\documentclass[10pt,twocolumn,letterpaper]{article}

\usepackage[pagenumbers]{cvpr} 
\usepackage[utf8]{inputenc} 
\usepackage[T1]{fontenc}    
\usepackage{url}            
\usepackage{booktabs}       
\usepackage{amsfonts}       
\usepackage{nicefrac}       
\usepackage{microtype}      
\usepackage{xcolor}         
\usepackage{xspace}
\usepackage{amsmath}        
\usepackage{graphicx}       
\usepackage{colortbl}       
\usepackage{multirow}
\usepackage{makecell}
\usepackage{anyfontsize}

\usepackage{graphicx}
\usepackage{amsmath}
\usepackage{amssymb}
\usepackage{booktabs}
\usepackage{float}
\usepackage{caption}
\usepackage{colortbl}
\usepackage{xcolor}
\usepackage{float}
\usepackage[symbol]{footmisc}

\usepackage{multirow}
\usepackage{diagbox}
\usepackage{pifont}

\usepackage{float}
\usepackage[utf8]{inputenc}
\usepackage{etoolbox}
\usepackage{silence}
\makeatletter
\robustify\@latex@warning@no@line
\makeatother
\makeatletter
\def\thanks#1{\protected@xdef\@thanks{\@thanks
\protect\footnotetext{#1}}}
\makeatother
\usepackage{authblk}
\usepackage{color}
\definecolor{Green}{RGB}{0,176,80}
\definecolor{Purple}{RGB}{112,48,160}

\usepackage{booktabs} 
\usepackage{float}    
\usepackage{caption}  

\usepackage[pagebackref,breaklinks,colorlinks,allcolors=cvprblue]{hyperref}
\usepackage[capitalize]{cleveref}

\usepackage[demo, export]{adjustbox}

\newcommand{\bE}{\mathbf{E}}
\newcommand{\bF}{\mathbf{F}} 

\newcommand{\bI}{\mathbf{I}}

\newcommand{\bM}{\mathbf{M}}

\newcommand{\bp}{\mathbf{p}}

\newcommand{\bT}{\mathbf{T}}
\newcommand{\bu}{\mathbf{u}}
\newcommand{\bV}{\mathbf{V}}

\newcommand{\bx}{\mathbf{x}}

\newcommand{\cL}{\mathcal{L}}

\newcommand{\eqnref}[1]{Eq.~\eqref{#1}}

\DeclareMathOperator*{\argmin}{argmin~}

\makeatletter
\DeclareRobustCommand\onedot{\futurelet\@let@token\@onedot}
\def\@onedot{\ifx\@let@token.\else.\null\fi\xspace}
\def\eg{e.g\onedot} 
\def\ie{i.e\onedot}

\def\etal{et~al\onedot}

\makeatother

\newcommand{\PAR}[1]{\vspace{0.1cm}\noindent{\bf #1} }

\newcommand{\norm}[1]{\left\lVert#1\right\rVert}

\makeatletter
\DeclareRobustCommand\onedot{\futurelet\@let@token\@onedot}
\def\@onedot{\ifx\@let@token.\else.\null\fi\xspace}
\def\eg{\emph{e.g}\onedot} 
\def\ie{\emph{i.e}\onedot}

\def\etal{\emph{et al}\onedot}
\makeatother

\newcommand{\methodname}{IncEventGS\xspace}

\definecolor{cvprblue}{rgb}{0.21,0.49,0.74}
\definecolor{cvprred}{rgb}{0.894, 0.0, 0.498}

\def\paperID{10699} 
\def\confName{CVPR}
\def\confYear{2025}

\title{\methodname: Pose-Free Gaussian Splatting from a Single Event Camera}

\author{Jian Huang$^{1,2}$\qquad Chengrui Dong$^{1,2}$\qquad Xuanhua Chen$^{2,3}$\qquad Peidong Liu$^{2*}$\thanks{*Corresponding author.} \\$^{1}$Zhejiang University\qquad $^{2}$Westlake University \qquad $^{3}$Northeastern University
	\\{\tt\small \{huangjian39, dongchengrui, chenxuanhua, liupeidong\}@westlake.edu.cn} \tt\small{\href{https://github.com/WU-CVGL/IncEventGS}{\textcolor{cvprred}{https://github.com/WU-CVGL/IncEventGS}}}}

\begin{document}

\maketitle
\begin{abstract}
Implicit neural representation and explicit 3D Gaussian Splatting (3D-GS) for novel view synthesis have achieved remarkable progress with frame-based camera (\eg RGB and RGB-D cameras) recently. Compared to frame-based camera, a novel type of bio-inspired visual sensor, \ie event camera, has demonstrated advantages in high temporal resolution, high dynamic range, low power consumption, and low latency, which make it favored for many robotic applications. 
In this work, we present \textit{\methodname}, an incremental 3D Gaussian Splatting reconstruction algorithm with a single event camera, without the assumption of known camera poses. 
To recover the 3D scene representation incrementally, we exploit the tracking and mapping paradigm of conventional SLAM pipelines for \textit{\methodname}. 
Given the incoming event stream, the tracker first estimates an initial camera motion based on prior reconstructed 3D-GS scene representation. The mapper then jointly refines both the 3D scene representation and camera motion based on the previously estimated motion trajectory from the tracker.
The experimental results demonstrate that \textit{\methodname} delivers superior performance
compared to prior NeRF-based methods and other related baselines, even if we do not have the ground-truth camera poses. Furthermore, our method can also deliver better performance compared to state-of-the-art event visual odometry methods in terms of camera motion estimation. 
\end{abstract}
\section{Introduction}
\label{sec:intro}

Reconstructing accurate 3D scene representations from 2D images has been a long-standing challenge in computer vision and robotics, driving substantial efforts over the past few decades. Among those pioneering works, Neural Radiance Fields (NeRF) \cite{mildenhall2021nerf} and 3D Gaussian Splatting (3D-GS) \cite{kerbl3Dgaussians}, stand out for their utilization of differentiable rendering techniques, and have garnered significant attention due to their capability to recover high-quality 3D scene representation from 2D images.
Commonly used sensors for 3D scene reconstruction are usually frame-based cameras, such as the RGB and RGB-D cameras. They usually capture full-brightness intensity images within a short exposure time at a regular frequency. Due to the characteristic of this data-capturing process, they often suffer from motion blur or fail to capture accurate and informative intensity information under fast motion and low-light conditions, which would further affect the performance of downstream applications. 

The event camera, a bio-inspired sensor, has gained significant attention in recent years for its potential to address the limitations of frame-based cameras under challenging conditions. Unlike conventional cameras, event cameras record brightness changes asynchronously at each pixel, emitting events when a predefined threshold is surpassed. This unique operation offers several advantages over conventional cameras, in terms of high temporal resolution, high dynamic range, low latency, and power consumption. Although event cameras have attractive characteristics for challenging environments, they cannot be directly integrated into existing frame-based 3D reconstruction algorithms that rely on processing dense 2D brightness intensity images. 

%
Several pioneering works have been proposed to exploit event stream \cite{Kim2016eccv,Rebecq2017ral,Gallego2018cvpr} to recover the motion trajectory and scene representation. While existing methods deliver impressive performance, they usually exploit 2.5D semi-dense depth maps to represent the 3D scene, and bundle adjustment (BA) is hardly performed, due to the asynchronous and sparse characteristics of event data stream. Klenk \etal \cite{klenk2024deep} recently proposed to convert event stream into event voxel grids, and then adapt a previous frame-based deep visual odometry pipeline \cite{teed2023deep} for accurate camera motion estimation. As NeRF exhibited impressive scene representation capability recently, several works \cite{klenk2023nerf, hwang2023ev, rudnev2023eventnerf, low2023_robust-e-nerf} explore to recover the underlying dense 3D scene NeRF representation from event stream, by assuming ground-truth poses are available. 

In contrast to those works, we propose \textit{\methodname}, an incremental dense 3D scene reconstruction algorithm from a single event camera, by exploiting Gaussian Splatting as the underlying scene representation. Different from prior event-based NeRF reconstruction methods, \textit{\methodname} does not require any ground-truth camera poses, which is more challenging and provides more flexibility for real-world robotic application scenarios. To overcome the challenges brought by unknown poses, \textit{\methodname} adopts the tracking and mapping paradigm of conventional SLAM pipelines \cite{mur2017orb}. In particular, \textit{\methodname} exploits prior explored and reconstructed 3D scenes for camera motion estimation of incoming event stream during the tracking stage. Both the 3D-GS scene representation and camera motions are then jointly optimized (\ie event-based bundle adjustment) during the mapping stage, for more accurate scene representation and motion estimation. The 3D scene is progressively expanded and densified. 
The experimental results on both synthetic and real datasets demonstrate that \textit{\methodname} can recover the underlying 3D scene representation and camera motion trajectory accurately. In particular, \textit{\methodname} outperforms prior NeRF-based methods and other related baselines in terms of scene representation recovery, even without ground-truth poses. Furthermore, our method also delivers better camera motion estimation accuracy than the most recent state-of-the-art visual odometry algorithm, in terms of the Absolute Trajectory Error (ATE) metric. The recovered 3D scene representation can be further used to render novel brightness images. Our main contributions can be summarized as follows: 
\begin{itemize}
\itemsep0em
\item We present an incremental 3D Gaussian Splatting reconstruction algorithm from a single event camera, without requiring the ground-truth camera poses.
\item We propose a novel initialization strategy tailored to the event data stream, which is vital to the success of the algorithm. 
\item The experimental results on both the synthetic and real datasets demonstrate the superior performance of our method over prior NeRF-based methods and related baselines in terms of novel view synthesis and better performance over state-of-the-art event-based visual odometry algorithm in terms of camera motion estimation.
\end{itemize}

\section{Related Works}
\label{sec:related_works}

We review two main areas of prior works: event-based neural radiance fields and 3D Gaussian Splatting, which are the most related to our work.

\PAR{Event-based Neural Radiance Fields.} Prior works \cite{klenk2023nerf, hwang2023ev, rudnev2023eventnerf} propose to exploit event stream to recover the neural radiance fields with known camera motion trajectory.  
Low \etal \cite{low2023_robust-e-nerf} further improves the reconstruction algorithm to handle sparse and noisy events under non-uniform motion.
The recovered neural radiance fields can then be used to render novel view brightness images. The ground-truth poses are usually computed from corresponding brightness images via COLMAP \cite{colmap} or provided by the indoor motion-capturing system. 
Recently, Qu \etal \cite{qu2023implicit} proposed to integrate event measurements into an RGB-D implicit neural SLAM framework and achieved robust performance in motion blur scenarios. 
Li \etal \cite{li2024benerf} also proposed exploiting event measurements and a single blurry image to recover the underlying neural 3D scene representation. In contrast to those works, \textit{\methodname} conducts incremental 3D scene reconstruction without requiring any prior ground-truth poses, which is more challenging and provides more flexibility for practical robotic application scenarios. The method further exploits 3D Gaussian Splatting as the underlying scene representation, which demonstrates better image rendering quality and efficiency, compared to the NeRF-based representation. 

\PAR{3D Gaussian Splatting.} 
3D Gaussian Splatting \cite{kerbl3Dgaussians} proposes a novel explicit 3D representation to further improve both the training and rendering efficiency compared to Neural Radiance Fields. Due to its impressive efficient scene representation capability, several pioneering works have been proposed to exploit 3D-GS for incremental 3D reconstruction. For example, Keetha \etal \cite{keetha2024splatam} propose an RGBD-based 3D-GS SLAM, employing an online tracking and mapping system tailored to the underlying Gaussian representation. Yan \etal \cite{yan2023gs} implement a coarse-to-fine camera tracking approach based on the sparse selection of Gaussians. Matsuki \etal \cite{matsuki2024gaussian} propose to apply 3D Gaussian Splatting to do incremental 3D reconstruction using a single moving monocular or RGB-D camera. Huang \etal \cite{huang2023photo} exploit ORB-SLAM3 to compute accurate camera poses and feed it into a 3D-GS algorithm for dense mapping. Fu \etal \cite{fu2023colmap} use monocular depth estimation with 3D-GS. Yugay \etal \cite{yugay2023gaussian} combine  DROID-SLAM \cite{teed2021droid} based camera tracking with active and inactive 3D-GS sub-maps. Wang \etal \cite{wang2024mbaslam} integrate bundle-adujustment and 3DGS \cite{zhao2024badgaussians} to estimated camera trajectory within the exposure time. Hu \etal \cite{hu2024cg} propose a novel depth uncertainty model to ensure the selection of valuable Gaussian primitives during optimization. While those methods deliver impressive performance in terms of 3D scene recovery and motion estimation, they usually assume the usage of frame-based images (\ie either RGB or RGB-D date). On the contrary, we propose to exploit pure event measurements for incremental 3D-GS reconstruction.  
Several concurrent studies have recently explored the use of 3D Gaussians for event-based reconstruction, including EvGGS \cite{wang2024evggs}, Event3DGS \cite{xiong2024event3dgs}, and E2GS \cite{deguchi2024e2gs}. EvGGS and Event3DGS rely solely on event data, whereas E2GS incorporates both event data and blurry images. However, the key difference between these methods and ours is that they all rely on ground-truth poses.

\section{Method}
\label{sec:method}

\begin{figure*}
    \centering
    \includegraphics[width=0.85\linewidth]{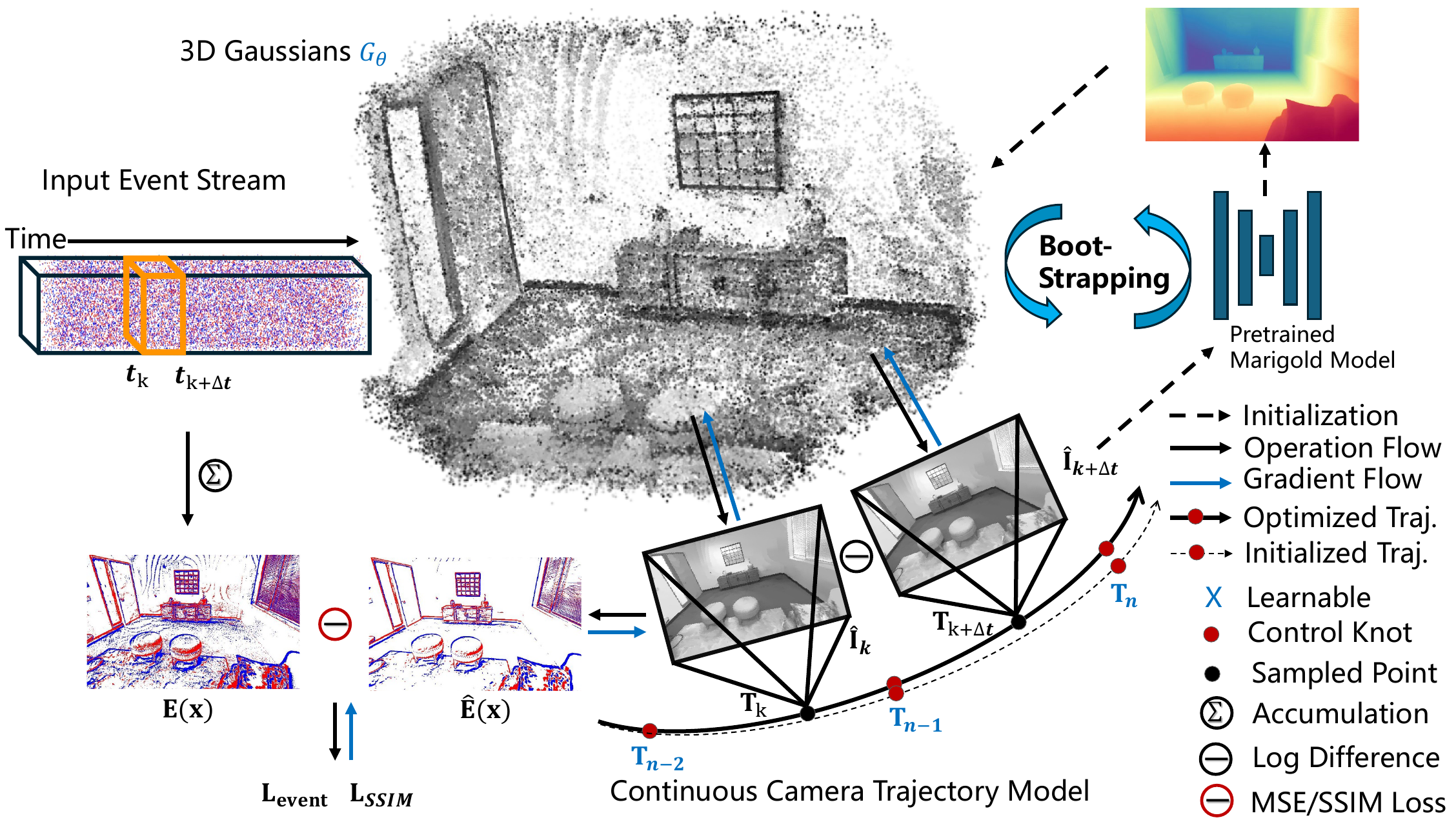}
    \caption{{\bf{The pipeline of \textit{\methodname}}}.  \textit{\methodname} processes incoming event stream by dividing it into chunks and representing the camera trajectory as a continuous model. It randomly samples two close consecutive timestamps to integrate the corresponding event streams. Two brightness images are rendered from 3D-GS at the corresponding poses, and we minimize the photometric loss between the synthesized and measured brightness change. During initialization, a pre-trained depth estimation model estimates depth from the rendered images to bootstrap the system.}
    \label{pic:pipeline}
    \vspace{-1.8em}
\end{figure*}

The overview of our \textit{\methodname} is shown in Fig. \ref{pic:pipeline}. Given only a single event camera, \textit{\methodname} incrementally performs tracking and dense mapping under the framework of 3D Gaussian Splatting, to recover both the camera motion trajectory and 3D scene representation simultaneously. 
Since event data are asynchronous, they cannot be directly integrated with the 3D-GS representation. We therefore process the event data stream into chunks according to a fixed time window. We associate each chunk with a continuous time trajectory parameterization in the $\mathfrak{se}(3)$ space. Two close consecutive timestamps (i.e., $t_{k}$ and $t_{k+\Delta t}$, where $\Delta t$ is a small time interval) can be randomly sampled, and the measured brightness change ${\bE}(x)$ can be computed from the corresponding event stream.
%
%
Based on the parameterized trajectory, the corresponding camera poses (\ie, $\bT_{k}$, $\bT_{k+\Delta t}$) can be determined and the brightness images (\ie, $\hat{\bI}_{k}$, $\hat{\bI}_{k+\Delta t}$) can be further rendered from the 3D-GS. The synthesized brightness change $\hat{\bE}(x)$ can be computed for event loss computation. 

During the tracking stage, we optimize only the camera motion trajectory of the newly accumulated event chunk and exploit the recovered trajectory to initialize the dense bundle adjustment (BA) algorithm for the mapping stage. During the mapping stage, we continuously densify 3D Gaussians for newly explored areas and prune transparent 3D Gaussians. For computational efficiency, we exploit a sliding window of the latest chunks and perform BA only within this window for both 3D-GS reconstruction and motion trajectory estimation. We will detail each component as follows.

\subsection{3D Scene Representation}
Following 3D-GS \cite{kerbl3Dgaussians}, the scene is represented by a set of 3D Gaussian primitives, each of which contains mean position $ \boldsymbol{\mu} \in \mathbb{R}^3$ in the world coordinate, 3D covariance $ \boldsymbol{\Sigma} \in \mathbb{R}^{3 \times 3}$, opacity $ \mathbf{o} \in \mathbb{R}$, and color $ \mathbf{c} \in \mathbb{R}^3$. To ensure that the covariance matrix remains positive semi-definite throughout the gradient descent, the covariance $\boldsymbol{\Sigma}$ is parameterized using a scale vector $ \mathbf{s} \in \mathbb{R}^3$ and rotation matrix $ \mathbf{R} \in \mathbb{R}^{3 \times 3}$:
\begin{equation}\label{equ:gaussian_dis}
	\boldsymbol{\Sigma}=\mathbf{R S S}^{T} \mathbf{R}^{T},
\end{equation}
where scale matrix  $\mathbf{S} = diag([s])$ is derived from the scale vector $ \mathbf{s} \in \mathbb{R}^3$.

In order to enable rendering, 3D-GS projects 3D Gaussian primitives to the 2D image plane from a given camera pose $ \mathbf{T}_{c}=\left\{\mathbf{R}_{c} \in \mathbb{R}^{3 \times 3}, \mathbf{t}_{c} \in \mathbb{R}^{3}\right\}$ using following equation:
\begin{equation}\label{equ:gaussian_3d_to_2d}
	\boldsymbol{\Sigma}^{\prime}=\mathbf{J R}_{c} \boldsymbol{\Sigma} \mathbf{R}_{c}^{T} \mathbf{J}^{T},
\end{equation}
where  $ \boldsymbol{\Sigma}^{\prime} \in \mathbb{R}^{2 \times 2} $ is the 2D covariance matrix, $ \boldsymbol{J} \in \mathbb{R}^{2 \times 3}$ is the Jacobian of the affine approximation of the projective transformation. After projecting 3D Gaussians onto the image plane, the color of each pixel is determined by sorting the Gaussians according to their depth and then applying near-to-far ${\alpha}$-blending rendering via the following equation:
\begin{equation}\label{equ:gaussian_alpha_blending}
	\mathbf{I}=\sum_{i}^{N} \mathbf{c}_{i} \alpha_{i} \prod_{j}^{i-1}\left(1-\alpha_{j}\right),
\end{equation}
where $\mathbf{c}_{i}$ is the learnable color of each Gaussian, and $\alpha_{i}$ is the alpha value computed by evaluating the 2D covariance  $ \boldsymbol{\Sigma}^{\prime}$ multiplied with the learned Gaussian opacity $\mathbf{o}$:
\begin{equation}\label{equ:alpha_c}
	\alpha_{i}=\mathbf{o}_{i} \cdot \exp \left(-\sigma_{i}\right), \quad \sigma_{i}=\frac{1}{2} \Delta_{i}^{T} \boldsymbol{\Sigma}^{\prime-1} \Delta_{i},
\end{equation}
where $\Delta_{i} \in \mathbb{R}^{2} $ is the offset between the pixel center and the 2D Gaussian center. Depth is rendered by:
\begin{equation}\label{equ:gaussian_d}
	\mathbf{D}=\sum_{i}^{N} \mathbf{d}_{i} \alpha_{i} \prod_{j}^{i-1}\left(1-\alpha_{j}\right),
\end{equation}
where ${d}_{i}$ denotes the z-depth of the center of the i-th 3D Gaussian to the camera. We also render alpha map to determine visibility:
\begin{equation}\label{equ:gaussian_alpha}
	\mathbf{V}=\sum_{i}^{N} \alpha_{i} \prod_{j}^{i-1}\left(1-\alpha_{j}\right),
\end{equation}

The derivations presented above demonstrate that the rendered pixel color, denoted as I in Eq. \eqref{equ:gaussian_alpha_blending}, is a function that is differentiable with respect to the learnable attributes of 3D-GS, and the camera poses $\bT_c$. This facilitates our bundle adjustment formulation, accommodating a set of event chunks and inaccurate camera motion trajectories within the framework of 3D-GS.

\subsection{Event Data Formation Model}
\vspace{-0.3em}

An event camera records changes in brightness as a stream of events asynchronously. 
%
To relate 3D-GS representation with the event stream, we sample two close consecutive timestamps (i.e., $t_{k}$ and $t_{k+\Delta t}$, where $\Delta t$ is a small time interval), and the measured brightness change between $\Delta t$ is:
\begin{equation}\label{eq_event_accumulate}
	\bE(\bx) = C \{e_i(\bx,t_i,p_i)\}_{t_k<t_i<t_k+\Delta t},
\end{equation}
where $e(\bx, t_i, p_i)$ is the $i^{th}$ event within the defined time interval corresponding to pixel $\bx$ and $C$ is the fixed contrast threshold. 
The corresponding camera poses \( T_k \) and \( T_{k + \Delta t} \) can be interpolated from the camera motion trajectory parameterization, and two corresponding brightness images (\ie $\hat{\bI}_{k}$ and $\hat{\bI}_{k+\Delta t}$) can be rendered from 3D-GS. The synthesized brightness change $\hat{\bE}$ is modelled as:
\begin{equation}
	\hat{\bE}(\bx) = \log(\hat{\bI}_{k+\Delta t}(\bx)) - \log(\hat{\bI}_{k}(\bx)),
\end{equation}
where $\hat{\bE}(\bx)$ depends on the parameters of both the motion trajectory parameters and 3D-GS, and is differentiable with respect to them.

Both in tracking and mapping, inspired by the work of \cite{rudnev2023eventnerf},  we segment the current event chunks into \( n_{seg} \) equal segments according to the number of events, obtaining \( n_{seg} \) timestamps that correspond to the end of each segment. We then randomly select one timestamp from these \( n_{seg} \) timestamps to serve as \( t_{k+\Delta t} \), and we randomly sample an integer \( n_{win} \) between the integer bounds \( n_{low} \) and \( n_{up} \). The index of \( t_{k} \) is equal to the index of \( t_{k+\Delta t} \) subtract $n_{win}$.  $n_{seg}$,  $n_{low}$ and $n_{up}$ are hyperparameters. This sampling strategy enables the model to capture both local and global information.

\subsection{Camera Motion Trajectory Modeling}
\label{subset:traj_interp}
Since each event chunk usually contains too many events, we sample a portion of them according to the total number of events during optimization. Following \cite{zhao2024badgaussians, wang2023badnerf}, we formulate the corresponding poses (\ie $\bT_{k}$ and $\bT_{k+\Delta t}$) at the beginning and end of the sampled event portion within each chunk, by employing a camera motion trajectory. The trajectory is represented through linear interpolation between two camera poses, one at the beginning of the chunk $\bT_\mathrm{start} \in \mathbf{SE}(3)$ and the other at the end $\bT_\mathrm{end} \in \mathbf{SE}(3)$. The camera pose at time $t_k$ can thus be expressed as follows:
\begin{equation} \label{eq_trajectory}
	\bT_k = \bT_\mathrm{start} \cdot \mathrm{exp}(\frac{t_k - t_{start}}{t_{end} - t_{start}} \cdot \mathrm{log}(\bT_\mathrm{start}^{-1} \cdot \bT_\mathrm{end})),
\end{equation} 
where $t_{start}$ and $t_{end}$ represent the timestamps corresponding to the boundary of the event chunk. It follows that $\bT_k$ is differentiable with respect to both $\bT_{\mathrm{start}}$ and $\bT_{\mathrm{end}}$. The objective of \textit{\methodname} is thus to estimate both $\bT_\mathrm{start}$ and $\bT_\mathrm{end}$ for each event chunk, along with the learnable parameters of 3D-GS $\mathbf{G}_\theta$. 

\subsection{Incremental Tracking and Mapping}
For both tracking and mapping, we aim to minimize the difference between the synthesized and measured brightness changes. In particular, we compute the loss of the latest event chunk only for the tracking stage and minimize the following loss function:
\begin{equation}
    \bT_{start}^{*}, \bT_{end}^{*} = \argmin_{\bT_{start}, \bT_{end}} \norm{\bE(\bx) - \hat{\bE}(\bx)}_2,
\end{equation}
where $\hat{\bE}(\bx)$ and $\bE(\bx)$ are the synthesized and measured brightness changes respectively, corresponding to a randomly sampled event portion within the latest event chunk.

Once the tracking is done, we insert the latest event chunk to the mapper and exploit the estimated $\bT_{start}^{*}$ and $\bT_{end}^{*}$ as the initial value of the chunk to perform dense bundle adjustment. For computational consideration, we exploit a sliding window BA of the latest $n_w$ chunks, and $n_w$ is a hyperparameter. In particular, we optimize both the motion trajectories and the 3D-GS jointly by minimizing the following loss functions:
\begin{align}
    \cL &= (1-\lambda)\cL_{event} + \lambda \cL_{ssim}, \label{eq:loss_eq}\\
    \cL_{event} &=  \norm{\bE_i(\bx) - \hat{\bE}_i(\bx)}_2, \label{eq:loss_event}\\
    \cL_{ssim} &= SSIM(\bE_i(\bx), \hat{\bE}_i(\bx)), \label{eq:loss_ssim}
\end{align}
where $\lambda$ is a hyperparameter, SSIM is the structural dissimilarity loss \cite{wang2004image}. As the event data streams in, we alternatively perform tracking and mapping. 

\subsection{System Initialization and Boot-strapping} 
\label{subsec:reinit}
Conventional frame-based 3D-GS methods usually require a good initial point cloud and camera poses computed via COLMAP \cite{schoenberger2016sfm} for initialization. However, they are usually not easy to obtain if we are only given a single event camera.
We therefore initialize the 3D-GS by sampling point cloud randomly within a bounding box. The first $m$ event chunks (where $m$ is a hyperparameter) are selected for initialization, and all corresponding camera poses are randomly initialized to be near the identity matrix. We then minimize the loss computed by \eqnref{eq:loss_eq} with respect to the attributes of 3D-GS and the parameters of camera motion trajectories jointly. 

Through experiments, we found that the above initialization procedure consistently produces satisfactory brightness images. However, the 3D structure remains of low quality due to the short baselines of the moving event camera. We further find that it could potentially affect the performance of the whole pipeline without a good initial 3D structure as more event data is received. Therefore, we utilize a monocular depth estimation network \cite{ke2023repurposing} to predict a dense depth map from the rendered brightness image after the pipeline is trained for certain iterations. This depth map is then used to re-initialize the centers of the 3D Gaussians by unprojecting the pixel depths, after which we repeat the minimization of \eqnref{eq:loss_eq} for system bootstrapping. More details about system re-initialization can be found in our supplementary material.

\PAR{3D-GS Incrementally Growing.} As the camera moves, new Gaussians is periodically introduced to cover newly explored regions. After tracking, we obtain an accurate camera pose estimate for each new event chunk. The centers of new Gaussians are determined by:
\begin{equation}
    \bp = \bT \cdot \pi^{-1}(\bu, d_u)
\end{equation}
where $\bu \in \mathbb{R}^2 $ is pixel coordinate in the image plane, $d_u$ is depth of the 3D point $\bp$ projecting onto the image plane, which is rendered by Eq. \eqref{equ:gaussian_d}, $\pi^{-1}$ denotes camera inverse projection,  $\bT$ is the camera pose from tracking. To ensure that new Gaussians are only added in previously unmapped areas, a visibility mask is computed to guide the expansion of the Gaussian splatting process, as following: 
\begin{equation}
    \bM(p) = V<\lambda_V,
\end{equation}
where $\bV$ is the rendered alpha map and $\lambda_V$ is a hyperparameter.

\section{Experiments}
\label{experiment}
\subsection{Experimental Setups.}
\paragraph{Implementation Details.} 

All experiments are conducted on a desktop PC equipped with a 5.73GHz AMD Ryzen 9 7900x CPU and an NVIDIA RTX 3090 GPU. The first $m=3$ event chunks are used for initialization. During the mapping stage, a sliding window size of $n_w = 20$ is employed for the bundle adjustment algorithm. The hyperparameters are set as follows: $\lambda = 0.05$, $\lambda_V = 0.8$, and $n_{seg} = 100$. For the synthetic dataset, $n_{low} = 400k$ and $n_{up} = 500k$, while for the real dataset, $n_{low} = 60k$ and $n_{up} = 80k$. Each event chunk has a time interval of 50 ms. The learning rate of the camera poses is set to 1e-4 and that for the attributes of 3D-GS are set the same as the original 3D-GS work. The number of optimization steps for initialization is 4500, and that for tracking and mapping are set to 200 and 1500 respectively. The contrast threshold $C$ of the event camera is set to 0.1 for synthetic datasets and 0.2 for real datasets empirically.

\PAR{Baselines and Evaluation Metrics.}
To the best of our knowledge, there are no existing event-only NeRF or 3D-GS methods that do not rely on ground-truth poses, making direct comparisons challenging. Therefore, we conduct a thorough comparison of our method with several event-based NeRF approaches, including E-NeRF \cite{klenk2023nerf}, EventNeRF \cite{rudnev2023eventnerf}, and Robust e-NeRF \cite{low2023_robust-e-nerf}, as well as our custom implemented two-stage method (\ie E2VID \cite{rebecq2019high} + COLMAP \cite{schoenberger2016sfm} + 3DGS \cite{kerbl3Dgaussians}). E-NeRF, EventNeRF, and Robust e-NeRF leverage implicit neural radiance fields for 3D scene representation, requiring ground-truth camera poses for accurate NeRF reconstruction. For E2VID + COLMAP + 3DGS, event data is first converted into brightness images using E2VID. The camera poses are then estimated from these images using COLMAP, and 3D-GS is trained with the generated images and poses. Both the quantitative and qualitative comparisons are performed on the synthetic dataset. Since there are no paired ground truth images for the real dataset, we only perform qualitative comparisons on the real dataset. In terms of motion trajectory evaluations, we use the publicly available state-of-the-art event-only visual odometry methods, \ie DEVO (mono) \cite{klenk2024deep} and ESVO2(stereo) \cite{niu2025esvo2}, for comparison.

\begin{figure*}
    \centering
    \addtolength{\tabcolsep}{-6.5pt}
    \footnotesize{
        \setlength{\tabcolsep}{1pt} 
        \begin{tabular}{p{8.2pt}cccccccc}
            & room0 & room2 & office0 & office2 & office3  \\
        \raisebox{18pt}{\rotatebox[origin=c]{90}{\tiny E-NeRF}}&
         \includegraphics[width=0.186\textwidth]{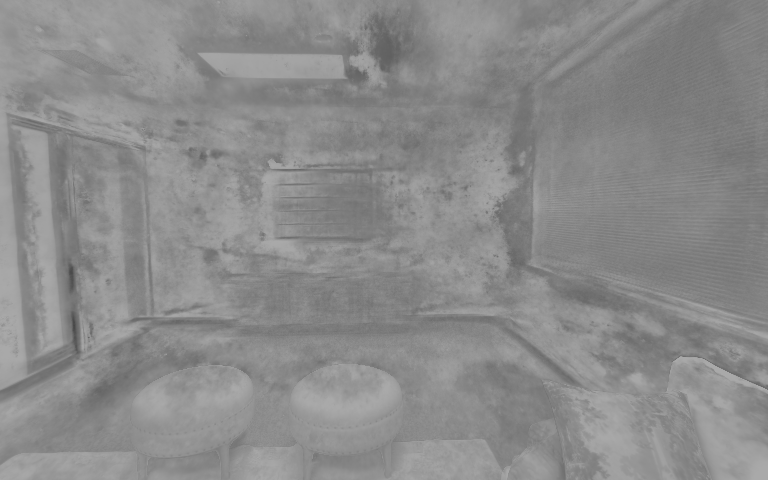} &
        \includegraphics[width=0.186\textwidth]{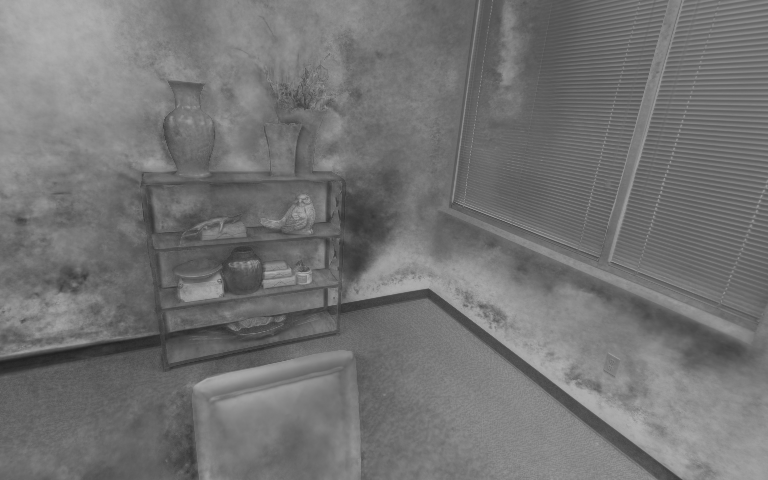} &
        \includegraphics[width=0.186\textwidth]{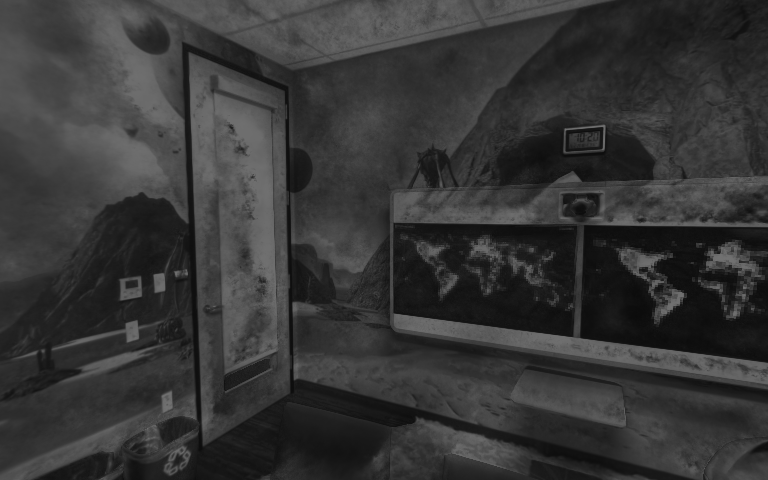} &
        \includegraphics[width=0.186\textwidth]{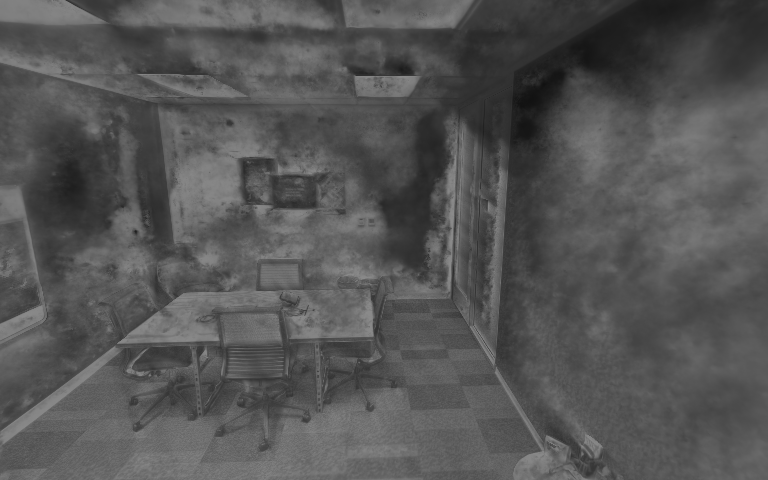} &
        \includegraphics[width=0.186\textwidth]{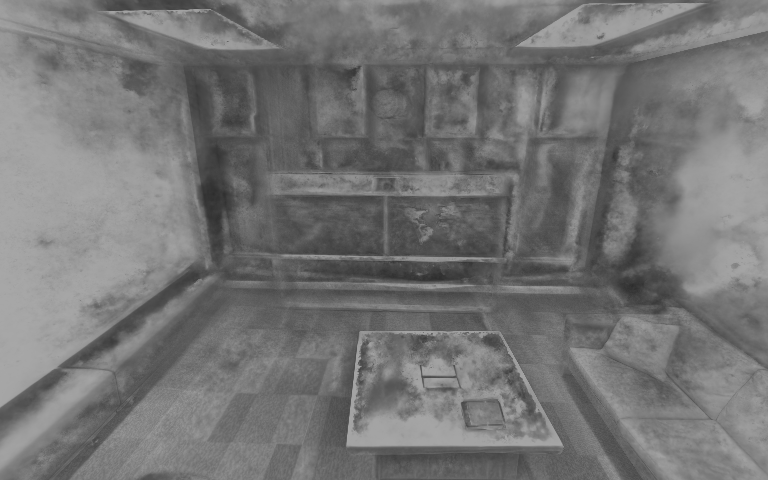} 
         \\

        \raisebox{18pt}{\rotatebox[origin=c]{90}{\tiny EventNeRF}}&
         \includegraphics[width=0.186\textwidth]{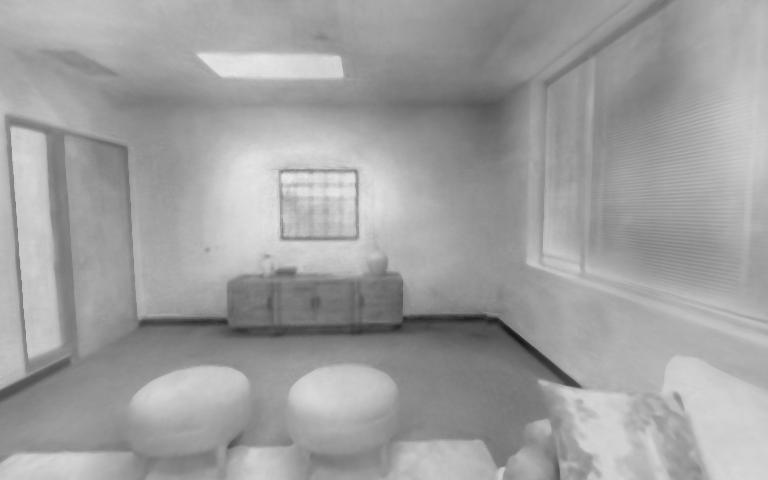} &
        \includegraphics[width=0.186\textwidth]{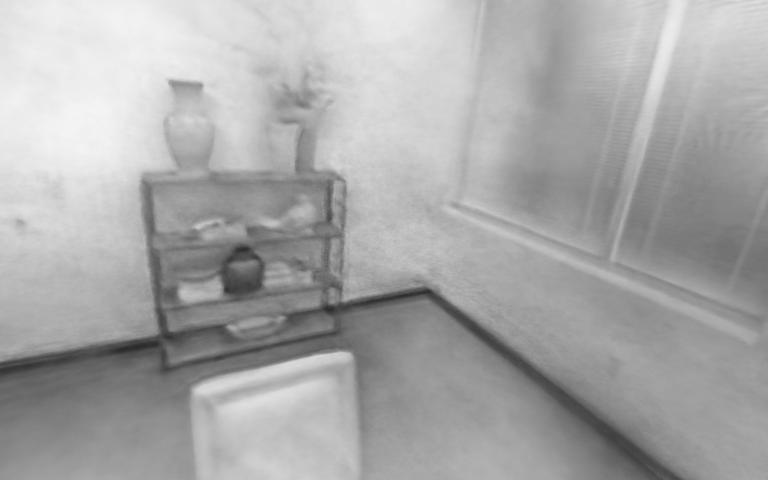} &
        \includegraphics[width=0.186\textwidth]{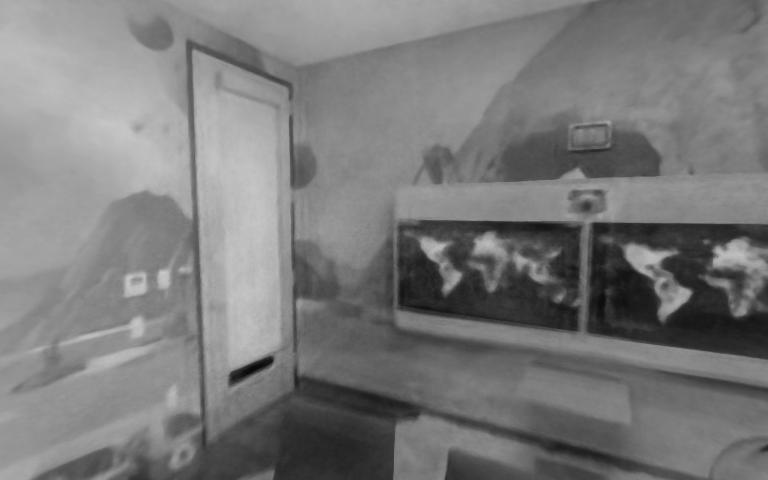} &
        \includegraphics[width=0.186\textwidth]{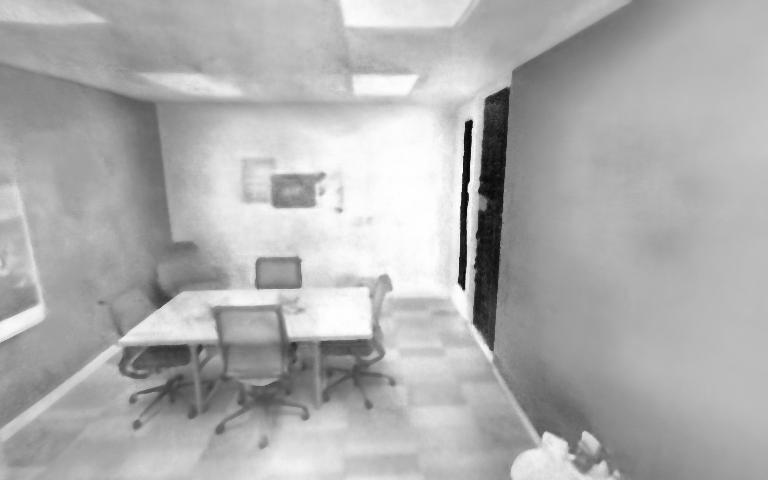} &
        \includegraphics[width=0.186\textwidth]{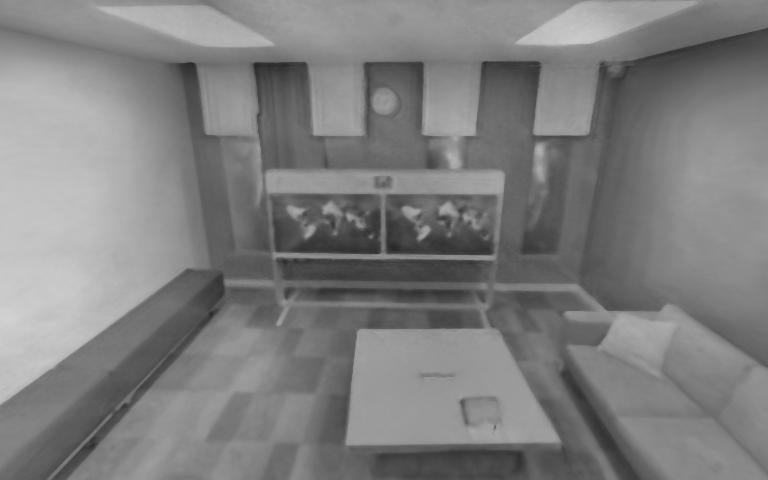} 
         \\

        \raisebox{18pt}{\rotatebox[origin=c]{90}{\tiny Robust e-NeRF}}&
         \includegraphics[width=0.186\textwidth]{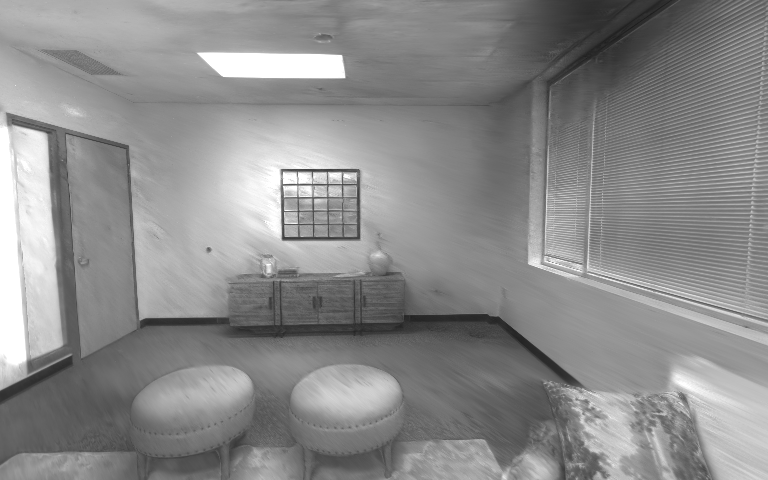} &
        \includegraphics[width=0.186\textwidth]{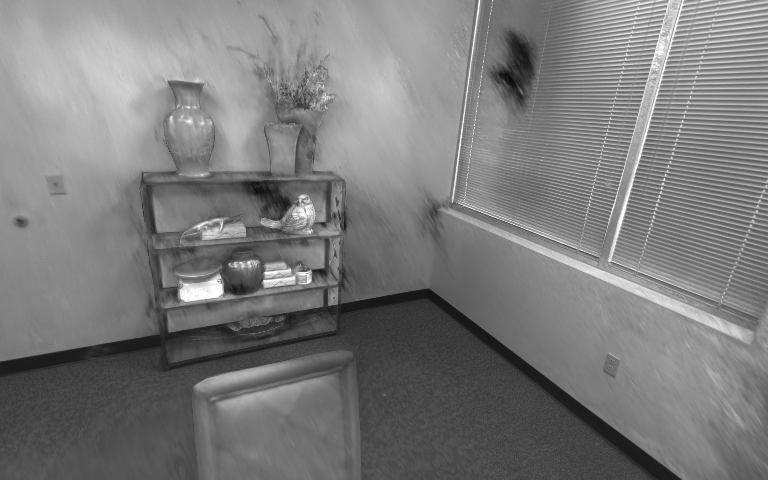} &
        \includegraphics[width=0.186\textwidth]{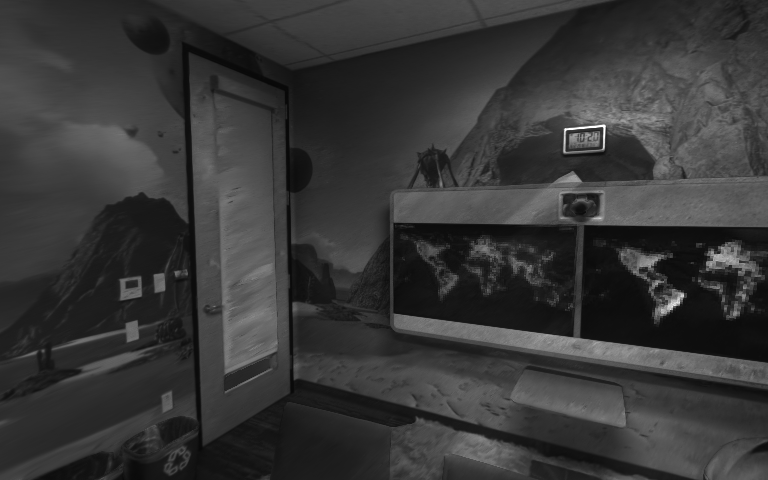} &
        \includegraphics[width=0.186\textwidth]{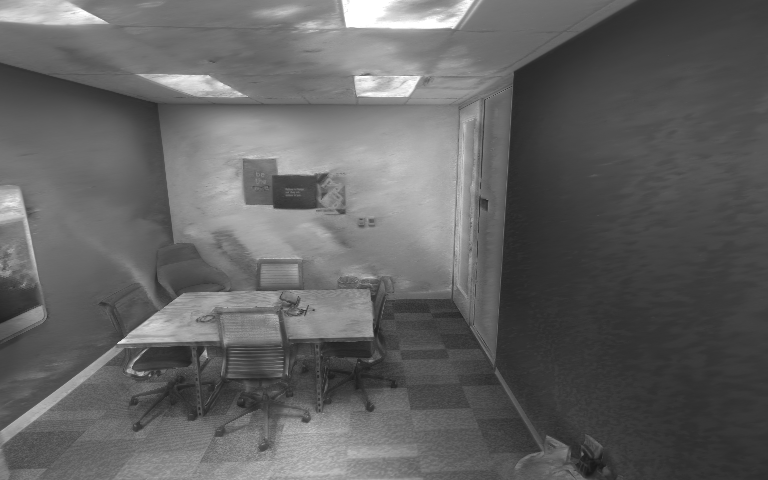} &
        \includegraphics[width=0.186\textwidth]{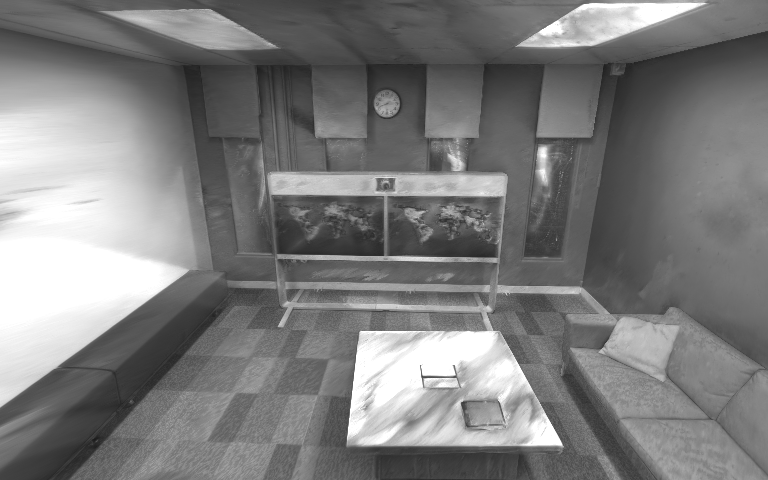} 
         \\

        \raisebox{18pt}{\rotatebox[origin=c]{90}{ \scalebox{0.6}{ \shortstack{E2VID+\\COLMAP+3DGS}} }}&
         \includegraphics[width=0.186\textwidth]{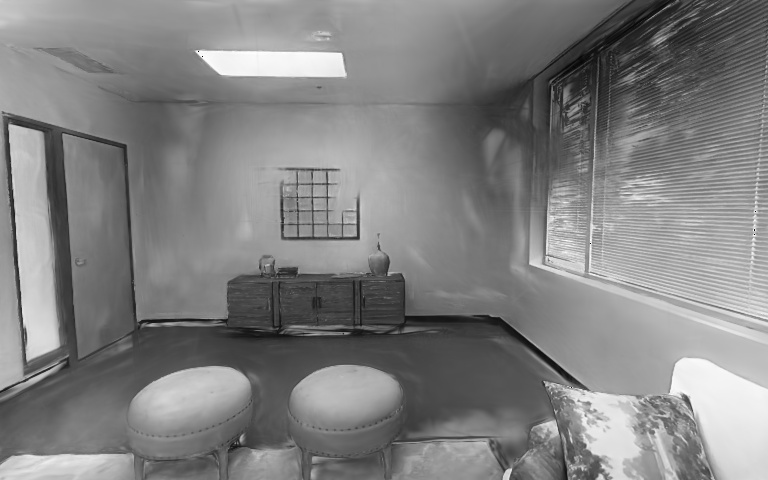} &
        \includegraphics[width=0.186\textwidth]{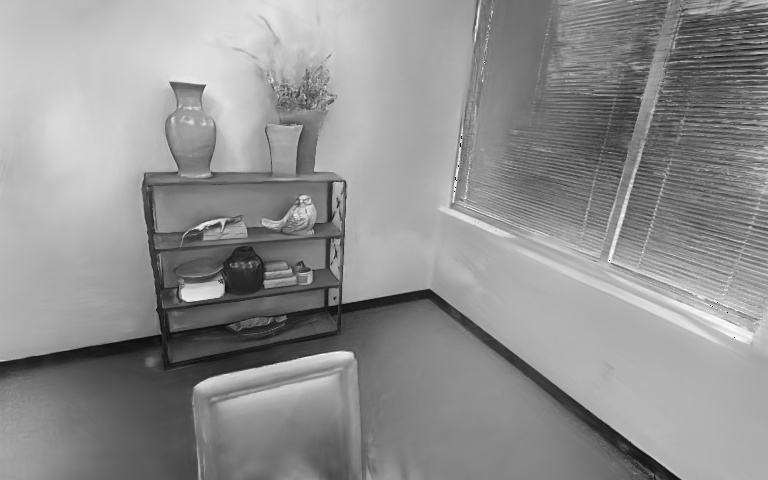} &
        \includegraphics[width=0.186\textwidth]{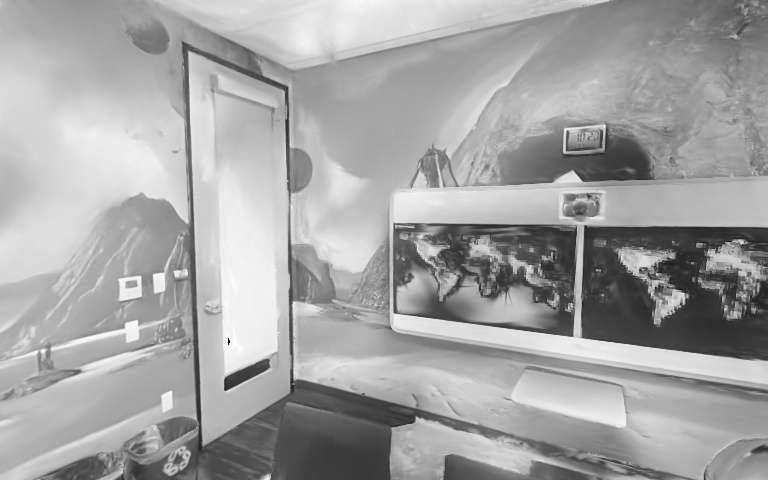} &
        \includegraphics[width=0.186\textwidth]{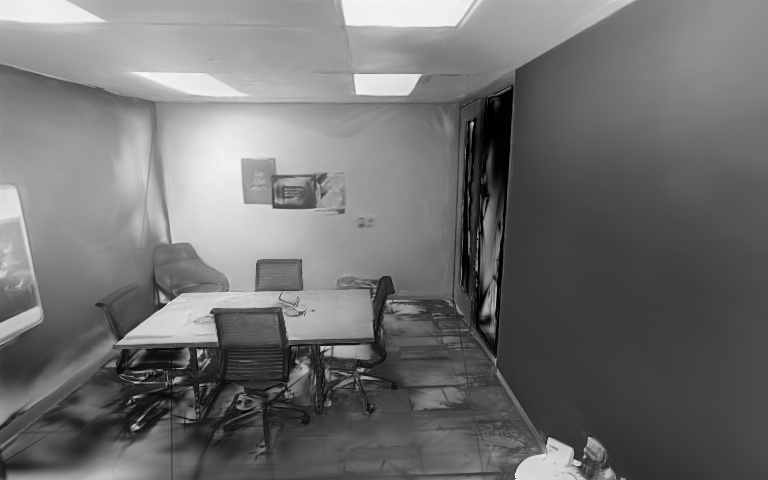} &
        \includegraphics[width=0.186\textwidth]{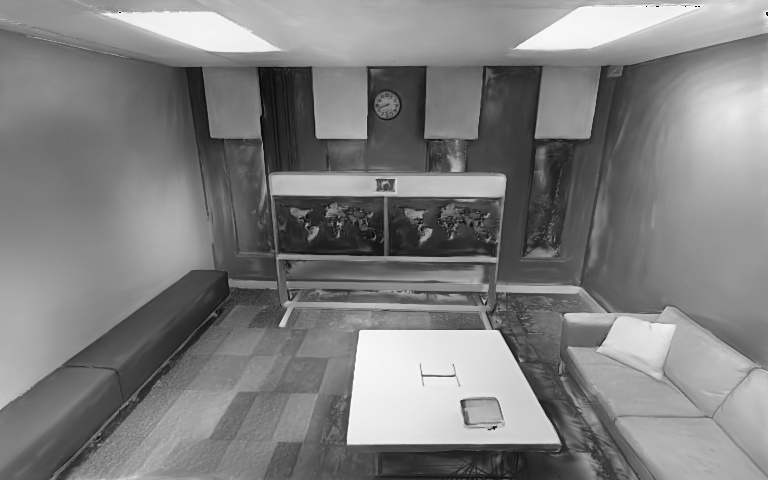} 
         \\


        \raisebox{18pt}{\rotatebox[origin=c]{90}{\tiny Ours}}&
         \includegraphics[width=0.186\textwidth]{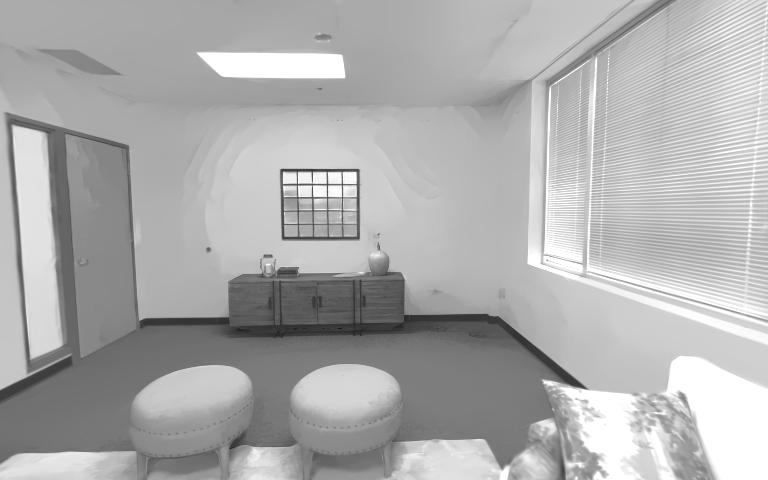} &
        \includegraphics[width=0.186\textwidth]{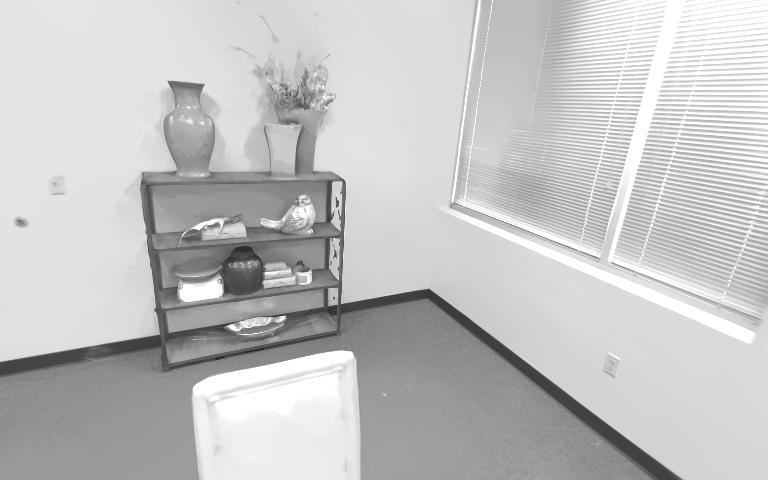} &
        \includegraphics[width=0.186\textwidth]{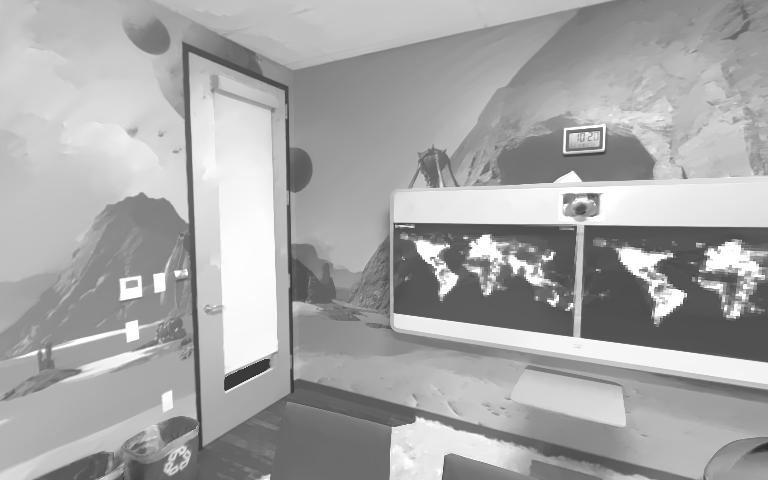} &
        \includegraphics[width=0.186\textwidth]{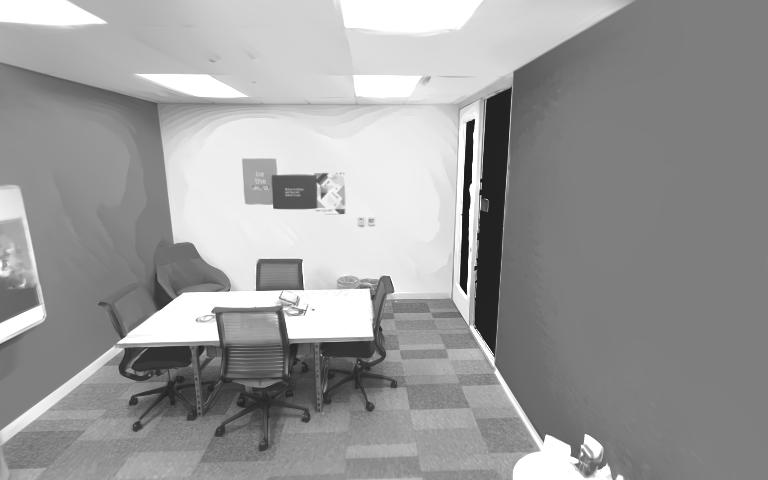} &
        \includegraphics[width=0.186\textwidth]{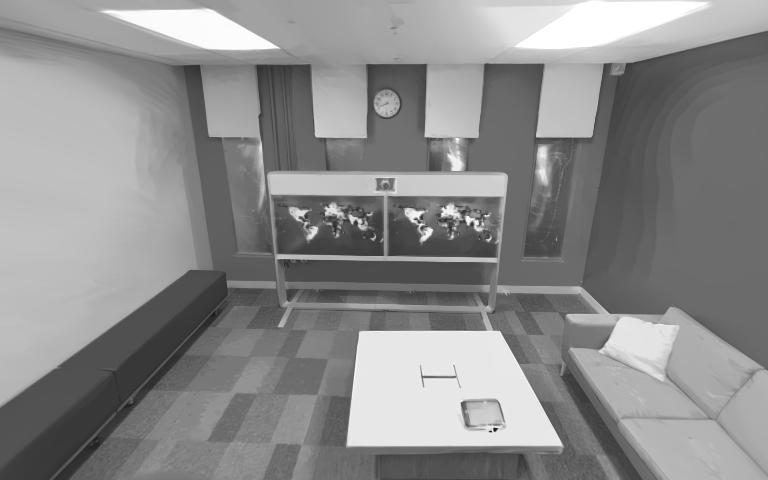} 
         \\

        \raisebox{18pt}{\rotatebox[origin=c]{90}{\tiny GT}}&
         \includegraphics[width=0.186\textwidth]{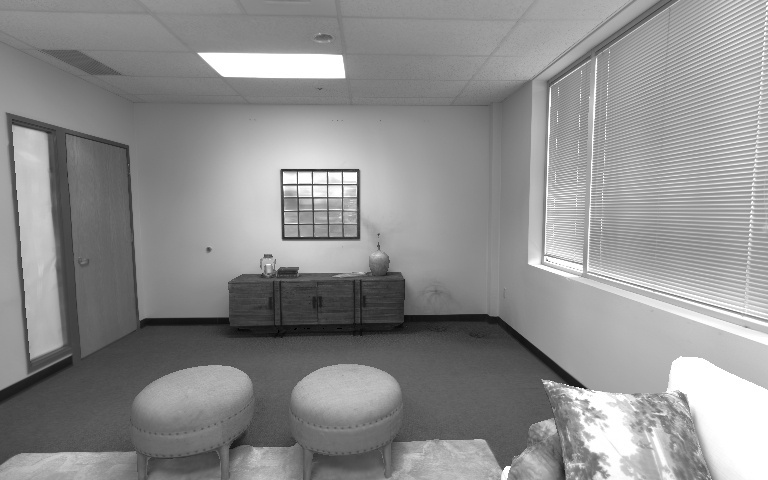} &
        \includegraphics[width=0.186\textwidth]{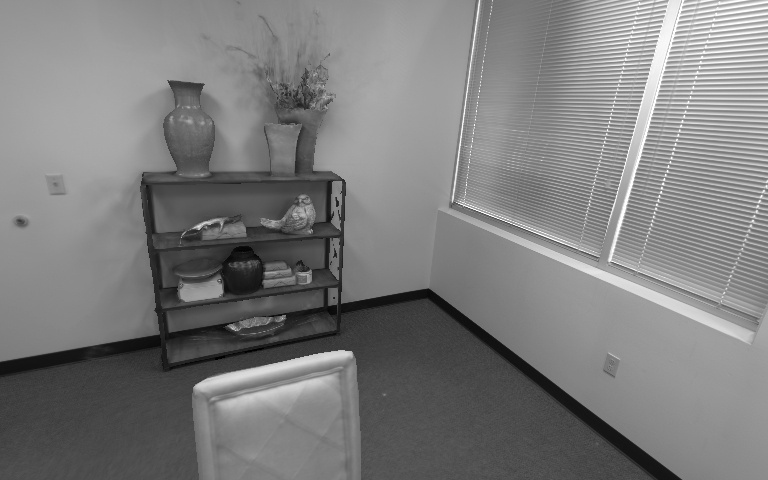} &
        \includegraphics[width=0.186\textwidth]{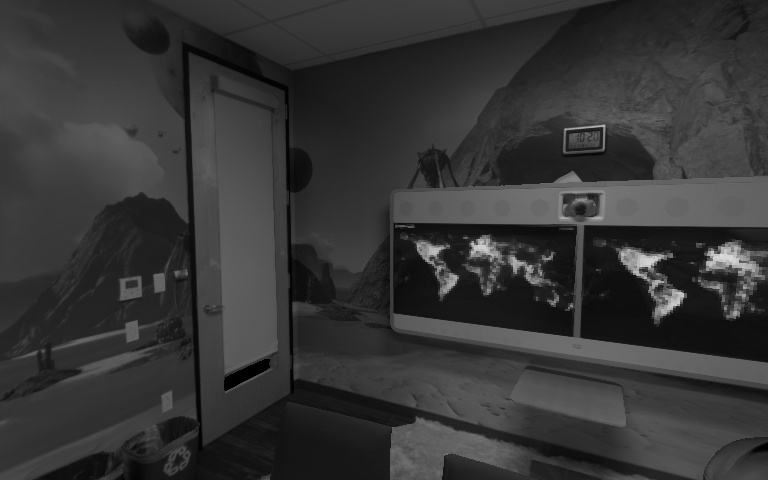} &
        \includegraphics[width=0.186\textwidth]{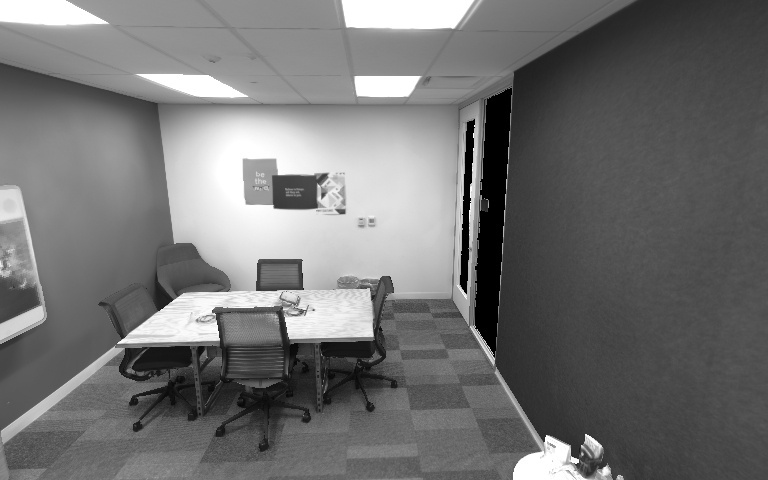} &
        \includegraphics[width=0.186\textwidth]{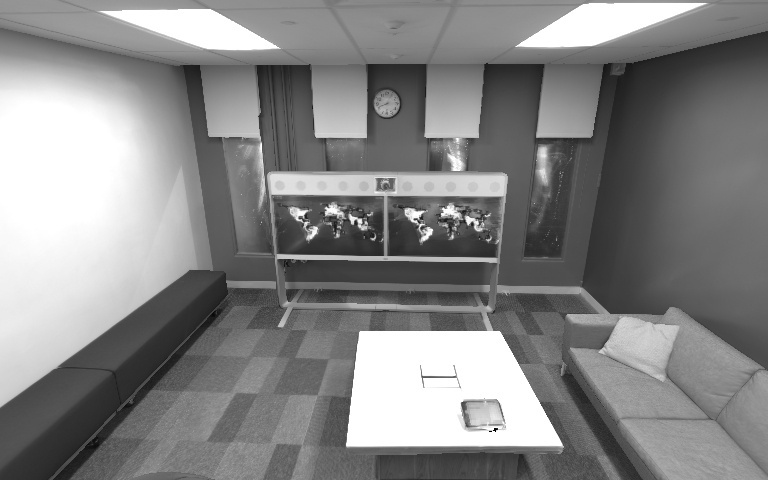} 
         \\

        \end{tabular}
    }
    \vspace{-5pt}
    \caption{Qualitative evaluation of novel view image synthesis on the Replica dataset. The experimental results demonstrate that our method renders higher-quality images with fewer artifacts compared to event-based NeRF and two-stage approaches.}
    \label{pic:syn_rendering}
    \vspace{-1.8em}
\end{figure*}

The metrics used for novel view synthesis (NVS) include the commonly employed PSNR, SSIM, and LPIPS. For motion trajectory evaluations, we utilize Absolute Trajectory Error (ATE). To ensure fair comparisons, we employ the evaluation code provided by EventNeRF to compute the NVS metrics, which applies a linear color transformation between predictions and ground truth. Additionally, we use the public EVO toolbox \cite{grupp2017evo} to compute the trajectory metrics.

\paragraph{Benchmark Datasets.}
To properly evaluate the performance of NVS and motion trajectory estimation, we synthesized event data using the 3D scene models from the Replica dataset \cite{replica19arxiv}. In particular, we exploit the \textit{room0}, \textit{room2}, \textit{office0}, \textit{office2}, and \textit{office3} scenes. We render high frame rate RGB images at 1000 Hz with a resolution of 768x480 pixels. These images are then converted to grayscale, and the event data is generated via the events simulator \cite{gehrig2020video}. The contrast threshold is set to 0.1. To simulate real-world camera motions, we exploit the same motion trajectories as that of NICE-SLAM \cite{zhu2022nice} for data generation. 

We use the event dataset provided by TUM-VIE \cite{klenk2021tum} for real data evaluations, which is also used by E-NeRF and Robust e-NeRF. TUM-VIE captures the event datasets by a pair of Prophesee Gen4 HD event cameras with a resolution of 1280x720 pixels. We only use the left-event camera data for our experiment. 

\begin{table*}
    \centering
    \vspace{1em}
    \resizebox{\linewidth}{!}{
    \begin{tabular}{c*{15}{c}}
        \toprule
        \multirow{3}{*}{} & \multicolumn{3}{c}{\textit{room0}} & \multicolumn{3}{c}{\textit{room2}} & \multicolumn{3}{c}{\textit{office0}} & \multicolumn{3}{c}{\textit{office2}} & \multicolumn{3}{c}{\textit{office3}} \\
        \cmidrule(lr){2-4} \cmidrule(lr){5-7} \cmidrule(lr){8-10} \cmidrule(lr){11-13} \cmidrule(lr){14-16}
        & {\scriptsize PSNR$\uparrow$} & {\scriptsize SSIM$\uparrow$} & {\scriptsize LPIPS$\downarrow$}  & {\scriptsize PSNR$\uparrow$} & {\scriptsize SSIM$\uparrow$} & {\scriptsize LPIPS$\downarrow$}  & {\scriptsize PSNR$\uparrow$} & {\scriptsize SSIM$\uparrow$} & {\scriptsize LPIPS$\downarrow$}  & {\scriptsize PSNR$\uparrow$} & {\scriptsize SSIM$\uparrow$} & {\scriptsize LPIPS$\downarrow$}  & {\scriptsize PSNR$\uparrow$} & {\scriptsize SSIM$\uparrow$} & {\scriptsize LPIPS$\downarrow$}  \\
        \midrule
        E-NeRF       & 13.99 & 0.58 & 0.51 & 15.56 & 0.47 & 0.58 & 18.91 & 0.51 & 0.57 & 13.05 & 0.65 & 0.44 & 14.01 & 0.62 & 0.48 \\
            EventNeRF     & 17.29 & 0.62 & 0.39 & 16.02 & 0.54 & 0.64 & 18.90 & 0.43 & 0.62 & 15.18 & 0.66 & 0.45 & 16.77 & 0.73 & 0.33 \\
            Robust e-NeRF & 17.26 & 0.84 & 0.18 & 16.43 & 0.50 & 0.52 & 18.93 & 0.52 & 0.56 & 16.81 & 0.81 & 0.25 & 19.22 & 0.84 & 0.18 \\
            \scalebox{0.8}{ \shortstack{E2VID+\\COLMAP+3DGS}}  & 14.45 & 0.44 & 0.52 & 15.74 & 0.51 & 0.55 & 18.91 & 0.31 & 0.68 & 14.03 & 0.57 & 0.48 & 13.25 & 0.47 & 0.53 \\
            Ours       & \textbf{24.31} & \textbf{0.85} & \textbf{0.17} & \textbf{23.75}  & \textbf{0.79} & \textbf{0.23} & \textbf{25.64} & \textbf{0.54} & \textbf{0.30} & \textbf{21.74} & \textbf{0.82} & \textbf{0.23} & \textbf{21.18} & \textbf{0.88} & \textbf{0.13}   \\
        \bottomrule
    \end{tabular}
    }
    \caption{NVS performance comparison on Replica dataset. All results are calculated using the same code and ground-truth images. The results demonstrate that our method outperforms NeRF-based and two-stage methods.}
    \label{tab:render1}
\end{table*}

\begin{table*}
	\centering
        \fontsize{9pt}{10pt}\selectfont
	\begin{tabular}{ccccccccccc}
		\toprule
		    & room0  & room2  & office0 & office2&office3 & 1d   & 3d  &6dof  & desk   & desk2 \\
            \midrule
            DEVO          & 0.289  & 0.266  & 0.138   & 0.281  & 0.156  &0.147 &0.303& 2.93 & 0.732 & 0.201 \\
            E2VID+COLMAP  & 17.93  & 59.96  & 105.19  & 18.414 & 17.28  &4.268   &16.90 & 9.88  & 21.57  & 10.13 \\
            ESVO2  & -  & -  & -  & - & -  & 0.337   &1.066 & 0.587  & 1.147  & 2.506 \\
            Ours          & \textbf{0.046} & \textbf{0.067} & \textbf{0.045}  & \textbf{0.046} & \textbf{0.054}  &\textbf{0.115} &\textbf{0.298} & \textbf{0.251} & \textbf{0.231}   & \textbf{0.129} \\
		\bottomrule
	\end{tabular}
    \caption{Pose accuracy (ATE, cm) on Replica and TUM-VIE datasets. The results demonstrate that our method delivers better performance in terms of camera motion estimation. Since ESVO2 is a stereo event camera method, it cannot run on the Replica dataset.}
    \label{traj_trans}
\end{table*}

\subsection{Ablation Study}

\begin{table}[]
    \centering
    \fontsize{9pt}{10pt}\selectfont
    \begin{tabular}{lllll}
            \toprule
            Setting & {\scriptsize PSNR$\uparrow$} & {\scriptsize SSIM$\uparrow$} & {\scriptsize LPIPS$\downarrow$} & ATE \\ 
            \midrule
            full    & \textbf{21.74}  & \textbf{0.82} & \textbf{0.23} & \textbf{0.046} \\	
            w/o     & 17.80  & 0.76 & 0.26 & 1.534 \\
            \bottomrule
        \end{tabular}
    \caption{Ablation Study about Depth Initialization. The unit of ATE is cm. The experimental results demonstrate the effectiveness of the initialization strategy. It not only improves the quality of rendered images, but also improves the accuracy of the camera motion estimation significantly.}
    \label{ablation_d}
\end{table}

\begin{table}[]
    \centering
    \fontsize{9pt}{10pt}\selectfont
    \begin{tabular}{lllll}
            \toprule
            Setting & {\scriptsize PSNR$\uparrow$} & {\scriptsize SSIM$\uparrow$} & {\scriptsize LPIPS$\downarrow$} & ATE  \\ 
            \midrule
            1k-10k     & 16.07  & 0.64 & 0.46 & 0.167  \\
            10k-50k     & 18.41  & 0.72 & 0.33 & 0.079  \\
            80k-200k    & 20.99  & 0.79 & 0.25 & 0.079  \\
            \textbf{400k-500k}   & \textbf{21.74}  & \textbf{0.82} & \textbf{0.23} & \textbf{0.046}  \\
            500k-600k   & 20.95  & 0.79 & 0.23 & 0.050  \\
            600k-700k   & 18.06  & 0.75 & 0.28 & 0.214  \\
            \bottomrule
        \end{tabular}
    \caption{Ablation Study on Event Slice Window Size (Hyperparameters $n_{\text{low}}$ and $n_{\text{up}}$). The unit of ATE is cm.}
    \label{ablation_win}
    \vspace{-1.8em}
\end{table}

We conduct ablation studies to confirm our design choices. In particular, we study the effect of a monocular depth estimation network for system bootstrapping and event slicing hyperparameters $n_{low}$, $n_{up}$. The experiments are conducted with the Replica dataset, and the results are shown in Table \ref{ablation_d} and Table \ref{ablation_win}, respectively. 

We found that depth initialization significantly impacts pose estimation, reducing the Average Trajectory Error (ATE) from 1.534 cm to 0.064 cm. Additionally, this improvement in pose estimation leads to a slight enhancement in Novel View Synthesis (NVS) performance. These results verify the importance of using depth initialization during the boot-strapping stage.

We compare several combinations of hyperparameters $n_{\text{low}}$ and $n_{\text{up}}$, which refer to the range of event slicing window size. Table \ref{ablation_win} demonstrates that both too small and too large window sizes negatively impact the performance of Novel View Synthesis (NVS) and pose estimation. Consequently, we select $n_{\text{low}} = 400k$ and $n_{\text{up}} = 500k$ for our experiments on the Replica dataset.

\subsection{Quantitative Evaluations.}
We conduct quantitative evaluations against event NeRF methods(E-NeRF, EventNeRF, and Robust e-NeRF) and our custom implemented two-stage method (\ie E2VID + COLMAP + 3DGS) in terms of the quality of NVS and pose estimation performance. 

The NVS performance is evaluated on Replica-dataset and the results are presented in Table \ref{tab:render1}. It is important to note that the metrics are lower than those typically observed in standard NeRF/3D-GS methods for RGB images, primarily due to the lack of absolute brightness supervision. Even though NeRF-based methods use ground truth poses for training, \textit{\methodname} still significantly outperforms them, highlighting the advantages of our approach utilizing a 3D Gaussian representation. Additionally, our method greatly surpasses two-stage method that also employs 3D Gaussian representation, demonstrating superior pose estimation and the effectiveness of our bundle adjustment technique.

We evaluate pose estimation performance using the ATE metric on both synthetic and real datasets, comparing our method with DEVO, ESVO2 and E2VID + COLMAP. The results, presented in Table \ref{traj_trans}, show that our method outperforms both baselines, validating the effectiveness of our incremental tracking and mapping technique.

\begin{figure*}
    \vspace{-0.5em}
    \centering
    \addtolength{\tabcolsep}{-6.5pt}
    \footnotesize{
        \setlength{\tabcolsep}{1pt} 
        \begin{tabular}{p{8.2pt}cccccccc}
            & desk & desk2 & 1d-trans & 3d-trans & 6dof  \\

        \raisebox{18pt}{\rotatebox[origin=c]{90}{ \scalebox{0.6}{ \shortstack{E2VID+\\COLMAP+3DGS}}  }}&
         \includegraphics[width=0.186\textwidth]{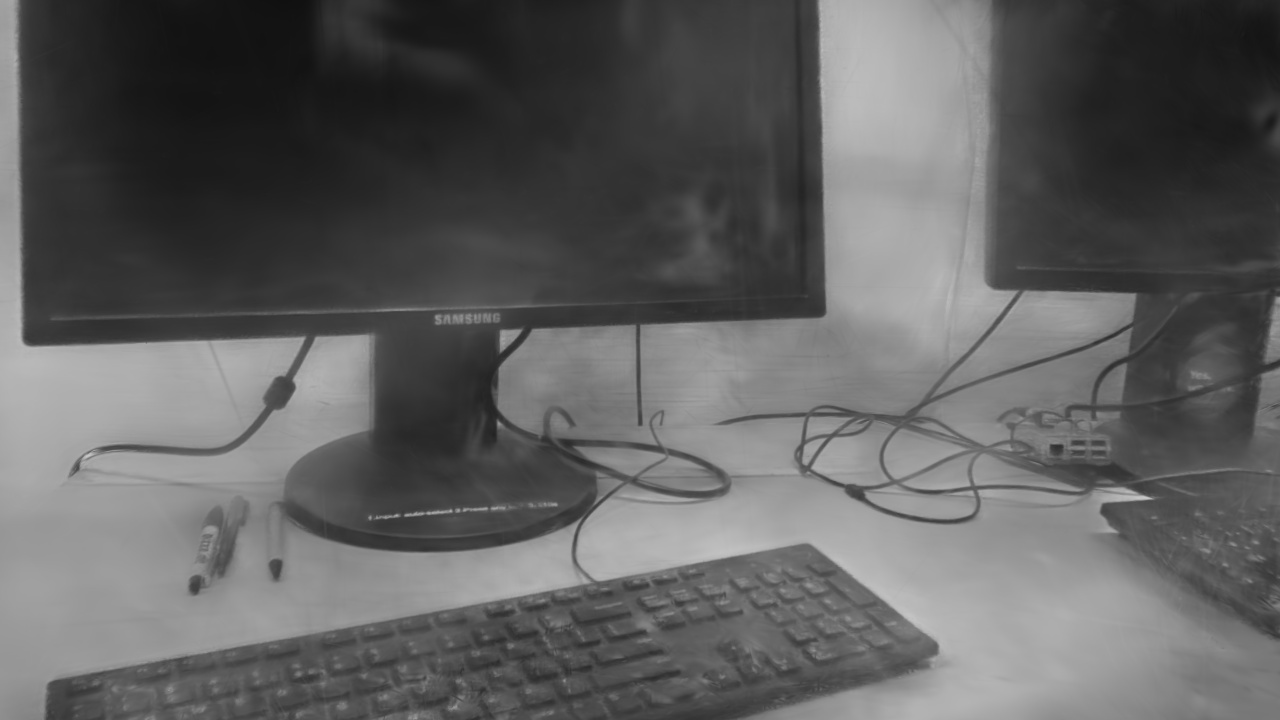} &
        \includegraphics[width=0.186\textwidth]{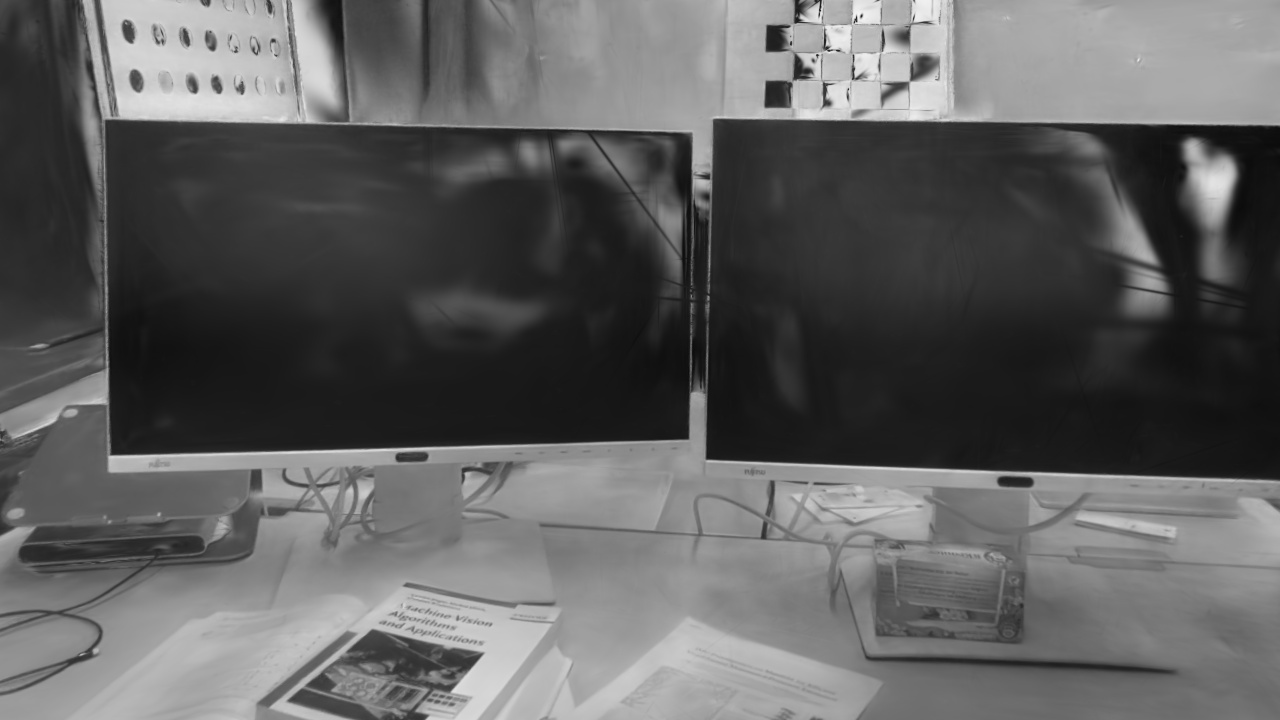} &
        \includegraphics[width=0.186\textwidth]{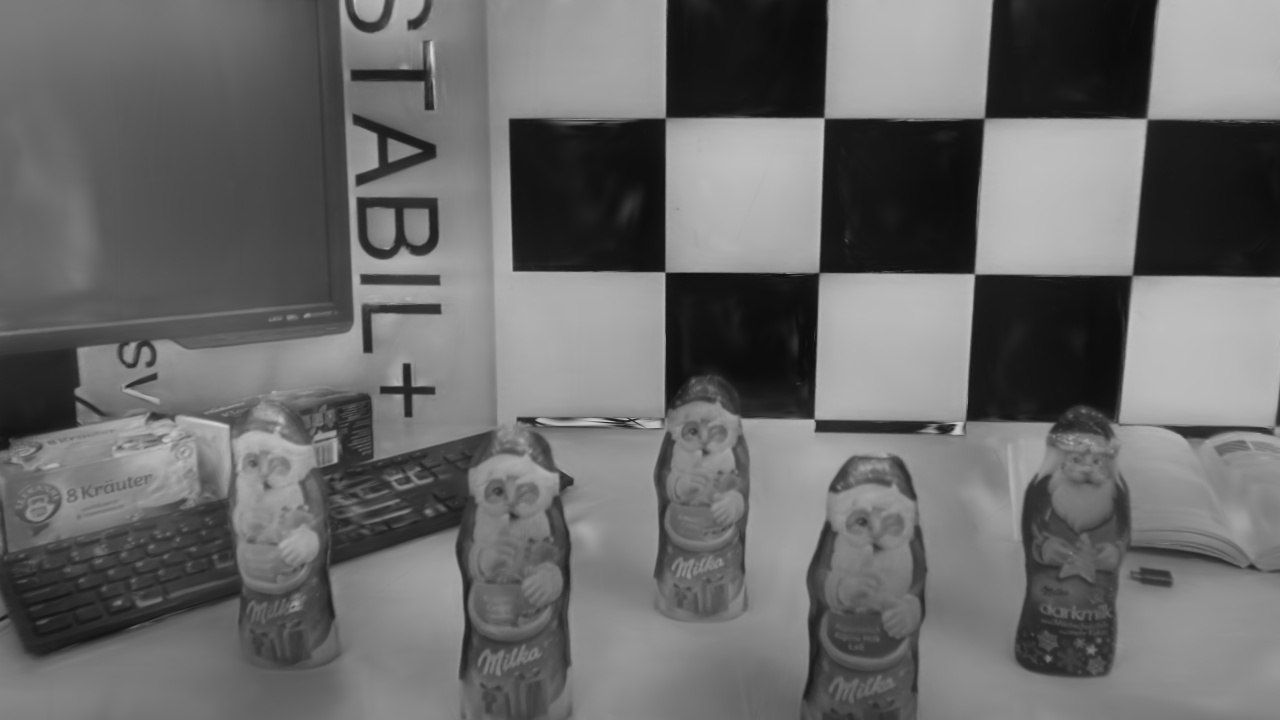} &
        \includegraphics[width=0.186\textwidth]{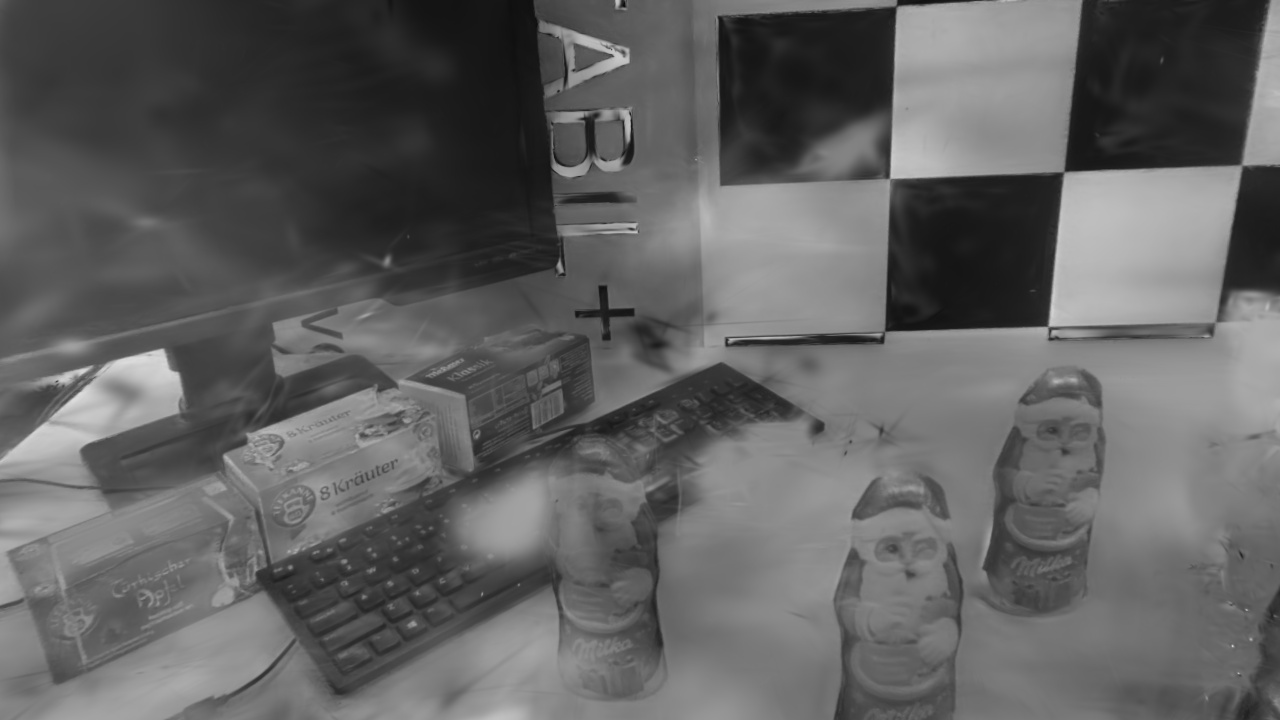} &
        \includegraphics[width=0.186\textwidth]{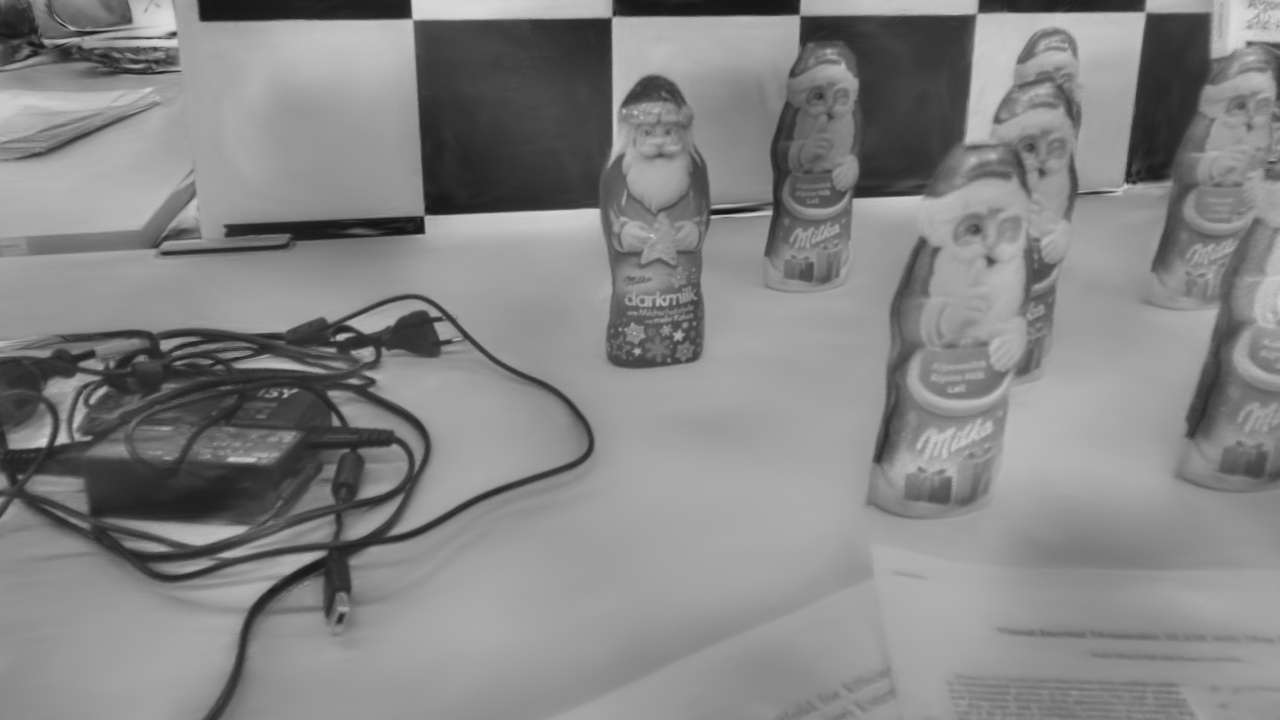} 
         \\

        \raisebox{18pt}{\rotatebox[origin=c]{90}{\tiny E-NeRF}}&
         \includegraphics[width=0.186\textwidth]{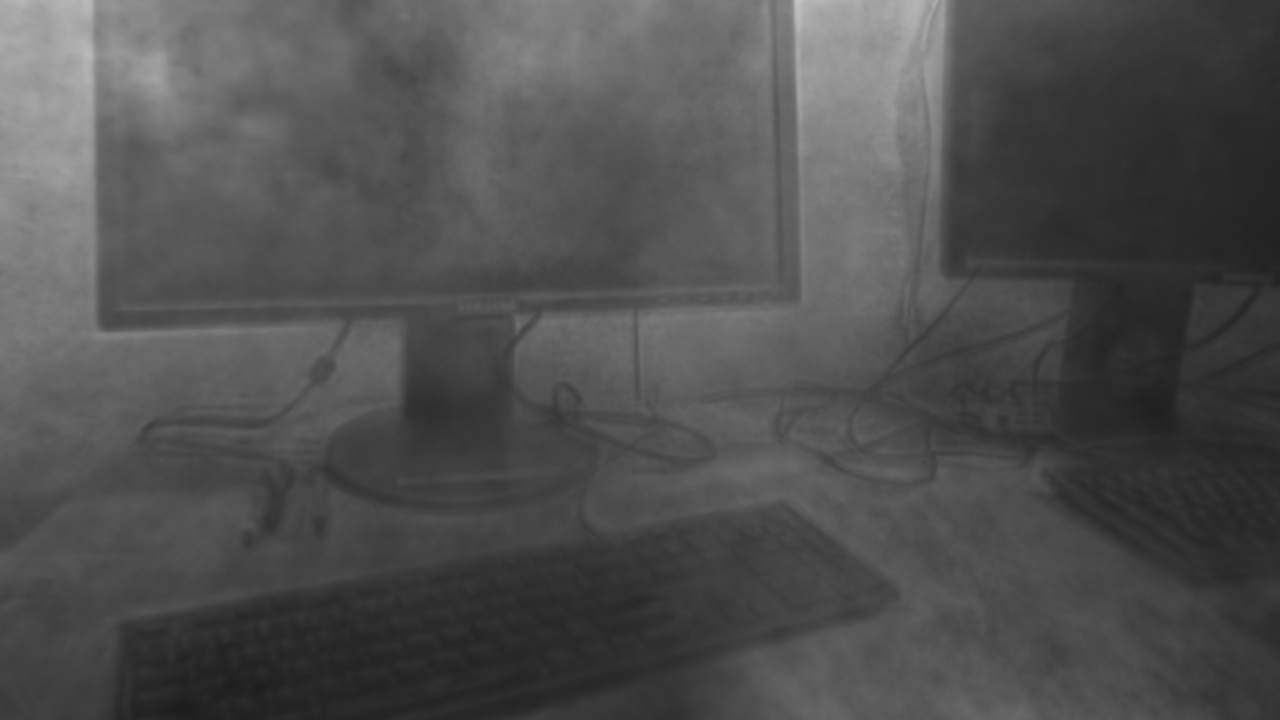} &
        \includegraphics[width=0.186\textwidth]{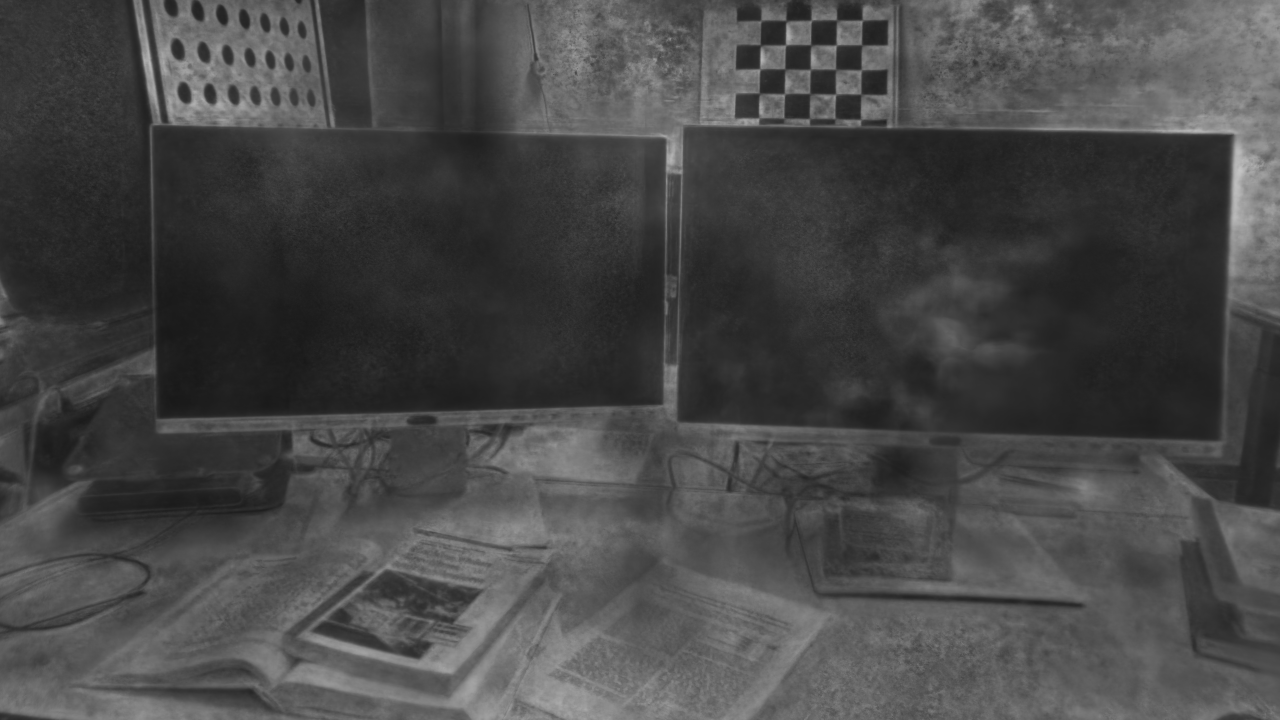} &
        \includegraphics[width=0.186\textwidth]{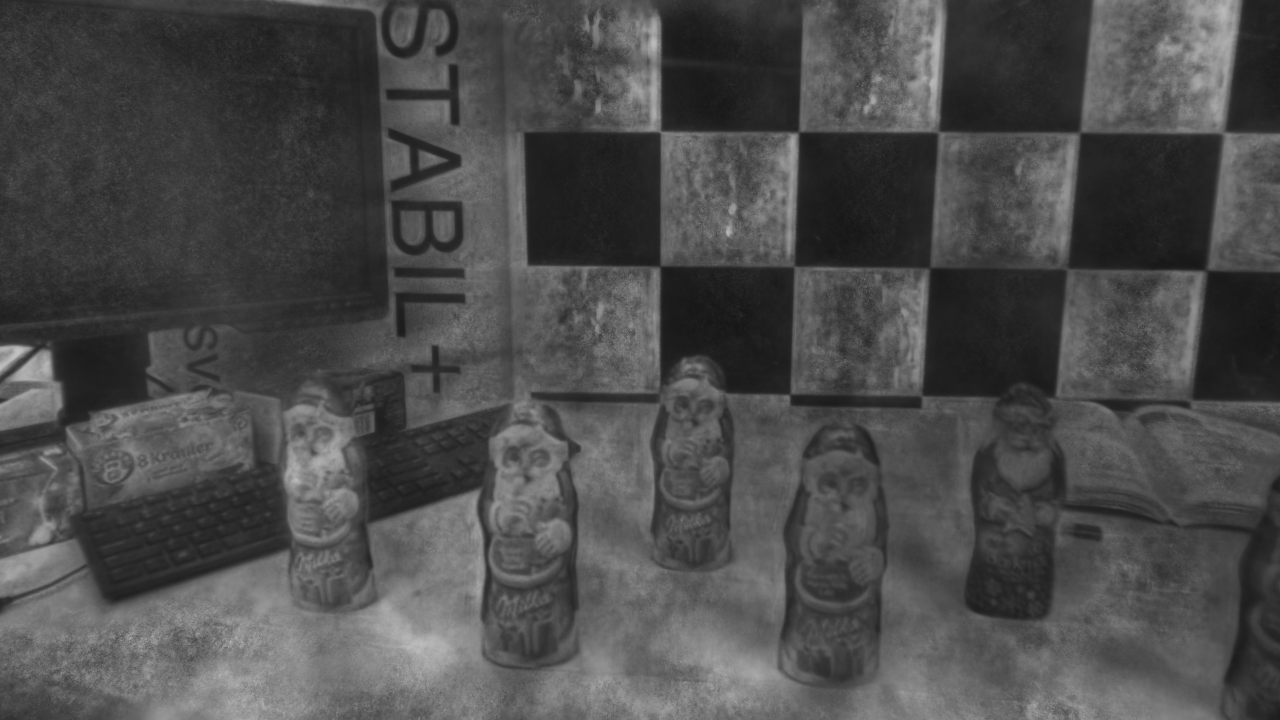} &
        \includegraphics[width=0.186\textwidth]{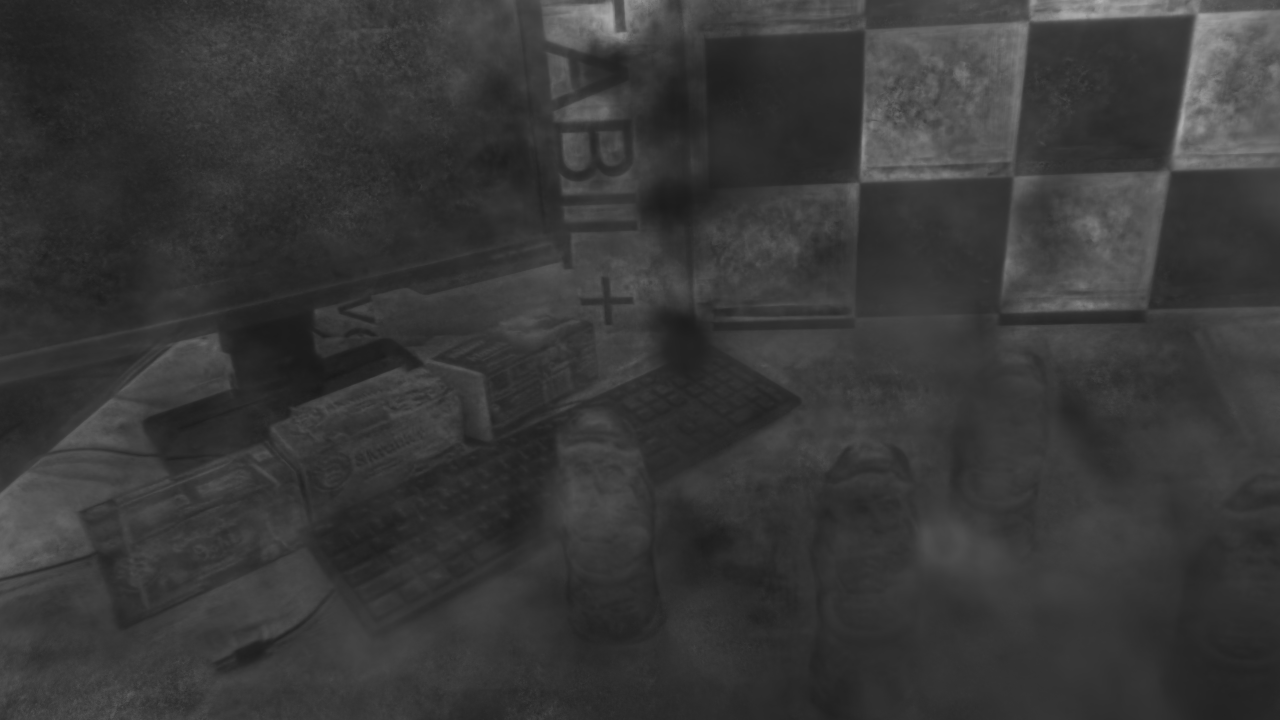} &
        \includegraphics[width=0.186\textwidth]{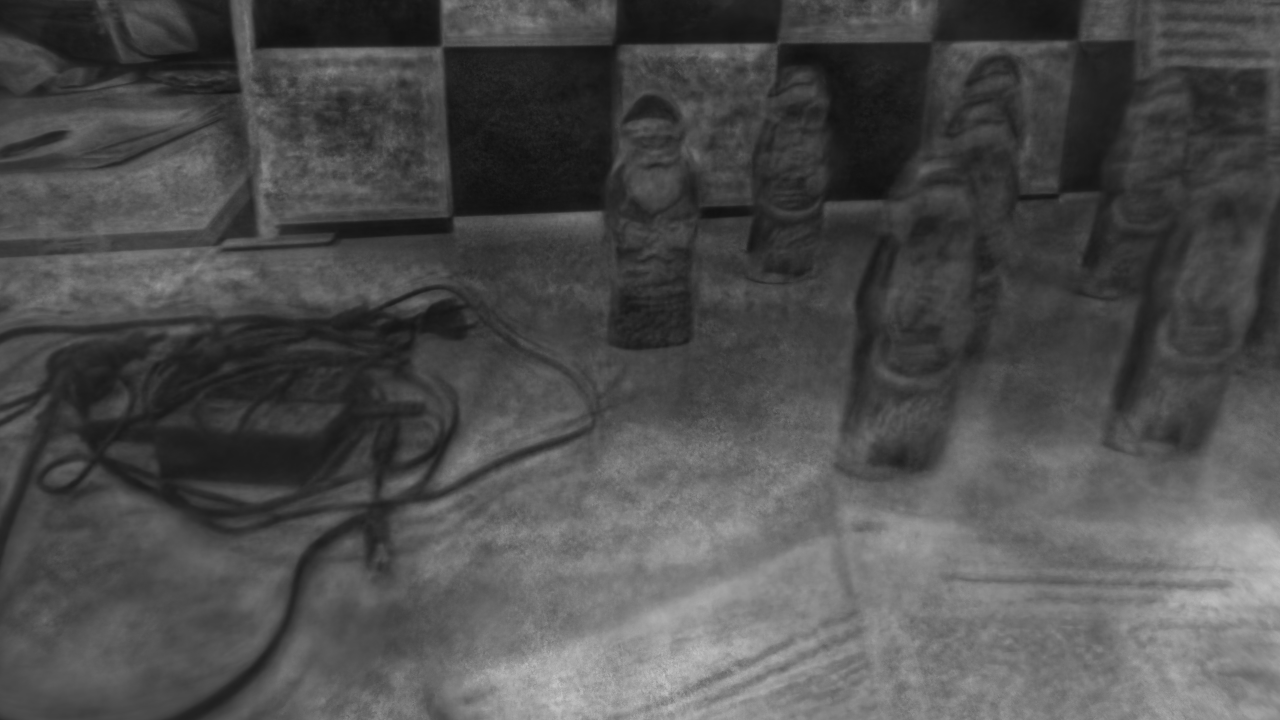} 
         \\

        \raisebox{18pt}{\rotatebox[origin=c]{90}{\tiny EventNeRF}}&
         \includegraphics[width=0.186\textwidth]{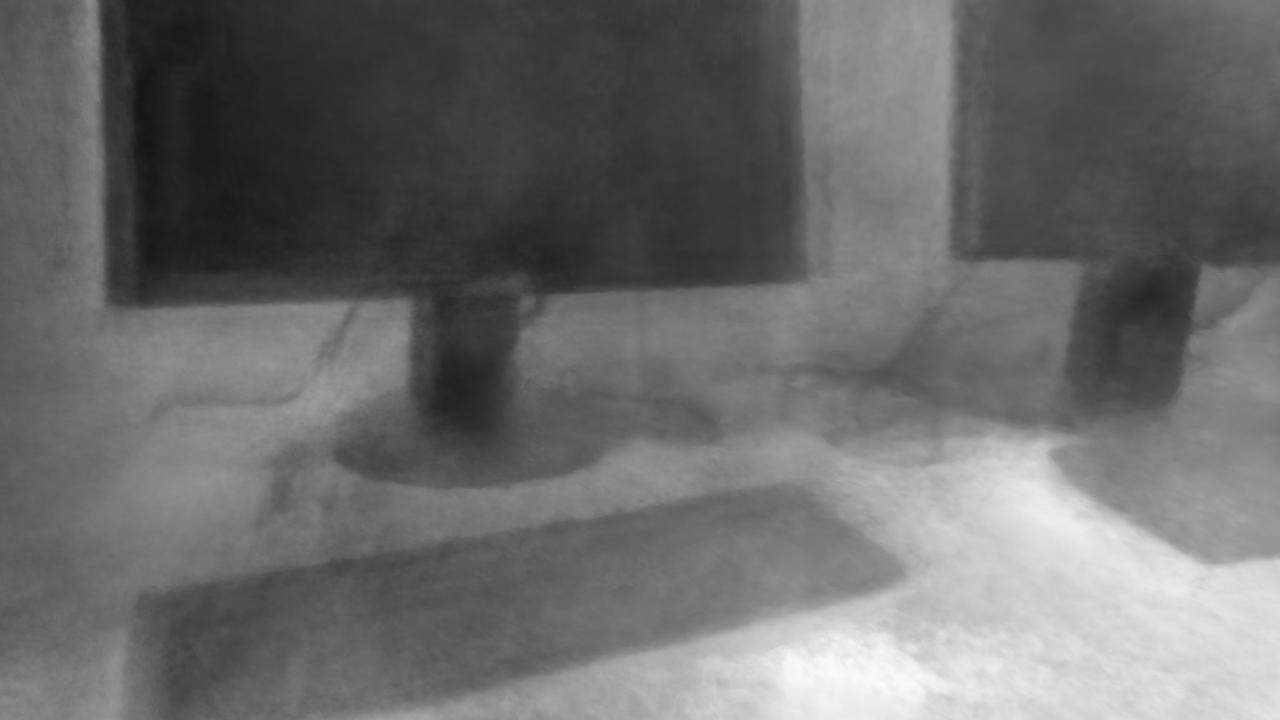} &
        \includegraphics[width=0.186\textwidth]{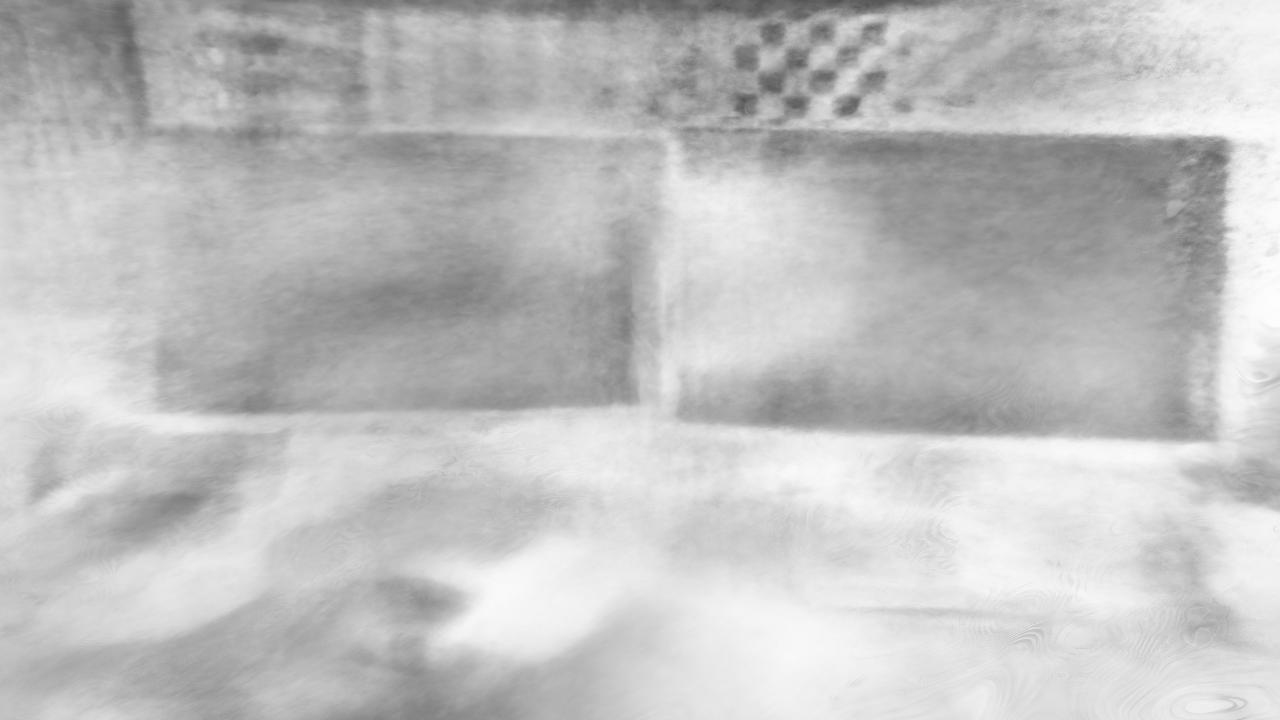} &
        \includegraphics[width=0.186\textwidth]{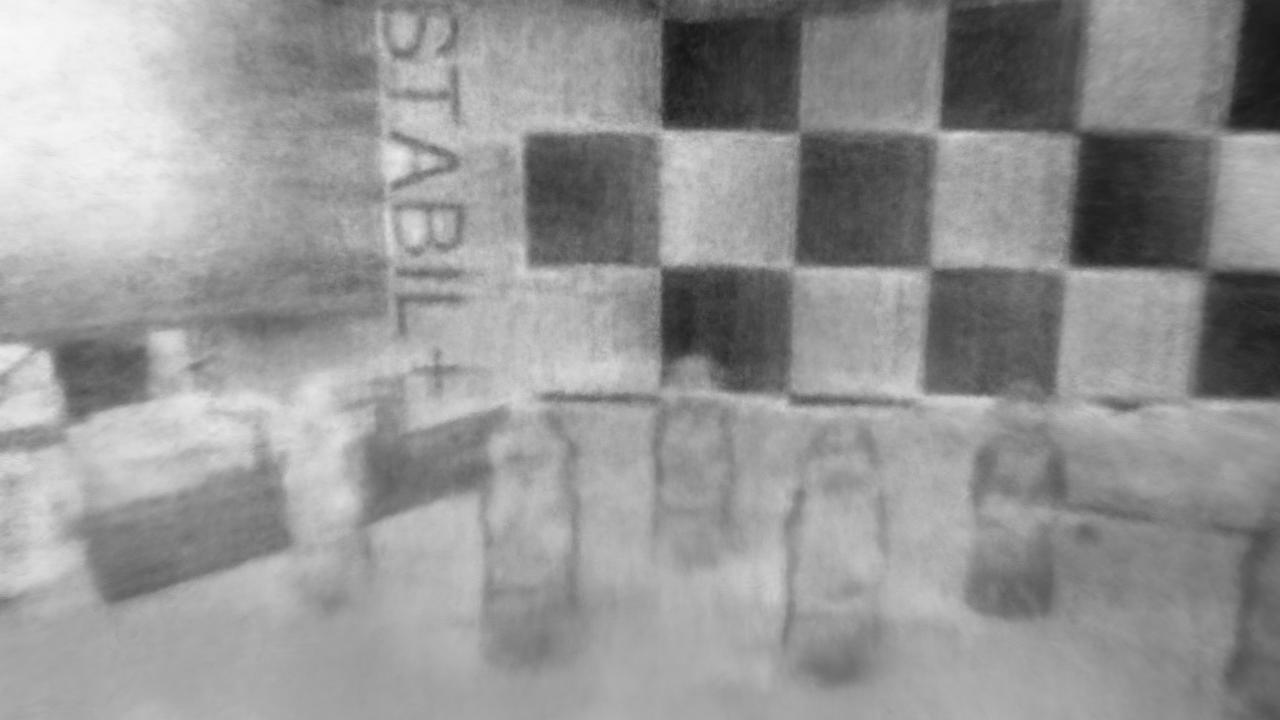} &
        \includegraphics[width=0.186\textwidth]{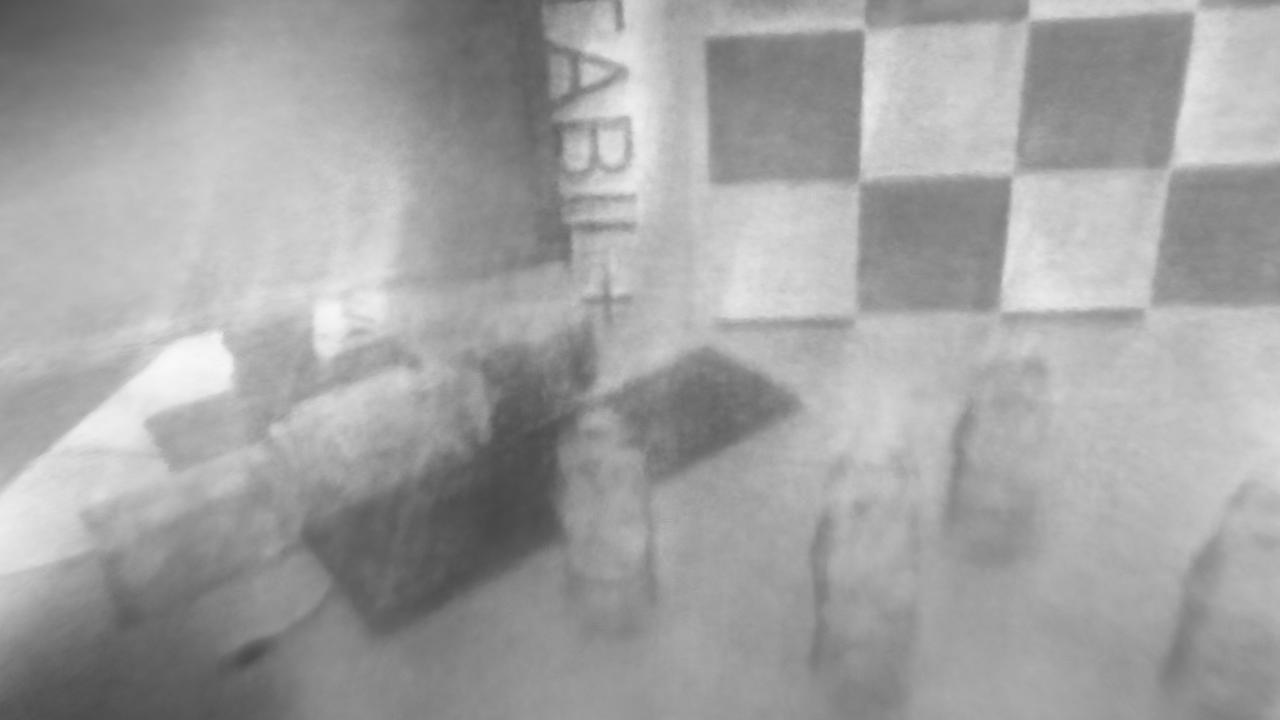} &
        \includegraphics[width=0.186\textwidth]{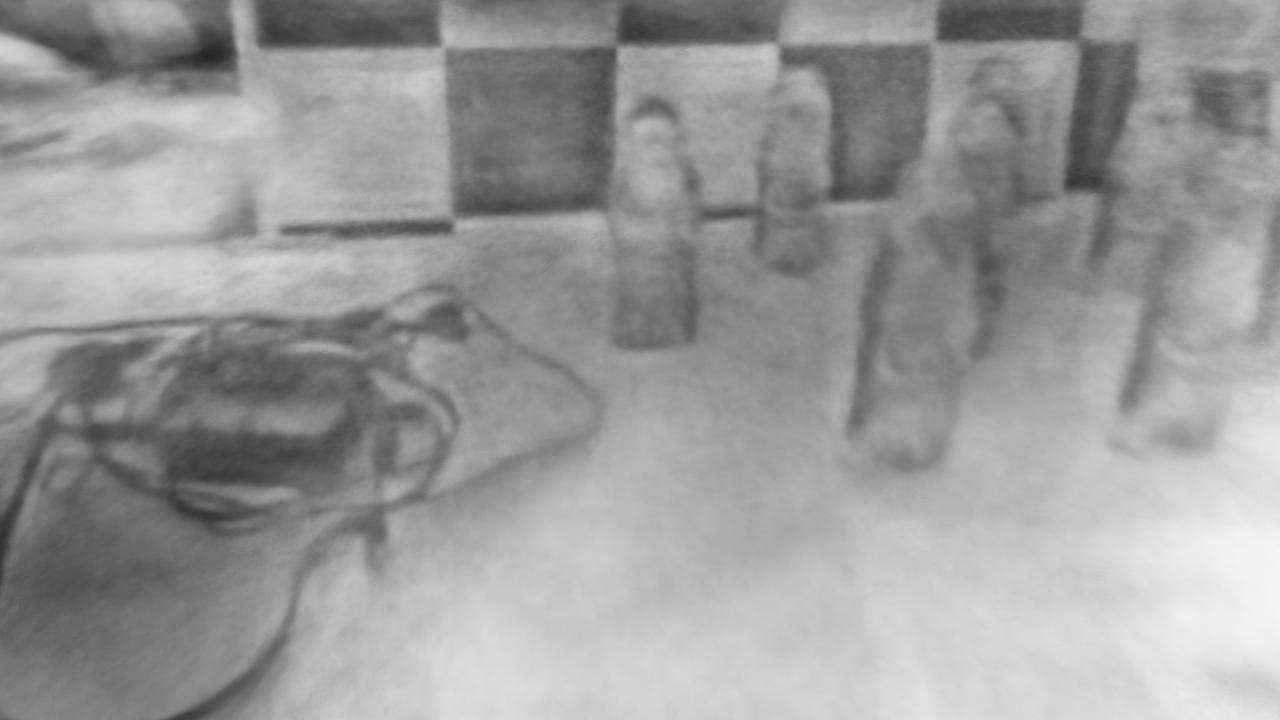} 
         \\

        \raisebox{18pt}{\rotatebox[origin=c]{90}{\tiny Robust e-NeRF}}&
         \includegraphics[width=0.186\textwidth]{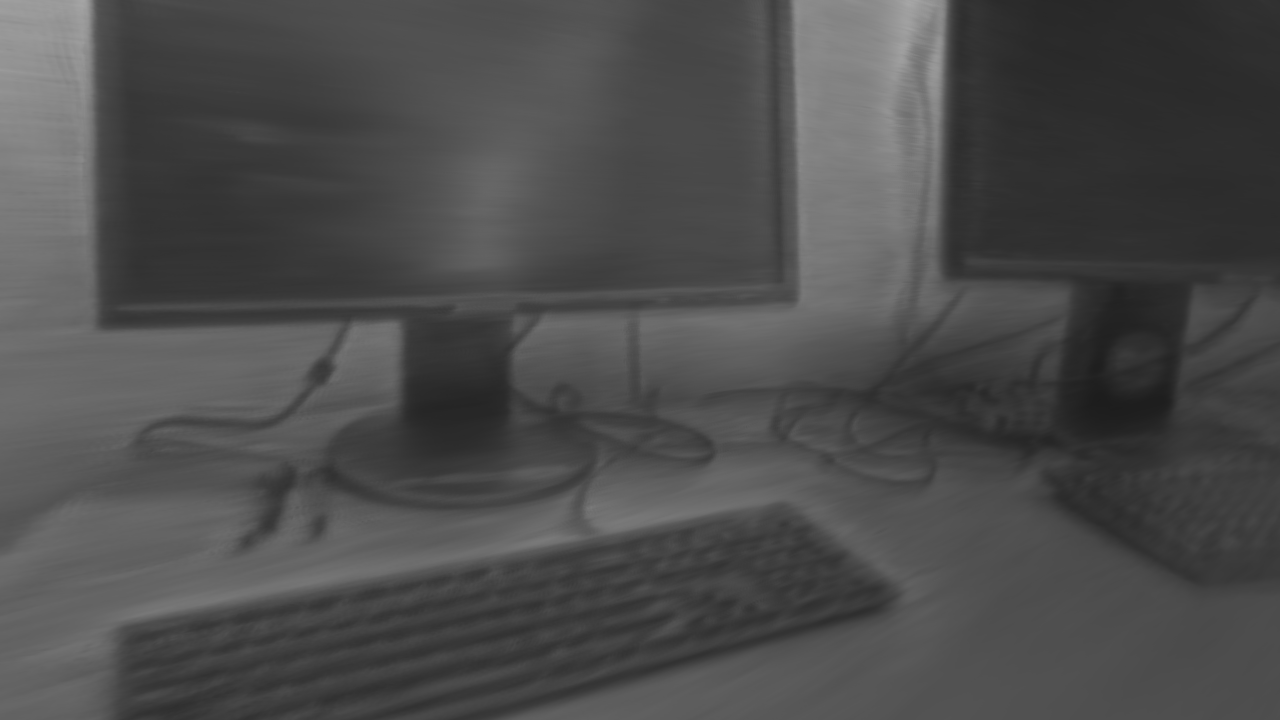} &
        \includegraphics[width=0.186\textwidth]{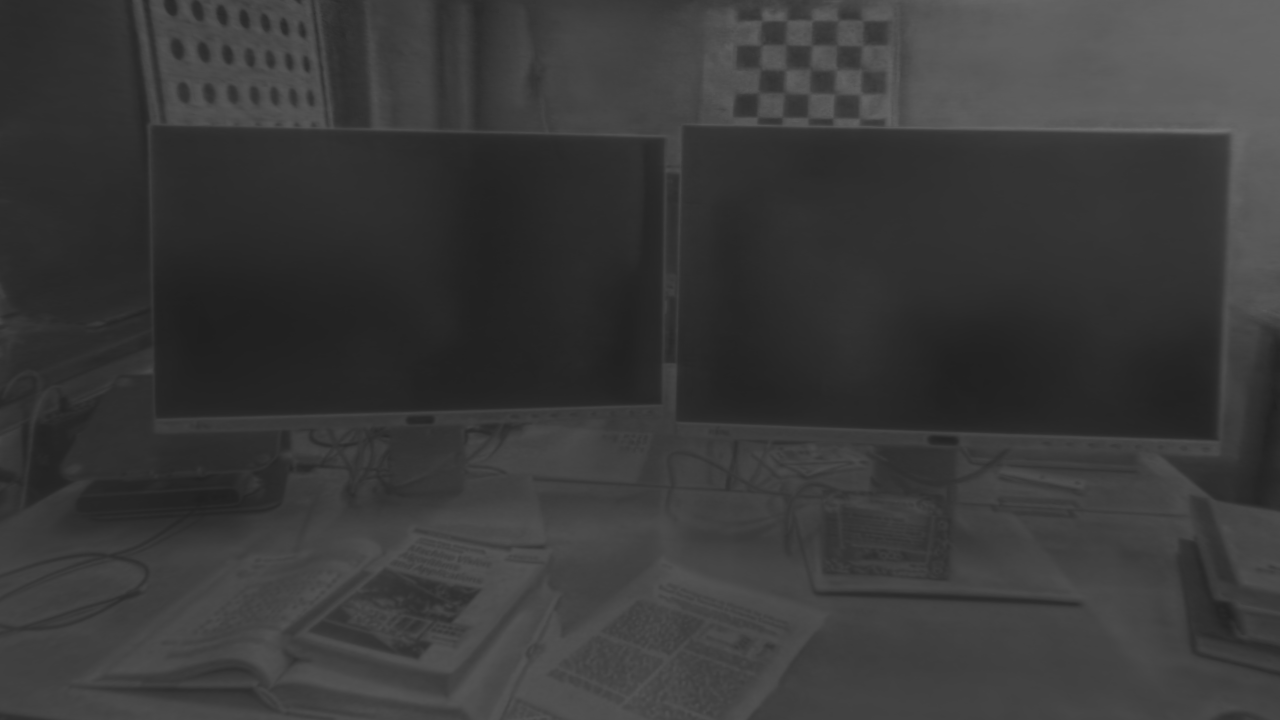} &
        \includegraphics[width=0.186\textwidth]{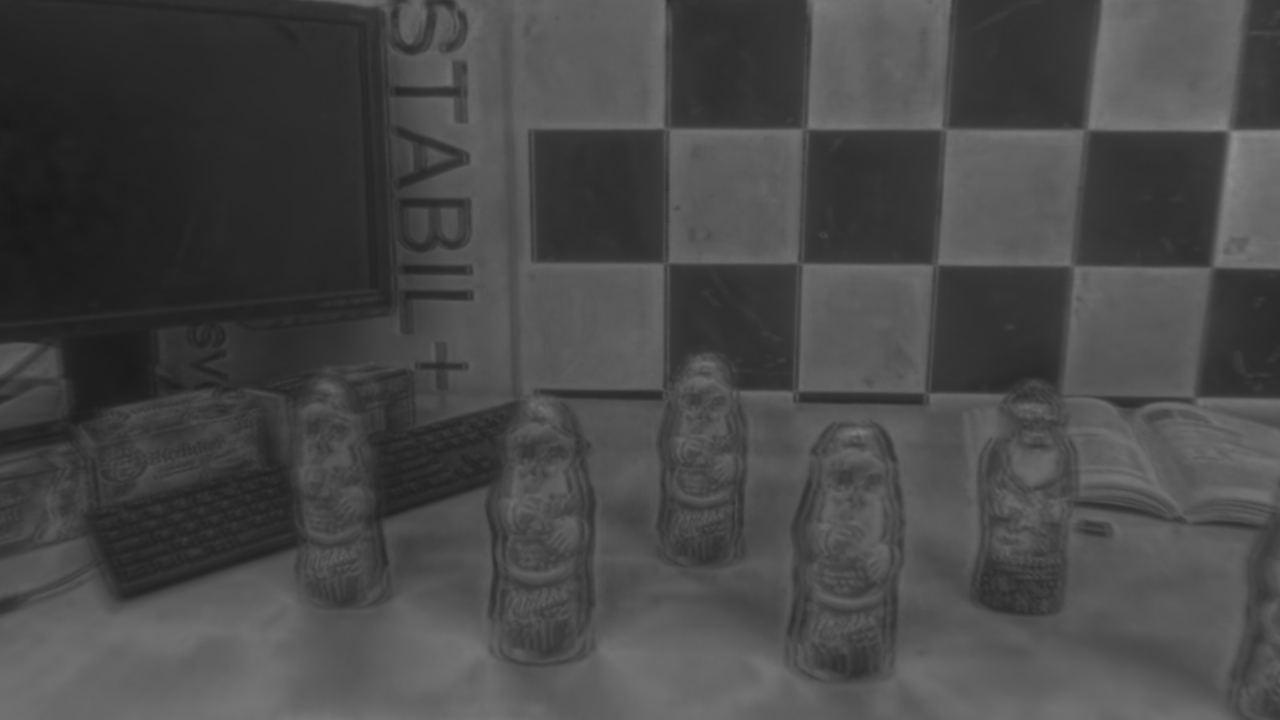} &
        \includegraphics[width=0.186\textwidth]{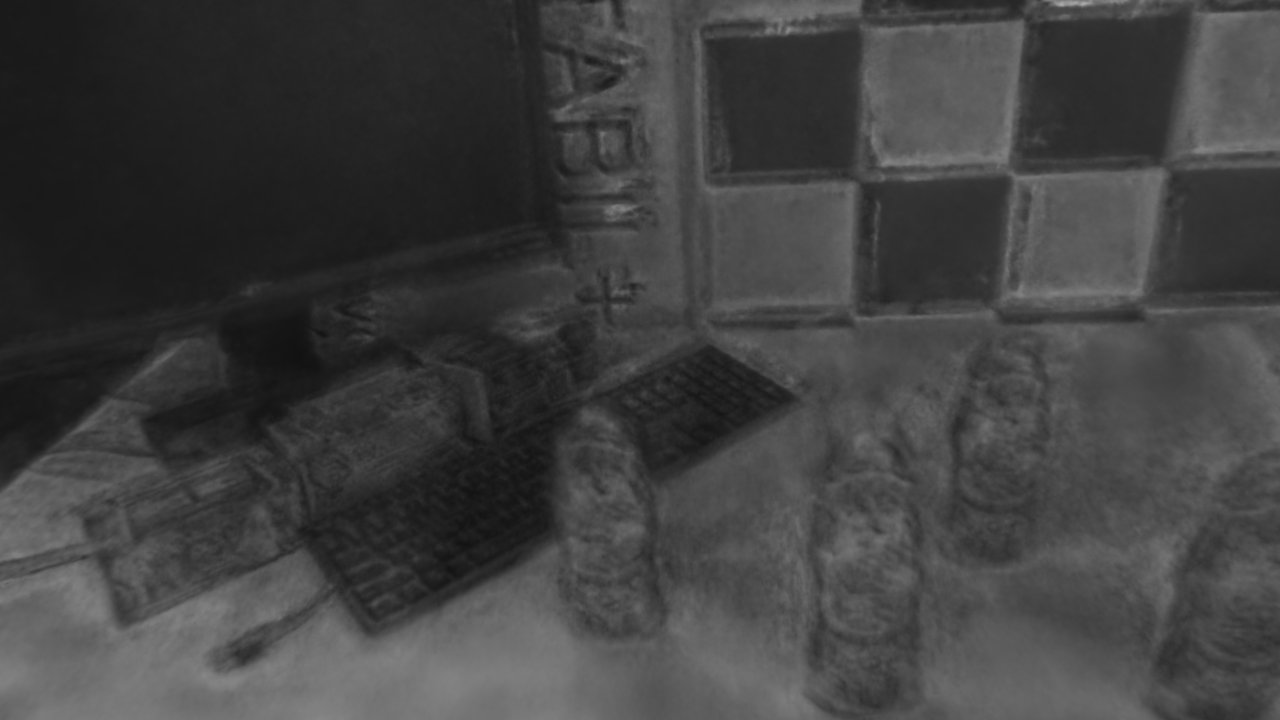} &
        \includegraphics[width=0.186\textwidth]{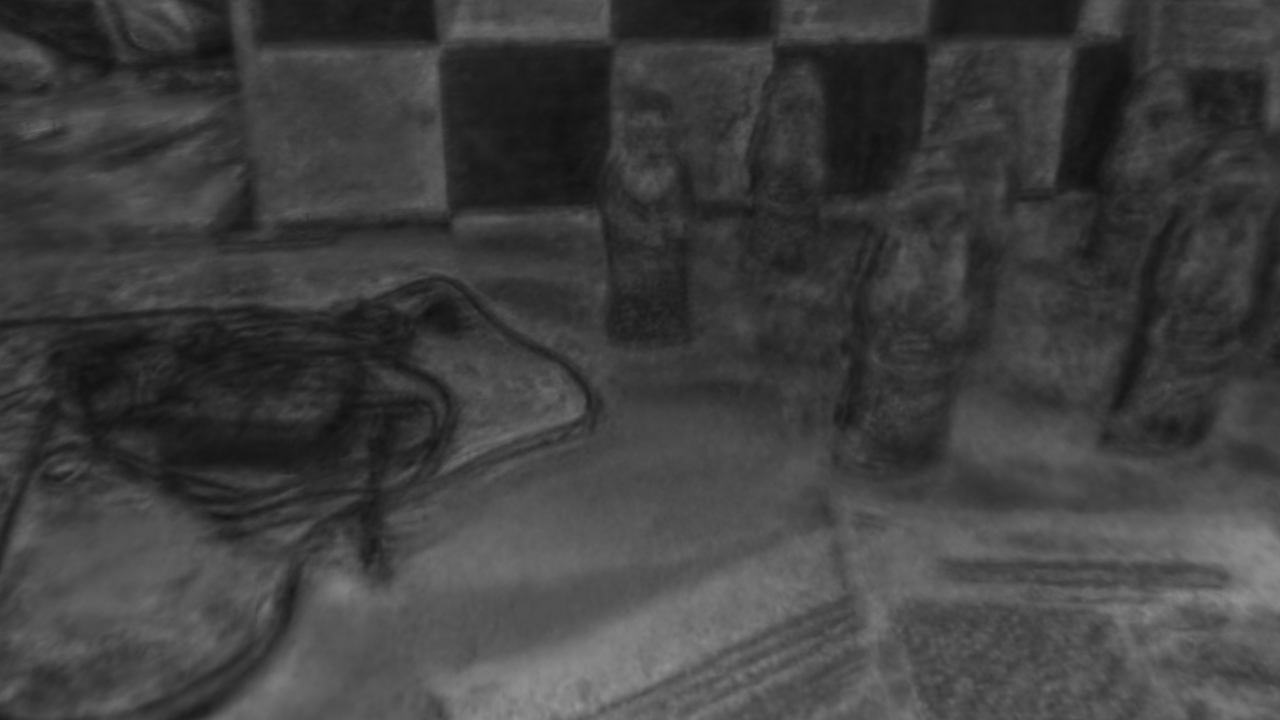} 
         \\

        \raisebox{18pt}{\rotatebox[origin=c]{90}{\tiny Ours}}&
         \includegraphics[width=0.186\textwidth]{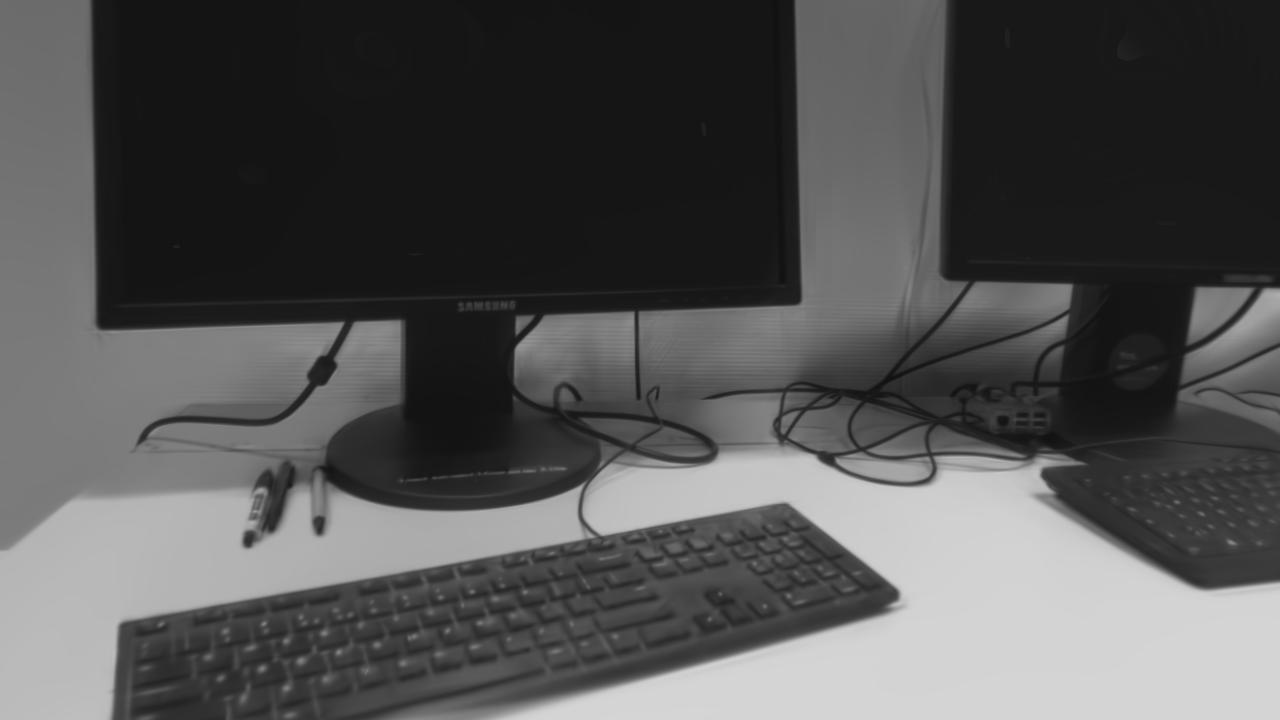} &
        \includegraphics[width=0.186\textwidth]{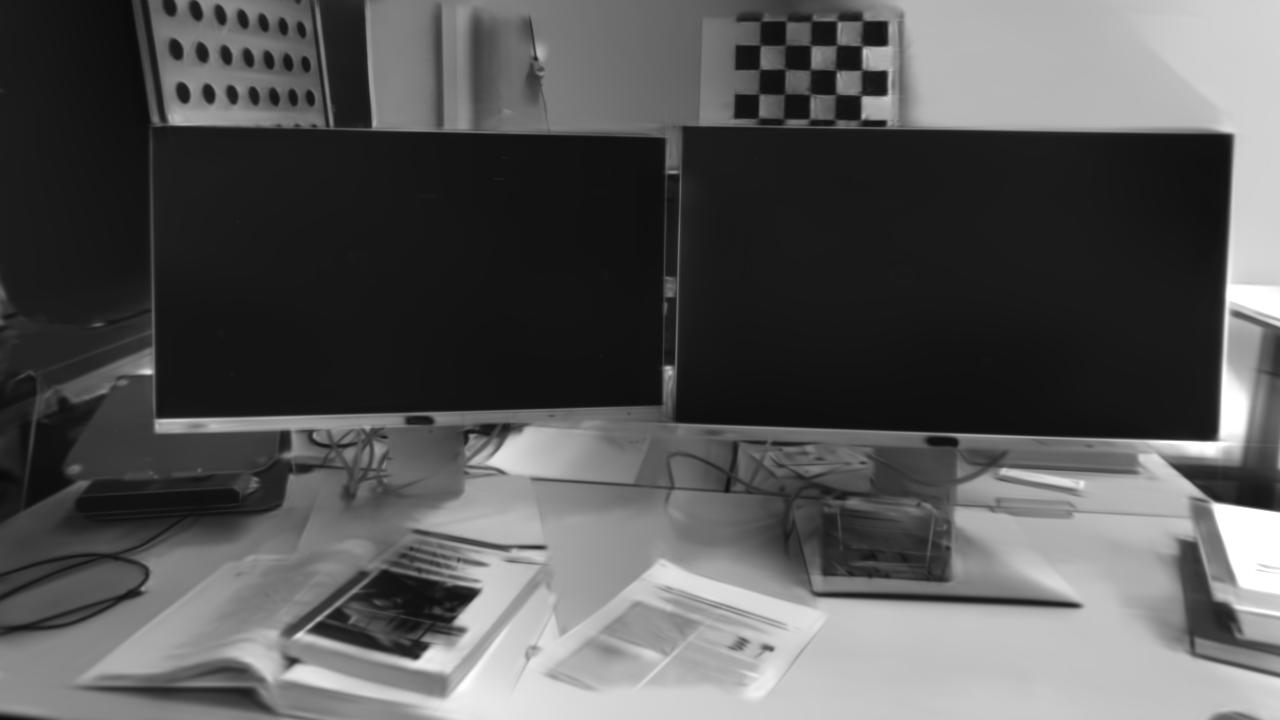} &
        \includegraphics[width=0.186\textwidth]{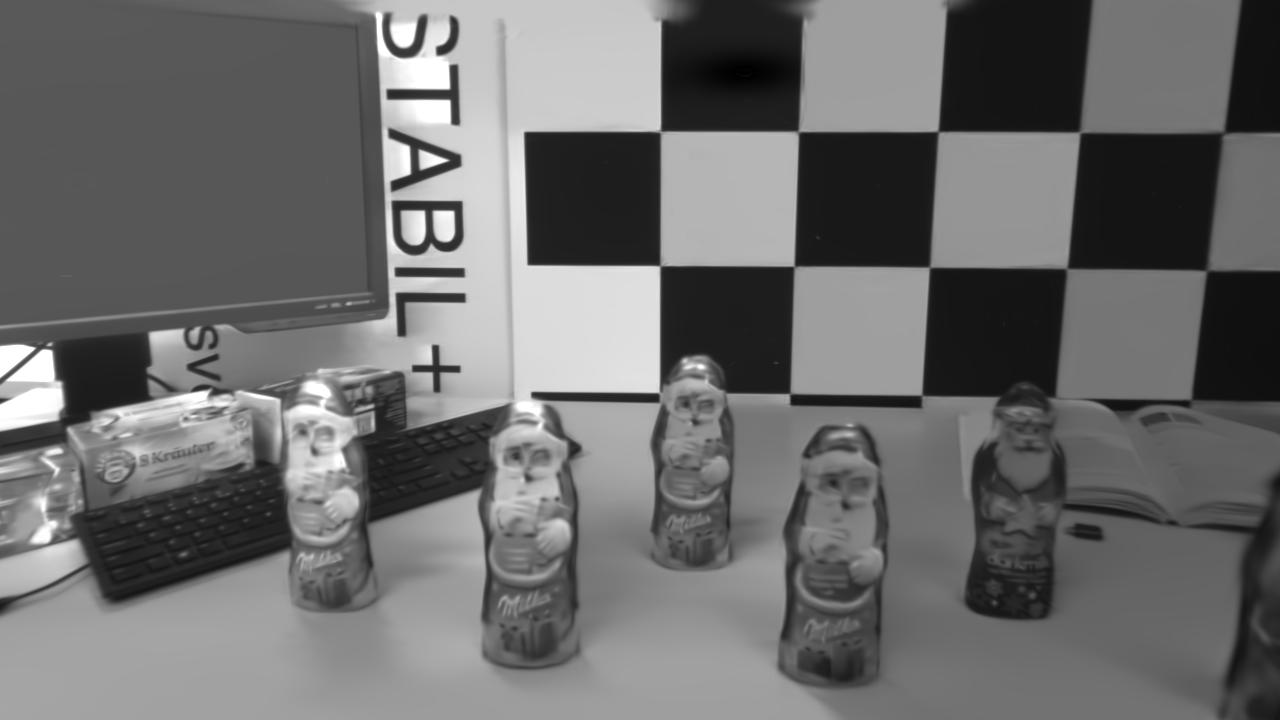} &
        \includegraphics[width=0.186\textwidth]{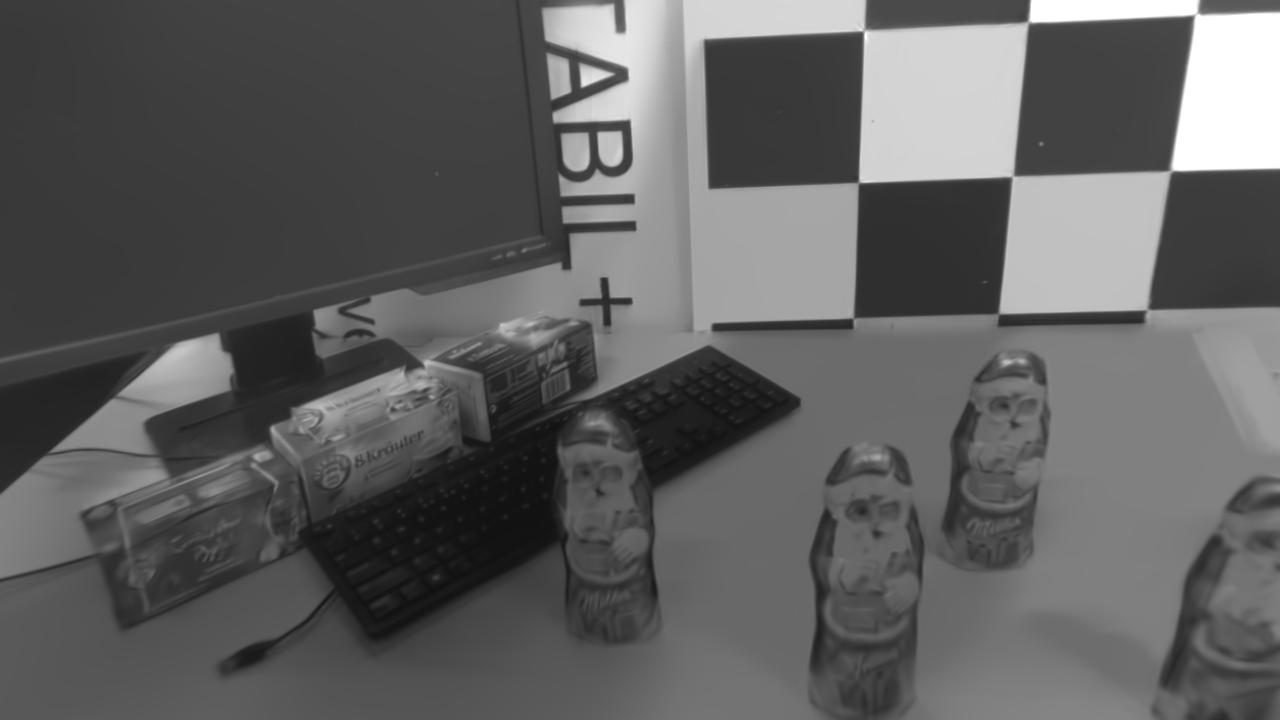} &
        \includegraphics[width=0.186\textwidth]{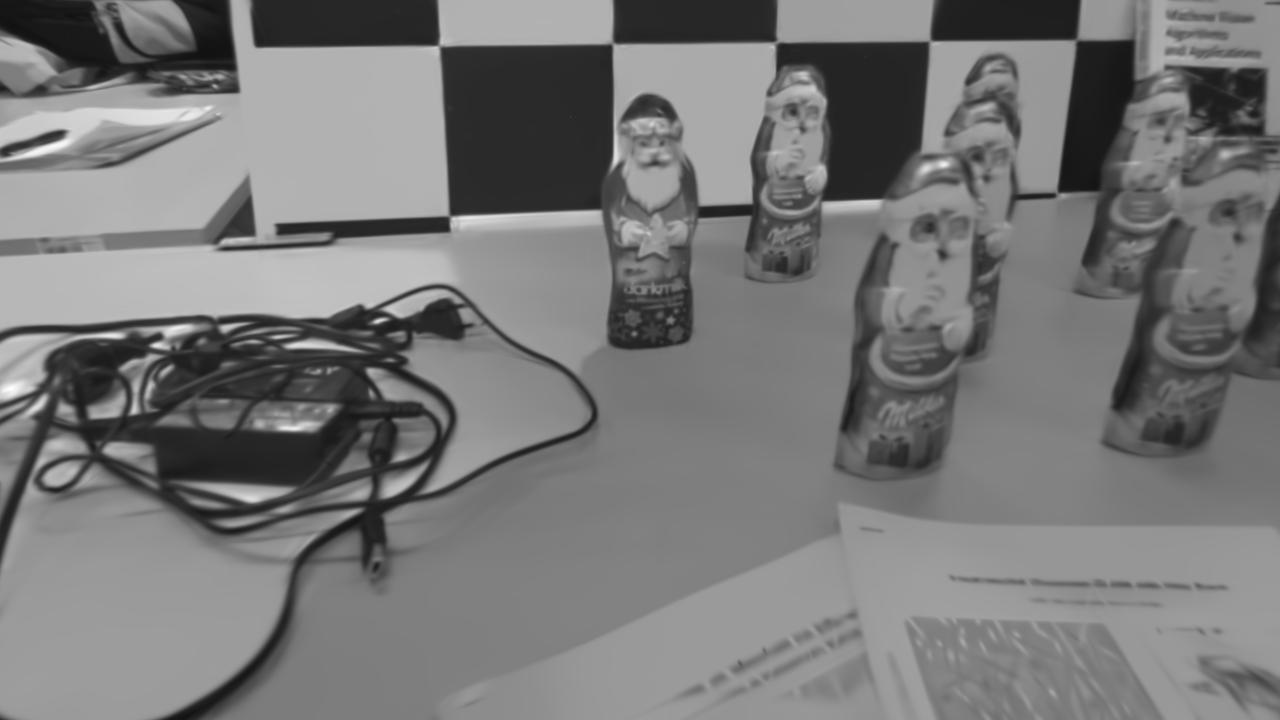} 
         \\

        \raisebox{18pt}{\rotatebox[origin=c]{90}{\tiny reference}}&
         \includegraphics[width=0.186\textwidth]{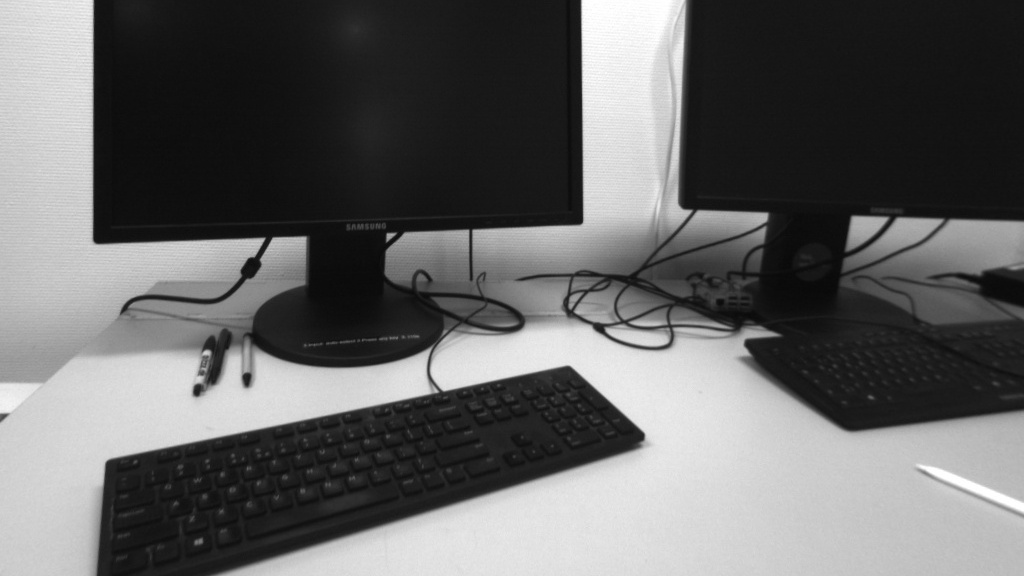} &
        \includegraphics[width=0.186\textwidth]{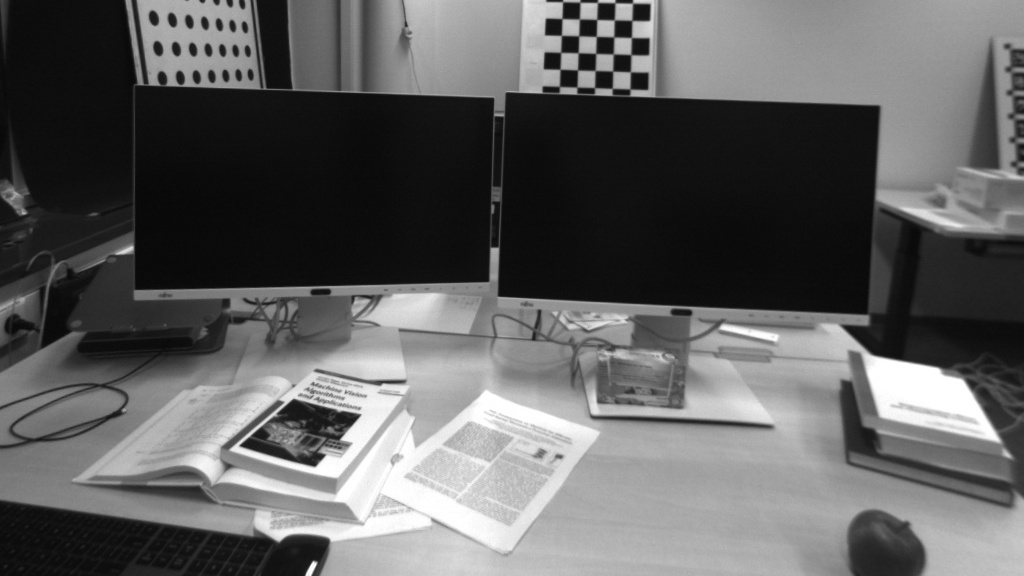} &
        \includegraphics[width=0.186\textwidth]{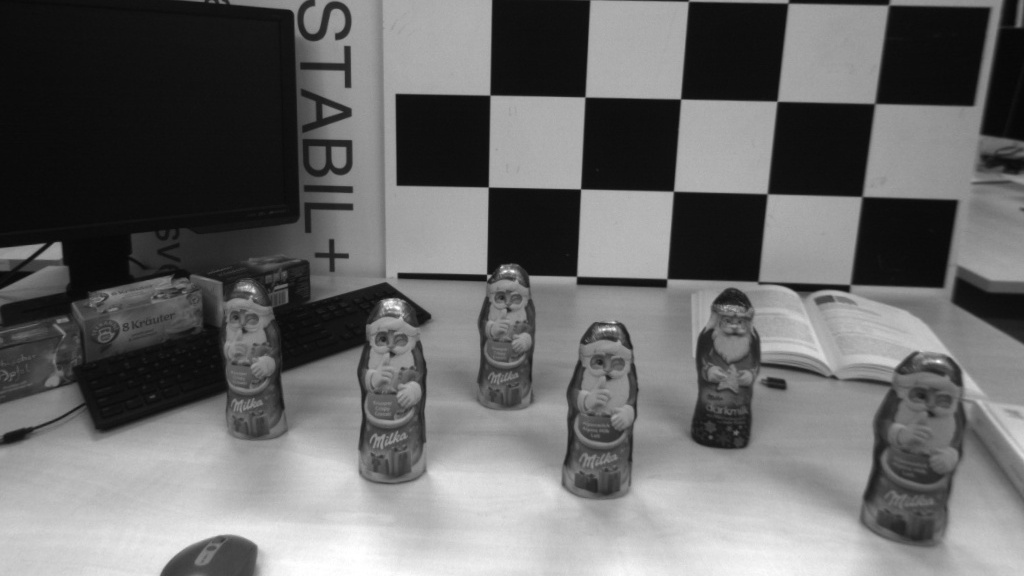} &
        \includegraphics[width=0.186\textwidth]{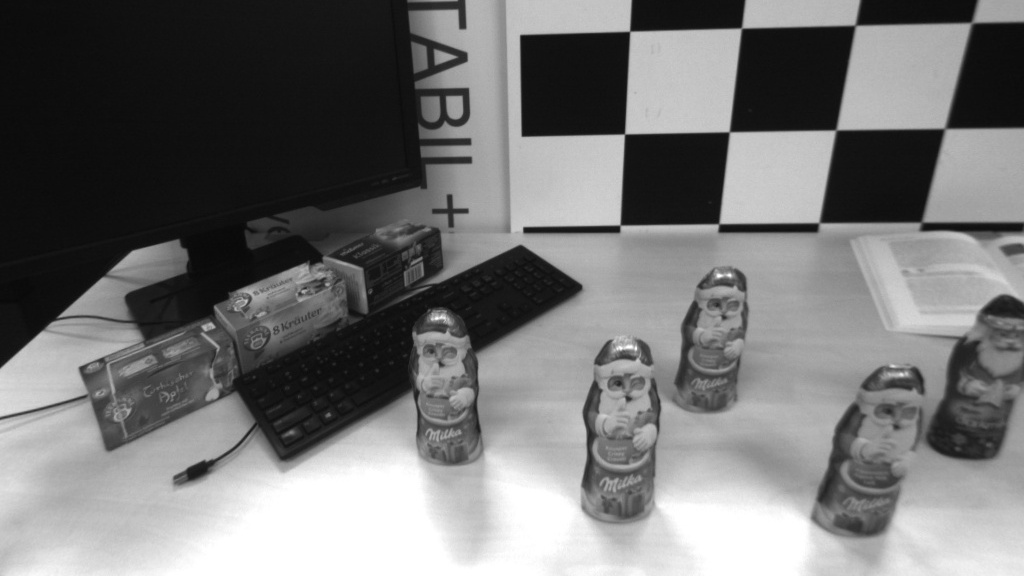} &
        \includegraphics[width=0.186\textwidth]{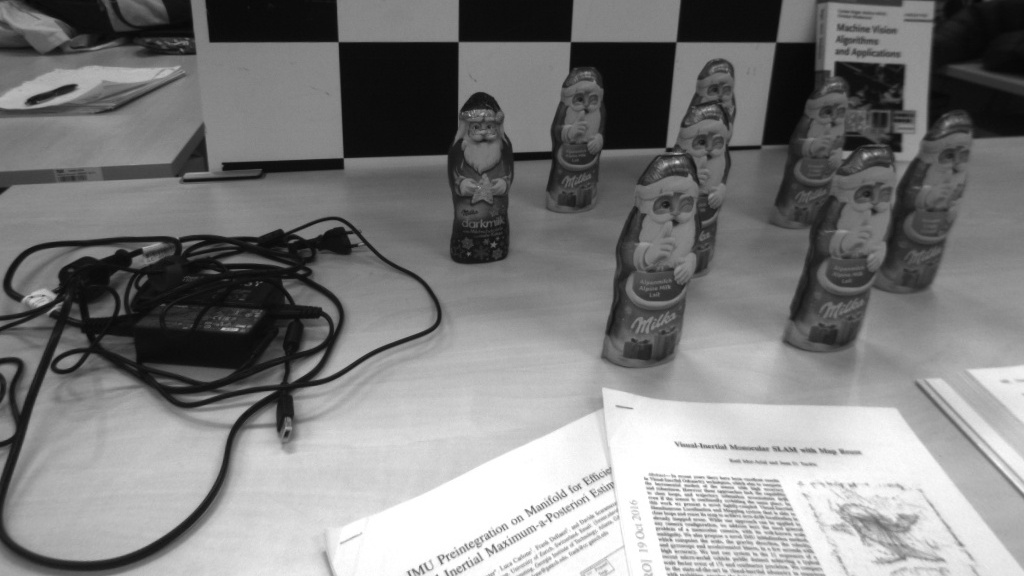} 
         \\
        \end{tabular}
    }
    \vspace{-5pt}
    \caption{Qualitative evaluation for novel view image synthesis on real dataset. It demonstrates that our method is able to render better images with fewer artifacts than event NeRF methods and two-stage methods. Note that there are no GT images aligned with the event camera, and we choose the closest images from the RGB camera and crop them to the same size as the rendered images for visual comparisons.}
    \label{pic:real_rendering}
    \vspace{-1.5em}
\end{figure*}

\begin{figure*}
    \centering
    \vspace{0.1em}
    \resizebox{\linewidth}{!}{
        \begin{tabular}{c c c c}
       
            \multirow{2}{*}{} & \textit{DEVO} & \textit{E2VID+COLMAP} & \textit{ours} \\
            
            \rotatebox{90}{\makecell[c]{office0}} & 
            \includegraphics[width=0.55\textwidth,valign=c]{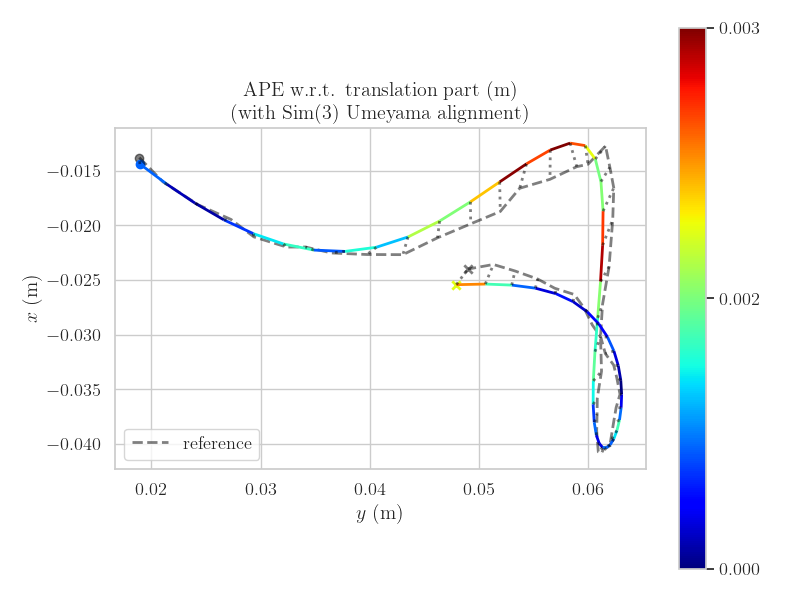} &
            \includegraphics[width=0.55\textwidth,valign=c]{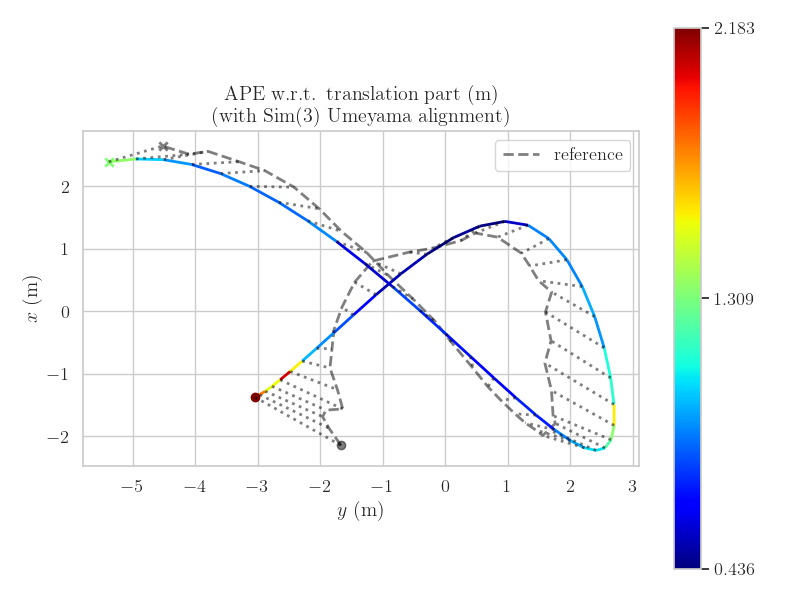} &
            \includegraphics[width=0.55\textwidth,valign=c]{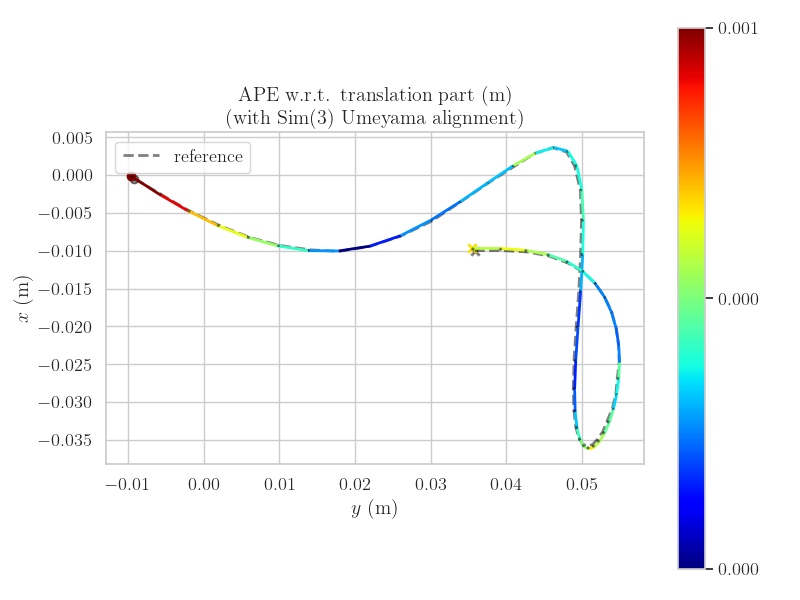} \\
            
            \rotatebox{90}{\makecell[c]{6dof}} & 
            \includegraphics[width=0.55\textwidth,valign=c]{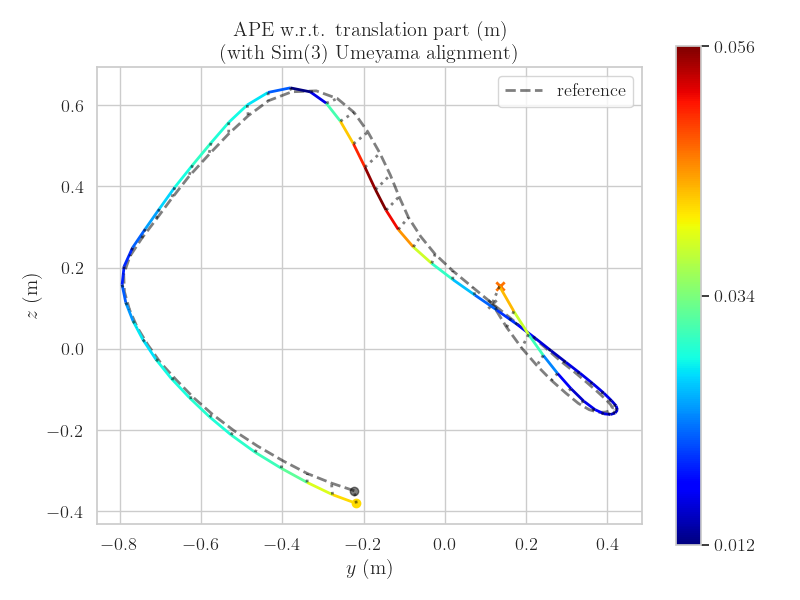} &
            \includegraphics[width=0.55\textwidth,valign=c]{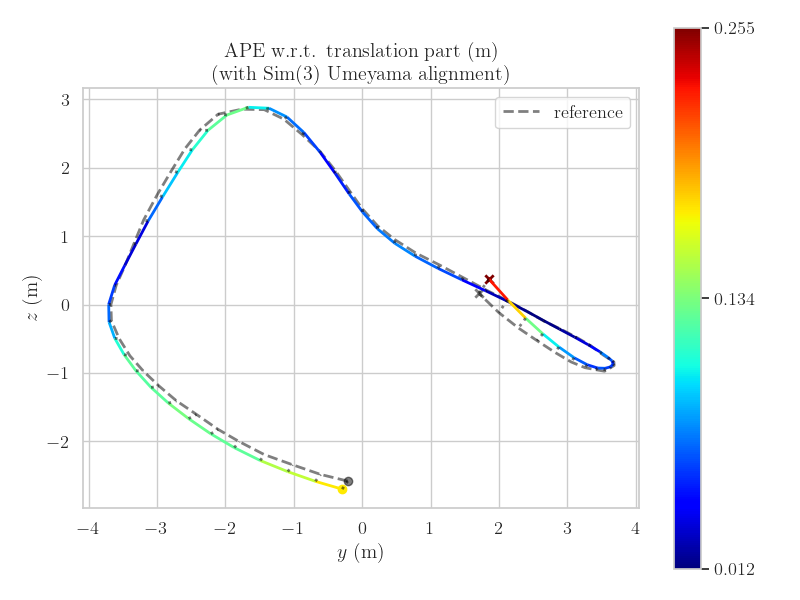} &
            \includegraphics[width=0.55\textwidth,valign=c]{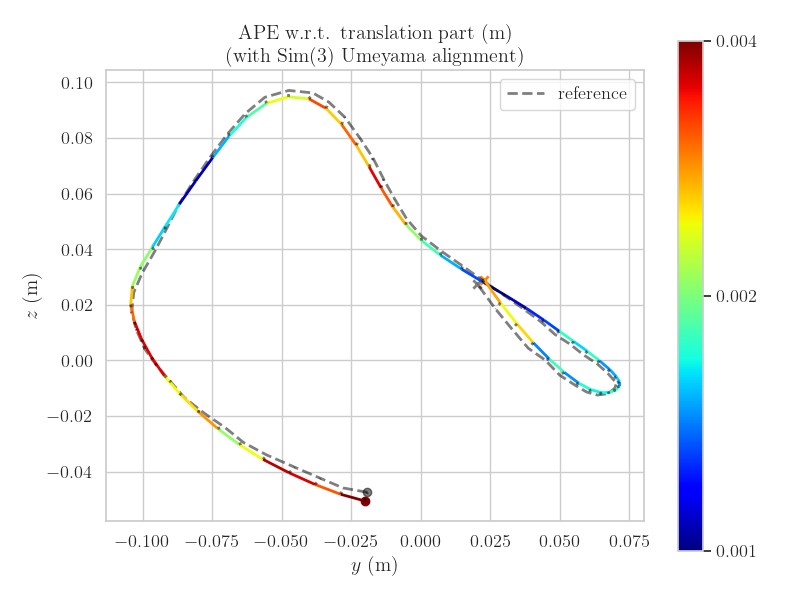} \\
        \end{tabular}
    }
    
    \caption{Representative visualization of ATE error mapped onto trajectories for the synthetic (office0) and real (6dof) datasets, generated by the EVO toolbox using the same ground truth poses, demonstrating the superior performance of our method in pose estimation.}
    \label{pic:traj}
    \vspace{-1.5em}
\end{figure*}

\subsection{Qualitative Evaluations.}

We evaluate our method against event NeRF methods and two-stage method qualitatively in terms of novel view image synthesis, both on synthetic and real data. The results are presented in both Fig. \ref{pic:syn_rendering} and Fig. \ref{pic:real_rendering}. It demonstrates that our method can deliver better novel view images, while event NeRF methods and two-stage method render images with additional artifacts. Compared to NeRF-based methods, our approach demonstrates the advantage of \textit{\methodname} by leveraging 3D Gaussian Splatting as the underlying scene representation. In contrast to two-stage method, our dense bundle adjustment optimizes both 3D Gaussian Splatting and camera pose using event data, whereas two-stage approaches tend to accumulate errors over time, as confirmed by the experimental results. We also provide representative visualization of ATE error mapped onto trajectories in Fig. \ref{pic:traj}, both on synthetic and real dataset. It demonstrates that \textit{\methodname} is able to recover more accurate motion trajectories.

\section{Conclusion}
We present the first pose-free 3D Gaussian splatting reconstruction model based on event camera, \ie \textit{\methodname}, and is useful for real world applications. We adopt the tracking and mapping paradigm in conventional SLAM pipeline to do incremental motion estimation and 3D scene reconstruction simultaneously. To handle the continuous and asynchronous characteristics of event stream, we exploit a continuous trajectory model to model the event data formation process. The experimental results on both synthetic and real datasets demonstrate the superior performance of \textit{\methodname} over prior state-of-the-art methods in terms of high-quality novel image synthesis and camera pose estimation. 


\PAR{Acknowledgements.}
This work was supported in part by NSFC under Grant 62202389, in part by a grant from the Westlake University-Muyuan Joint Research Institute, and in part by the Westlake Education Foundation.

\label{conclusion}

{
    \small
    \bibliographystyle{ieeenat_fullname}
    \bibliography{main}

\begin{thebibliography}{39}
\providecommand{\natexlab}[1]{#1}
\providecommand{\url}[1]{\texttt{#1}}
\expandafter\ifx\csname urlstyle\endcsname\relax
  \providecommand{\doi}[1]{doi: #1}\else
  \providecommand{\doi}{doi: \begingroup \urlstyle{rm}\Url}\fi

\bibitem[Deguchi et~al.(2024)Deguchi, Masuda, Nakabayashi, and Saito]{deguchi2024e2gs}
Hiroyuki Deguchi, Mana Masuda, Takuya Nakabayashi, and Hideo Saito.
\newblock E2gs: Event enhanced gaussian splatting.
\newblock In \emph{2024 IEEE International Conference on Image Processing (ICIP)}, pages 1676--1682. IEEE, 2024.

\bibitem[Fu et~al.(2023)Fu, Liu, Kulkarni, Kautz, Efros, and Wang]{fu2023colmap}
Yang Fu, Sifei Liu, Amey Kulkarni, Jan Kautz, Alexei~A Efros, and Xiaolong Wang.
\newblock Colmap-free 3d gaussian splatting.
\newblock \emph{arXiv preprint arXiv:2312.07504}, 2023.

\bibitem[Gallego et~al.(2018)Gallego, Rebecq, and Scaramuzza]{Gallego2018cvpr}
Guillermo Gallego, Henri Rebecq, and Davide Scaramuzza.
\newblock A unifying contrast maximization framework for event cameras, with applications to motion, depth and optical flow estimation.
\newblock In \emph{CVPR}, 2018.

\bibitem[Gehrig et~al.(2020)Gehrig, Gehrig, Hidalgo-Carri{\'o}, and Scaramuzza]{gehrig2020video}
Daniel Gehrig, Mathias Gehrig, Javier Hidalgo-Carri{\'o}, and Davide Scaramuzza.
\newblock Video to events: Recycling video datasets for event cameras.
\newblock In \emph{Proceedings of the IEEE/CVF Conference on Computer Vision and Pattern Recognition}, pages 3586--3595, 2020.

\bibitem[Grupp(2017)]{grupp2017evo}
Michael Grupp.
\newblock evo: Python package for the evaluation of odometry and slam.
\newblock \url{https://github.com/MichaelGrupp/evo}, 2017.

\bibitem[Hu et~al.(2024)Hu, Chen, Feng, Li, Yang, Bao, Zhang, and Cui]{hu2024cg}
Jiarui Hu, Xianhao Chen, Boyin Feng, Guanglin Li, Liangjing Yang, Hujun Bao, Guofeng Zhang, and Zhaopeng Cui.
\newblock Cg-slam: Efficient dense rgb-d slam in a consistent uncertainty-aware 3d gaussian field.
\newblock \emph{arXiv preprint arXiv:2403.16095}, 2024.

\bibitem[Huang et~al.(2024)Huang, Li, Cheng, and Yeung]{huang2023photo}
Huajian Huang, Longwei Li, Hui Cheng, and Sai-Kit Yeung.
\newblock Photo-slam: Real-time simultaneous localization and photorealistic mapping for monocular, stereo, and rgb-d cameras.
\newblock In \emph{Proceedings of the IEEE/CVF Conference on Computer Vision and Pattern Recognition}, 2024.

\bibitem[Hwang et~al.(2023)Hwang, Kim, and Kim]{hwang2023ev}
Inwoo Hwang, Junho Kim, and Young~Min Kim.
\newblock Ev-nerf: Event based neural radiance field.
\newblock In \emph{Proceedings of the IEEE/CVF Winter Conference on Applications of Computer Vision}, pages 837--847, 2023.

\bibitem[Ke et~al.(2024)Ke, Obukhov, Huang, Metzger, Daudt, and Schindler]{ke2023repurposing}
Bingxin Ke, Anton Obukhov, Shengyu Huang, Nando Metzger, Rodrigo~Caye Daudt, and Konrad Schindler.
\newblock Repurposing diffusion-based image generators for monocular depth estimation.
\newblock In \emph{Proceedings of the IEEE/CVF Conference on Computer Vision and Pattern Recognition (CVPR)}, 2024.

\bibitem[Keetha et~al.(2024)Keetha, Karhade, Jatavallabhula, Yang, Scherer, Ramanan, and Luiten]{keetha2024splatam}
Nikhil Keetha, Jay Karhade, Krishna~Murthy Jatavallabhula, Gengshan Yang, Sebastian Scherer, Deva Ramanan, and Jonathon Luiten.
\newblock Splatam: Splat, track \& map 3d gaussians for dense rgb-d slam.
\newblock In \emph{Proceedings of the IEEE/CVF Conference on Computer Vision and Pattern Recognition}, 2024.

\bibitem[Kerbl et~al.(2023)Kerbl, Kopanas, Leimk{\"u}hler, and Drettakis]{kerbl3Dgaussians}
Bernhard Kerbl, Georgios Kopanas, Thomas Leimk{\"u}hler, and George Drettakis.
\newblock 3d gaussian splatting for real-time radiance field rendering.
\newblock \emph{ACM Transactions on Graphics}, 42\penalty0 (4), 2023.

\bibitem[Kim et~al.(2016)Kim, Leutenegger, and Davison]{Kim2016eccv}
Hanme Kim, Stefan Leutenegger, and Andrew~J. Davison.
\newblock Real-time 3d reconstruction and 6-dof tracking with an event camera.
\newblock In \emph{ECCV}, 2016.

\bibitem[Klenk et~al.(2021)Klenk, Chui, Demmel, and Cremers]{klenk2021tum}
Simon Klenk, Jason Chui, Nikolaus Demmel, and Daniel Cremers.
\newblock Tum-vie: The tum stereo visual-inertial event dataset.
\newblock In \emph{2021 IEEE/RSJ International Conference on Intelligent Robots and Systems (IROS)}, pages 8601--8608. IEEE, 2021.

\bibitem[Klenk et~al.(2023)Klenk, Koestler, Scaramuzza, and Cremers]{klenk2023nerf}
Simon Klenk, Lukas Koestler, Davide Scaramuzza, and Daniel Cremers.
\newblock E-nerf: Neural radiance fields from a moving event camera.
\newblock \emph{IEEE Robotics and Automation Letters}, 8\penalty0 (3):\penalty0 1587--1594, 2023.

\bibitem[Klenk et~al.(2024)Klenk, Motzet, Koestler, and Cremers]{klenk2024deep}
Simon Klenk, Marvin Motzet, Lukas Koestler, and Daniel Cremers.
\newblock Deep event visual odometry.
\newblock In \emph{2024 International Conference on 3D Vision (3DV)}, pages 739--749. IEEE, 2024.

\bibitem[Li et~al.(2024)Li, Wan, Wang, Li, Zhou, and Liu]{li2024benerf}
Wenpu Li, Pian Wan, Peng Wang, Jinghang Li, Yi Zhou, and Peidong Liu.
\newblock Benerf: Neural radiance fields from a single blurry image and event stream.
\newblock In \emph{European Conference on Computer Vision (ECCV)}, 2024.

\bibitem[Low and Lee(2023)]{low2023_robust-e-nerf}
Weng~Fei Low and Gim~Hee Lee.
\newblock Robust e-nerf: Nerf from sparse and noisy events under non-uniform motion.
\newblock In \emph{Proceedings of the IEEE/CVF International Conference on Computer Vision (ICCV)}, 2023.

\bibitem[Matsuki et~al.(2024)Matsuki, Murai, Kelly, and Davison]{matsuki2024gaussian}
Hidenobu Matsuki, Riku Murai, Paul~HJ Kelly, and Andrew~J Davison.
\newblock Gaussian splatting slam.
\newblock In \emph{Proceedings of the IEEE/CVF Conference on Computer Vision and Pattern Recognition}, pages 18039--18048, 2024.

\bibitem[Mildenhall et~al.(2021)Mildenhall, Srinivasan, Tancik, Barron, Ramamoorthi, and Ng]{mildenhall2021nerf}
Ben Mildenhall, Pratul~P Srinivasan, Matthew Tancik, Jonathan~T Barron, Ravi Ramamoorthi, and Ren Ng.
\newblock Nerf: Representing scenes as neural radiance fields for view synthesis.
\newblock \emph{Communications of the ACM}, 65\penalty0 (1):\penalty0 99--106, 2021.

\bibitem[Mur-Artal and Tard{\'o}s(2017)]{mur2017orb}
Raul Mur-Artal and Juan~D Tard{\'o}s.
\newblock Orb-slam2: An open-source slam system for monocular, stereo, and rgb-d cameras.
\newblock \emph{IEEE Transactions on Robotics}, 33\penalty0 (5):\penalty0 1255--1262, 2017.

\bibitem[Niu et~al.(2025)Niu, Zhong, Lu, Shen, Gallego, and Zhou]{niu2025esvo2}
Junkai Niu, Sheng Zhong, Xiuyuan Lu, Shaojie Shen, Guillermo Gallego, and Yi Zhou.
\newblock Esvo2: Direct visual-inertial odometry with stereo event cameras.
\newblock \emph{IEEE Transactions on Robotics}, 2025.

\bibitem[Qu et~al.(2024)Qu, Yan, Wang, Yin, Xu, Zhao, and Li]{qu2023implicit}
Delin Qu, Chi Yan, Dong Wang, Jie Yin, Dan Xu, Bin Zhao, and Xuelong Li.
\newblock Implicit event-rgbd neural slam.
\newblock In \emph{Proceedings of the IEEE/CVF Conference on Computer Vision and Pattern Recognition}, 2024.

\bibitem[Rebecq et~al.(2017)Rebecq, Horstschaefer, Gallego, and Scaramuzza]{Rebecq2017ral}
Henri Rebecq, Timo Horstschaefer, Guillermo Gallego, and Davide Scaramuzza.
\newblock Evo: A geometric approach to event-based 6-dof parallel tracking and mapping in real time.
\newblock \emph{IEEE Robotics and Automation Letters}, 2017.

\bibitem[Rebecq et~al.(2019)Rebecq, Ranftl, Koltun, and Scaramuzza]{rebecq2019high}
Henri Rebecq, Ren{\'e} Ranftl, Vladlen Koltun, and Davide Scaramuzza.
\newblock High speed and high dynamic range video with an event camera.
\newblock \emph{IEEE transactions on pattern analysis and machine intelligence}, 43\penalty0 (6):\penalty0 1964--1980, 2019.

\bibitem[Rudnev et~al.(2023)Rudnev, Elgharib, Theobalt, and Golyanik]{rudnev2023eventnerf}
Viktor Rudnev, Mohamed Elgharib, Christian Theobalt, and Vladislav Golyanik.
\newblock Eventnerf: Neural radiance fields from a single colour event camera.
\newblock In \emph{Proceedings of the IEEE/CVF Conference on Computer Vision and Pattern Recognition}, pages 4992--5002, 2023.

\bibitem[Schonberger and Frahm(2016)]{colmap}
Johannes~L Schonberger and Jan-Michael Frahm.
\newblock {Structure-from-motion Revisited}.
\newblock In \emph{CVPR}, pages 4104--4113, 2016.

\bibitem[Sch\"{o}nberger and Frahm(2016)]{schoenberger2016sfm}
Johannes~Lutz Sch\"{o}nberger and Jan-Michael Frahm.
\newblock Structure-from-motion revisited.
\newblock In \emph{Conference on Computer Vision and Pattern Recognition (CVPR)}, 2016.

\bibitem[Straub et~al.(2019)Straub, Whelan, Ma, Chen, Wijmans, Green, Engel, Mur-Artal, Ren, Verma, Clarkson, Yan, Budge, Yan, Pan, Yon, Zou, Leon, Carter, Briales, Gillingham, Mueggler, Pesqueira, Savva, Batra, Strasdat, Nardi, Goesele, Lovegrove, and Newcombe]{replica19arxiv}
Julian Straub, Thomas Whelan, Lingni Ma, Yufan Chen, Erik Wijmans, Simon Green, Jakob~J. Engel, Raul Mur-Artal, Carl Ren, Shobhit Verma, Anton Clarkson, Mingfei Yan, Brian Budge, Yajie Yan, Xiaqing Pan, June Yon, Yuyang Zou, Kimberly Leon, Nigel Carter, Jesus Briales, Tyler Gillingham, Elias Mueggler, Luis Pesqueira, Manolis Savva, Dhruv Batra, Hauke~M. Strasdat, Renzo~De Nardi, Michael Goesele, Steven Lovegrove, and Richard Newcombe.
\newblock The {R}eplica dataset: A digital replica of indoor spaces.
\newblock \emph{arXiv preprint arXiv:1906.05797}, 2019.

\bibitem[Teed and Deng(2021)]{teed2021droid}
Zachary Teed and Jia Deng.
\newblock {DROID-SLAM: Deep Visual SLAM for Monocular, Stereo, and RGB-D Cameras}.
\newblock \emph{Advances in neural information processing systems}, 2021.

\bibitem[Teed et~al.(2023)Teed, Lipson, and Deng]{teed2023deep}
Zachary Teed, Lahav Lipson, and Jia Deng.
\newblock Deep patch visual odometry.
\newblock \emph{Advances in Neural Information Processing Systems}, 2023.

\bibitem[Wang et~al.(2024{\natexlab{a}})Wang, He, Zhang, Sun, Sun, and Xu]{wang2024evggs}
Jiaxu Wang, Junhao He, Ziyi Zhang, Mingyuan Sun, Jingkai Sun, and Renjing Xu.
\newblock Evggs: A collaborative learning framework for event-based generalizable gaussian splatting.
\newblock \emph{arXiv preprint arXiv:2405.14959}, 2024{\natexlab{a}}.

\bibitem[Wang et~al.(2023)Wang, Zhao, Ma, and Liu]{wang2023badnerf}
Peng Wang, Lingzhe Zhao, Ruijie Ma, and Peidong Liu.
\newblock {BAD-NeRF: Bundle Adjusted Deblur Neural Radiance Fields}.
\newblock In \emph{Proceedings of the IEEE/CVF Conference on Computer Vision and Pattern Recognition (CVPR)}, pages 4170--4179, 2023.

\bibitem[Wang et~al.(2024{\natexlab{b}})Wang, Zhao, Zhang, Zhao, and Liu]{wang2024mbaslam}
Peng Wang, Lingzhe Zhao, Yin Zhang, Shiyu Zhao, and Peidong Liu.
\newblock Mba-slam: Motion blur aware dense visual slam with radiance fields representation.
\newblock \emph{arXiv preprint arXiv:2411.08279}, 2024{\natexlab{b}}.

\bibitem[Wang et~al.(2004)Wang, Bovik, Sheikh, and Simoncelli]{wang2004image}
Zhou Wang, Alan~C Bovik, Hamid~R Sheikh, and Eero~P Simoncelli.
\newblock Image quality assessment: from error visibility to structural similarity.
\newblock \emph{IEEE transactions on image processing}, 13\penalty0 (4):\penalty0 600--612, 2004.

\bibitem[Xiong et~al.(2024)Xiong, Wu, He, Fermuller, Aloimonos, Huang, and Metzler]{xiong2024event3dgs}
Tianyi Xiong, Jiayi Wu, Botao He, Cornelia Fermuller, Yiannis Aloimonos, Heng Huang, and Christopher~A Metzler.
\newblock Event3dgs: Event-based 3d gaussian splatting for fast egomotion.
\newblock \emph{arXiv preprint arXiv:2406.02972}, 2024.

\bibitem[Yan et~al.(2024)Yan, Qu, Wang, Xu, Wang, Zhao, and Li]{yan2023gs}
Chi Yan, Delin Qu, Dong Wang, Dan Xu, Zhigang Wang, Bin Zhao, and Xuelong Li.
\newblock Gs-slam: Dense visual slam with 3d gaussian splatting.
\newblock In \emph{Proceedings of the IEEE/CVF Conference on Computer Vision and Pattern Recognition}, 2024.

\bibitem[Yugay et~al.(2023)Yugay, Li, Gevers, and Oswald]{yugay2023gaussian}
Vladimir Yugay, Yue Li, Theo Gevers, and Martin~R Oswald.
\newblock Gaussian-slam: Photo-realistic dense slam with gaussian splatting.
\newblock \emph{arXiv preprint arXiv:2312.10070}, 2023.

\bibitem[Zhao et~al.(2024)Zhao, Wang, and Liu]{zhao2024badgaussians}
Lingzhe Zhao, Peng Wang, and Peidong Liu.
\newblock Bad-gaussians: Bundle adjusted deblur gaussian splatting.
\newblock In \emph{European Conference on Computer Vision (ECCV)}, 2024.

\bibitem[Zhu et~al.(2022)Zhu, Peng, Larsson, Xu, Bao, Cui, Oswald, and Pollefeys]{zhu2022nice}
Zihan Zhu, Songyou Peng, Viktor Larsson, Weiwei Xu, Hujun Bao, Zhaopeng Cui, Martin~R Oswald, and Marc Pollefeys.
\newblock Nice-slam: Neural implicit scalable encoding for slam.
\newblock In \emph{Proceedings of the IEEE/CVF Conference on Computer Vision and Pattern Recognition}, pages 12786--12796, 2022.

\end{thebibliography}
}

\clearpage
\setcounter{page}{1}
\maketitlesupplementary

\begin{figure}[h]
    \centering
    \includegraphics[width=1.0\linewidth]{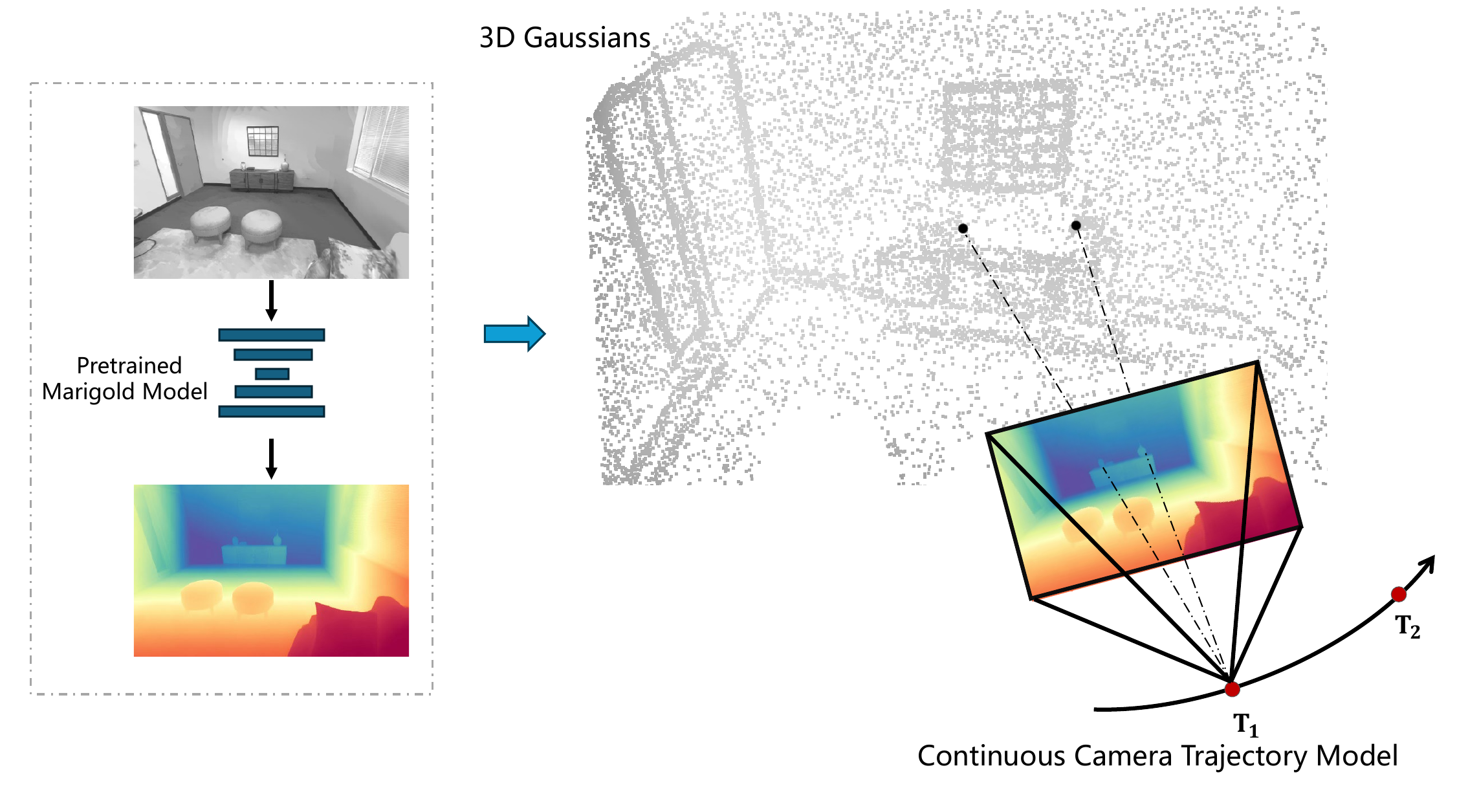}
    \vspace{-1.8em}
    \caption{{The re-initialization process of \textit{\methodname}}.}
    \label{pic:re-initialization}
    \vspace{-1.8em}
\end{figure}

\section{More Details about Re-initialization}
\label{sec:re-in}
The re-initialization process is illustrated in Fig. \ref{pic:re-initialization}. After the first-time initialization, we can render a brightness image from 3D-GS at pose $\bT_1$, where $\bT_1$ represents the camera pose at the end of the first event chunk. To improve the 3D structure of 3D-GS, we use a monocular depth estimation network \cite{ke2023repurposing} to predict a dense depth map from the rendered brightness image. This depth map is then used to re-initialize the centers of the 3D Gaussians by unprojecting the pixel depths at camera pose $\bT_1$, as illustrated in Fig. \ref{pic:re-initialization}. After re-centering the 3D Gaussians, we perform the initialization process again to achieve both accurate 3D structure and exceptional brightness image rendering performance.

\textbf{\section{Comparison with Gaussian-based Event Methods}}
To further evaluate our method, we conducted additional comparisons against state-of-the-art Gaussian-based event approaches. Since Event3DGS \cite{xiong2024event3dgs} had not been open-sourced, we chose to compare against E2GS\cite{deguchi2024e2gs} and EvGGS\cite{wang2024evggs}. In particular, we removed the supervision of blurred image in E2GS and exploited the pretrained weight of EvGGS for comparisons. As shown in Table \ref{tab:nvs_suppl}, our method still outperforms those two baselines event though they used ground truth poses. Since EvGGS is a generalizable method based on a feed-forward network, it has limited generalization capability on unseen dataset. 

\begin{figure*}
    \centering
    \vspace{0.3em}
    \addtolength{\tabcolsep}{-6.5pt}
    \footnotesize{
        \setlength{\tabcolsep}{1pt} 
        \begin{tabular}{p{8.2pt}cccccccc}
            & room0 & room2 & office0 & office2 & office3  \\

        \raisebox{24pt}{\rotatebox[origin=c]{90}{\tiny SplaTAM }}&
         \includegraphics[width=0.186\textwidth]{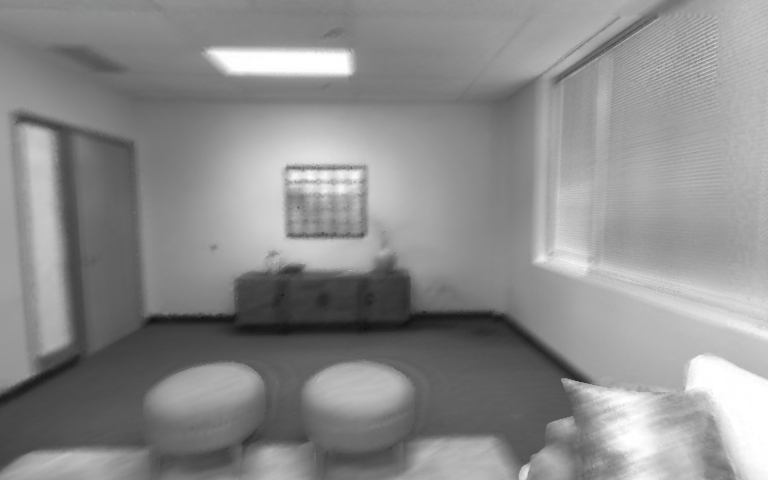} &
        \includegraphics[width=0.186\textwidth]{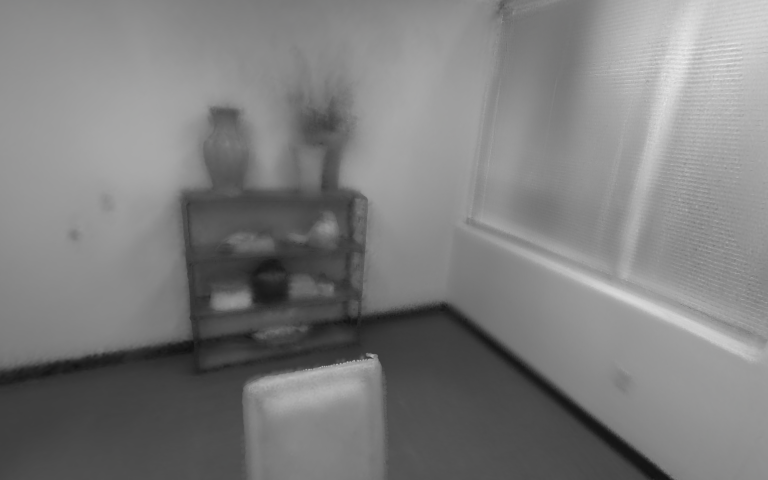} &
        \includegraphics[width=0.186\textwidth]{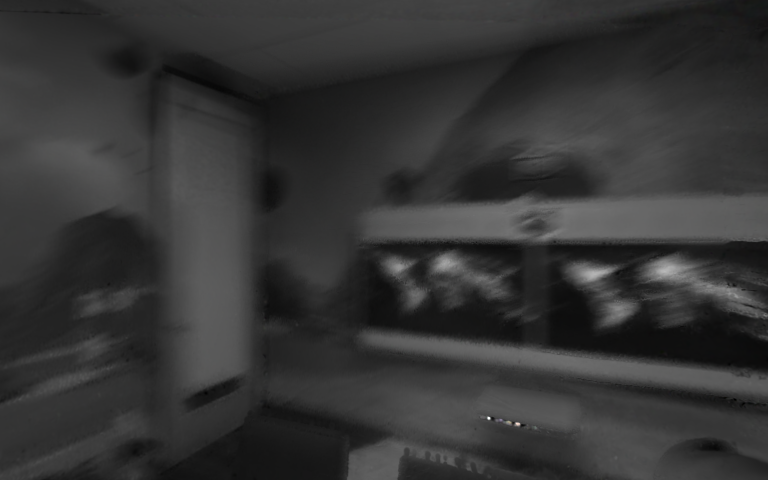} &
        \includegraphics[width=0.186\textwidth]{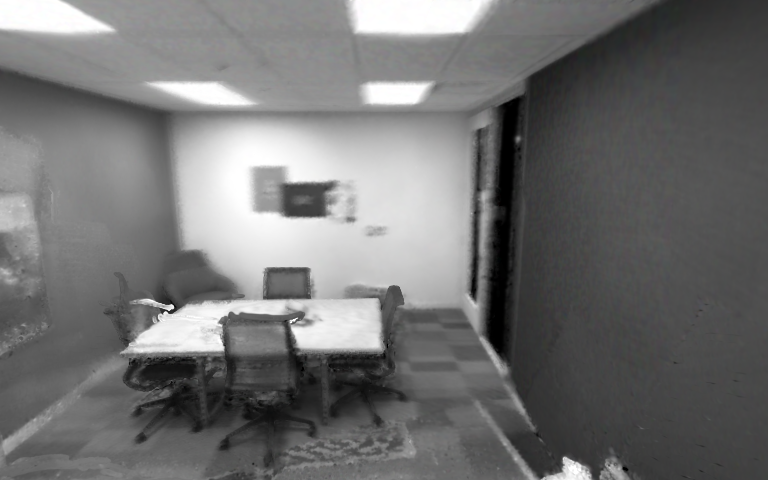} &
        \includegraphics[width=0.186\textwidth]{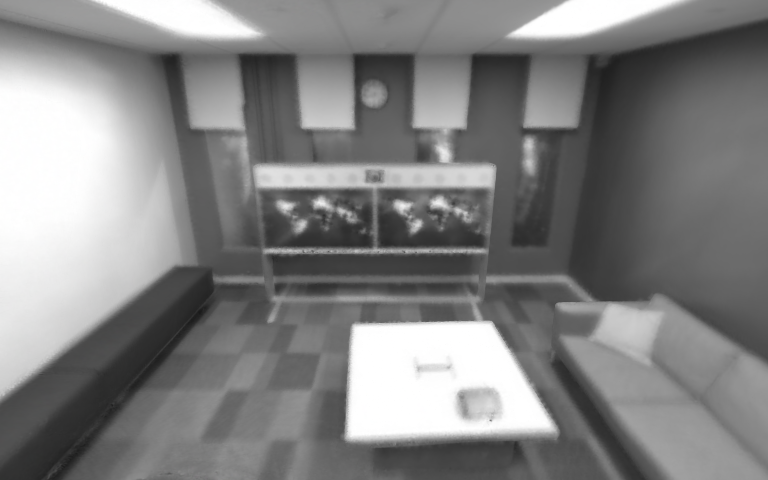} 
         \\

        \raisebox{24pt}{\rotatebox[origin=c]{90}{\tiny MonoGS }}&
         \includegraphics[width=0.186\textwidth]{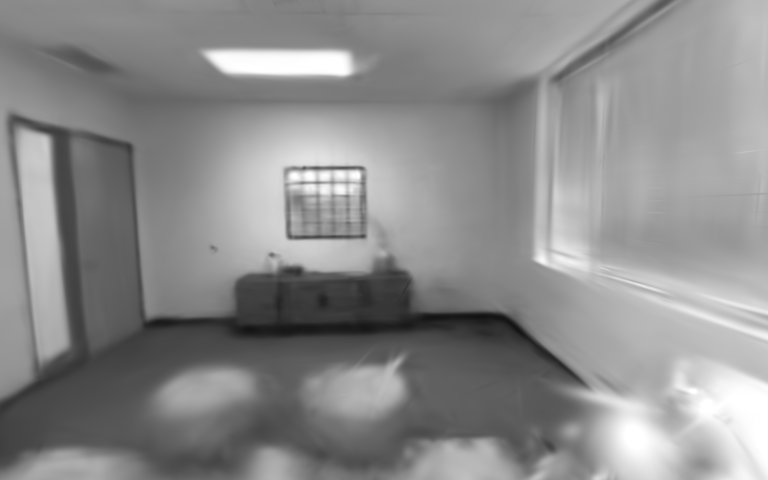} &
        \includegraphics[width=0.186\textwidth]{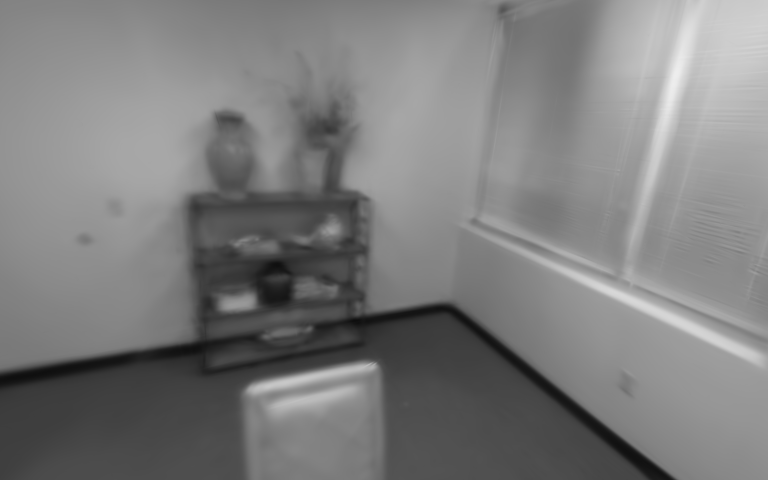} &
        \includegraphics[width=0.186\textwidth]{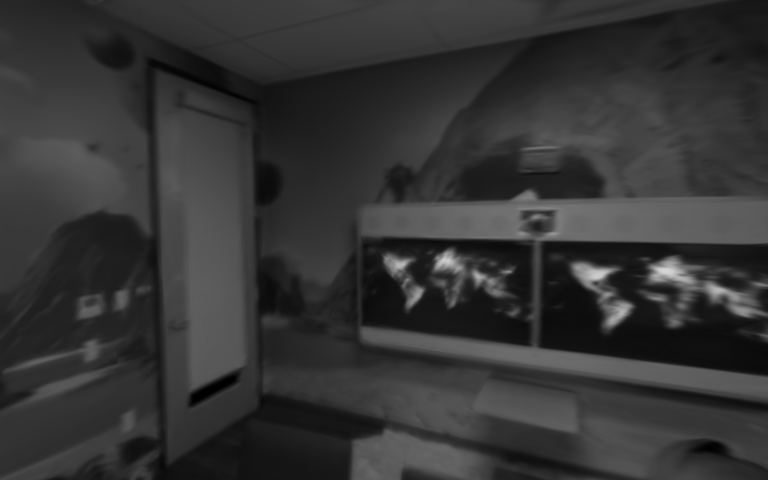} &
        \includegraphics[width=0.186\textwidth]{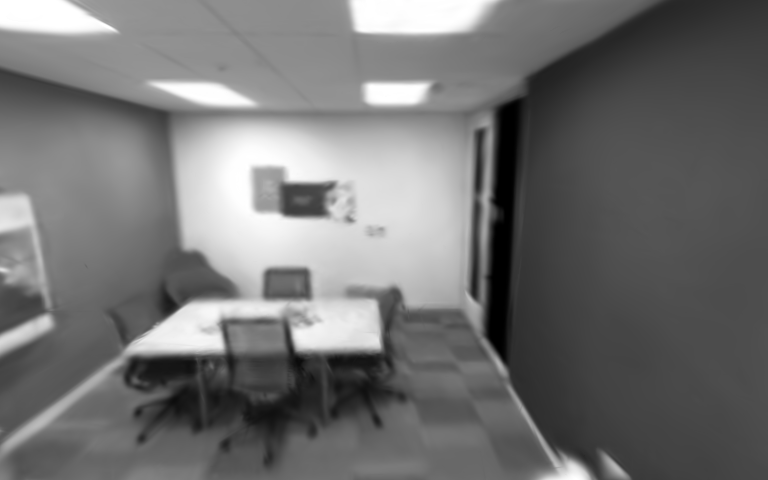} &
        \includegraphics[width=0.186\textwidth]{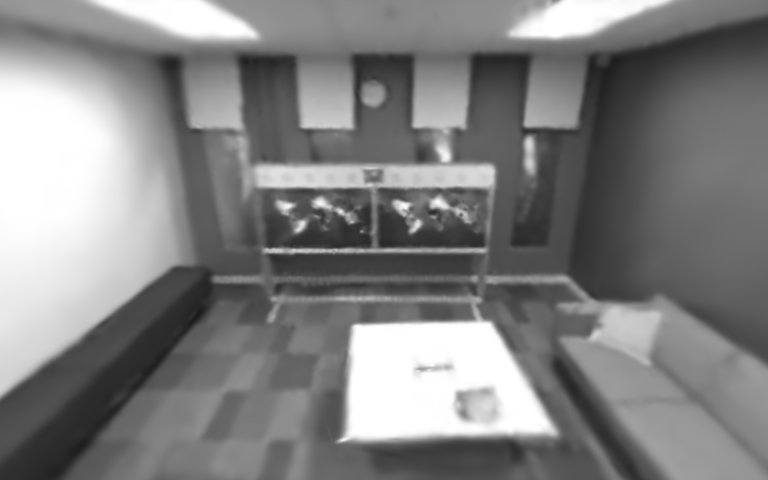} 
         \\

        \raisebox{24pt}{\rotatebox[origin=c]{90}{\tiny Ours}}&
         \includegraphics[width=0.186\textwidth]{imgs/ours/room0_long_f261_013.150s.jpg} &
        \includegraphics[width=0.186\textwidth]{imgs/ours/room2_long_f91_004.650s.jpg} &
        \includegraphics[width=0.186\textwidth]{imgs/ours/office0_long_f81_004.150s.jpg} &
        \includegraphics[width=0.186\textwidth]{imgs/ours/office2_long_f181_009.150s.jpg} &
        \includegraphics[width=0.186\textwidth]{imgs/ours/office3_long_f281_014.150s.jpg} 
         \\

        \raisebox{24pt}{\rotatebox[origin=c]{90}{\tiny Blur}}&
         \includegraphics[width=0.186\textwidth]{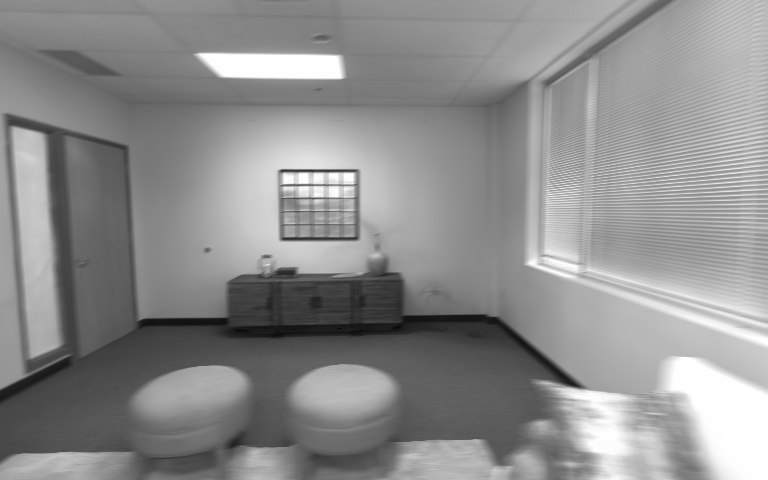} &
        \includegraphics[width=0.186\textwidth]{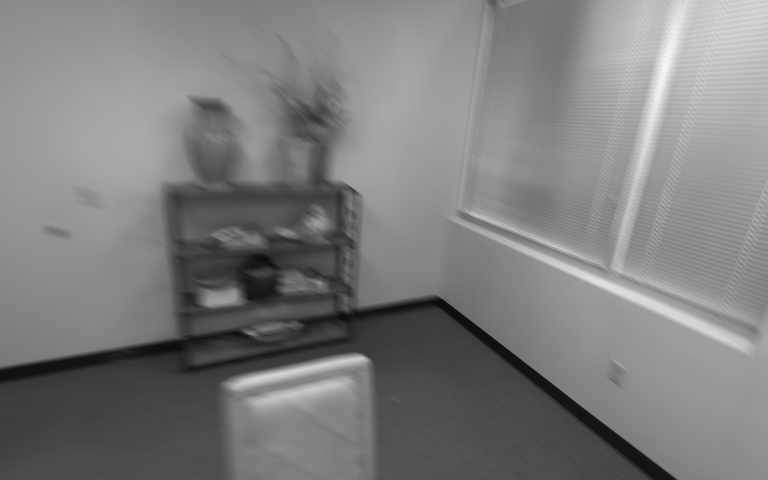} &
        \includegraphics[width=0.186\textwidth]{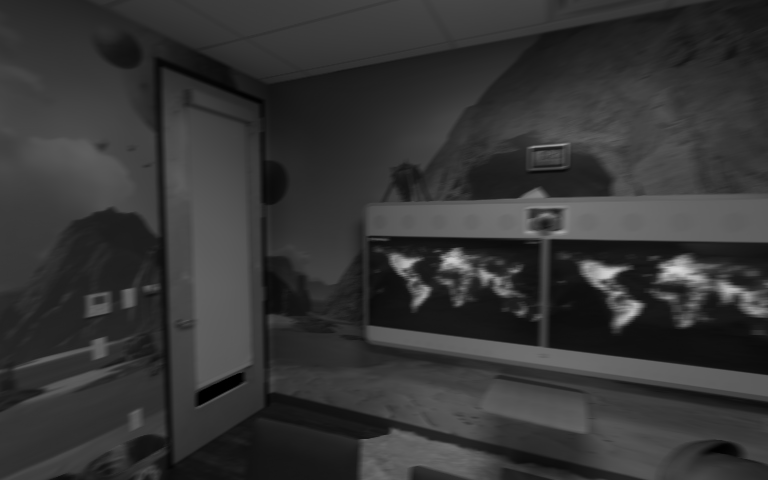} &
        \includegraphics[width=0.186\textwidth]{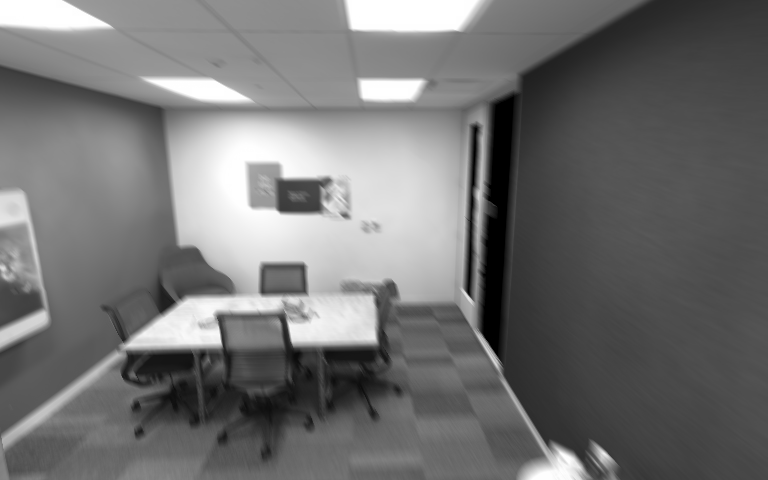} &
        \includegraphics[width=0.186\textwidth]{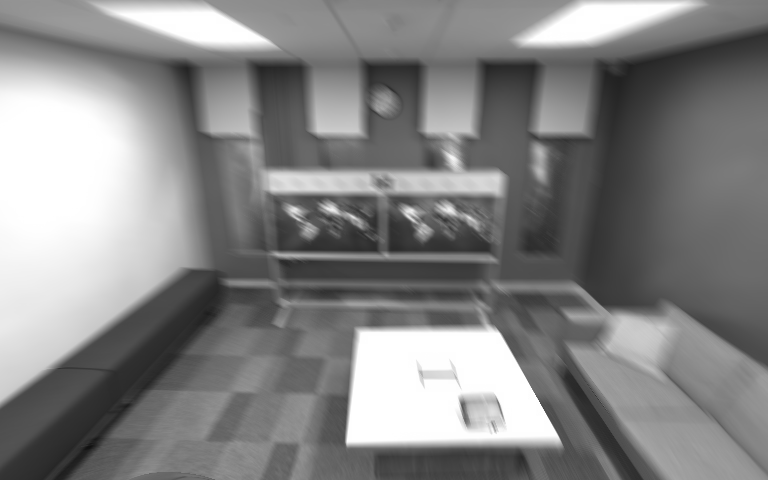} 
         \\
        
        \raisebox{24pt}{\rotatebox[origin=c]{90}{\tiny GT}}&
         \includegraphics[width=0.186\textwidth]{imgs/gt/room0_long_f261_013.150s.jpg} &
        \includegraphics[width=0.186\textwidth]{imgs/gt/room2_long_f91_004.650s.jpg} &
        \includegraphics[width=0.186\textwidth]{imgs/gt/office0_long_f81_004.150s.jpg} &
        \includegraphics[width=0.186\textwidth]{imgs/gt/office2_long_f181_009.150s.jpg} &
        \includegraphics[width=0.186\textwidth]{imgs/gt/office3_long_f281_014.150s.jpg} 
         \\

        \end{tabular}
    }
    \vspace{-10pt}
    \caption{Qualitative evaluation of novel view image synthesis on the Replica dataset. The experimental results demonstrate that our method renders higher-quality images when the camera is moving fast.}
    \label{pic:fast_motion}
    \vspace{-1.8em}
\end{figure*}

\textbf{\section{Experiments in Fast-Motion Scenarios}}
Fast camera movement can induce motion blur, making it challenging to reconstruct the scene and estimate camera poses using RGB-based algorithms. We compare our event-based method with two state-of-the-art pose-free Gaussian SLAM implementations: MonoGS \cite{matsuki2024gaussian} (RGB modality) and SplaTAM \cite{keetha2024splatam} (RGBD modality). By leveraging the high temporal resolution of event cameras, our method experiences minimal performance degradation, even under fast motion. Additionally, it is more effective at preserving high-frequency information in the scene. As shown in Fig. \ref{pic:fast_motion}, our approach delivers superior novel view synthesis results, particularly during rapid camera movement.

\textbf{\section{Experiments on Color Event Datasets}}

Our method can also be applied to color event datasets by integrating the Bayer filter \cite{rudnev2023eventnerf}, as shown below:

\begin{align}
    \cL_{event} &=  \norm{ \bF \odot \bE_i(\bx) - \bF \odot \hat{\bE}_i(\bx)}_2 \\
    \cL_{ssim} &= SSIM(\bF \odot \bE_i(\bx), \bF \odot \hat{\bE}_i(\bx)) 
\end{align}
Furthermore, our method can be extended to incorporate training with ground-truth poses.

\begin{figure*}
    \centering
    \vspace{0.3em}
    \addtolength{\tabcolsep}{-6.5pt}
    \footnotesize{
        \setlength{\tabcolsep}{1pt} 
        \begin{tabular}{p{8.2pt}cccccccc}
            & lego & materials & drums & chair & ficus \\

        \raisebox{36pt}{\rotatebox[origin=c]{90}{\tiny Ours(wo) }}&
         \includegraphics[width=0.186\textwidth]{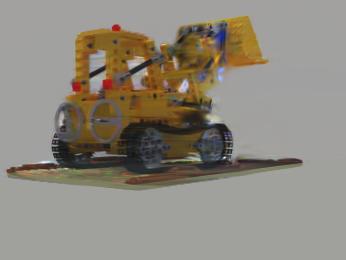} &
        \includegraphics[width=0.186\textwidth]{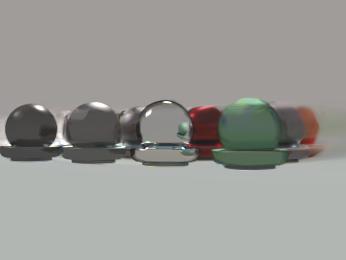} &
        \includegraphics[width=0.186\textwidth]{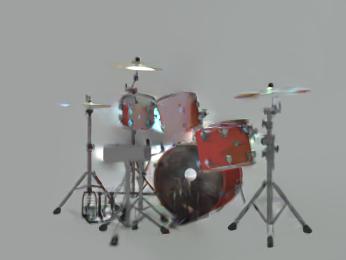} &
        \includegraphics[width=0.186\textwidth]{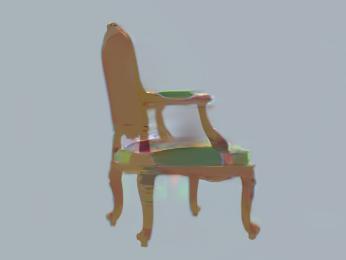} &
        \includegraphics[width=0.186\textwidth]{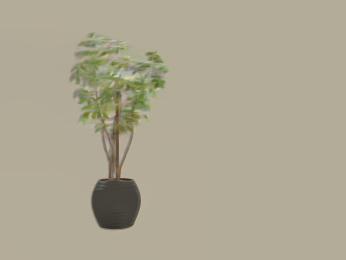} 
         \\

        \raisebox{36pt}{\rotatebox[origin=c]{90}{\tiny Ours(w)}}&
         \includegraphics[width=0.186\textwidth]{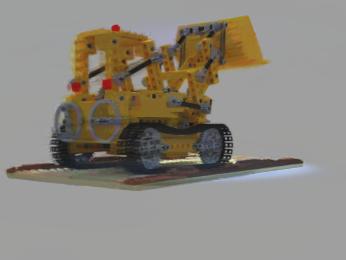} &
        \includegraphics[width=0.186\textwidth]{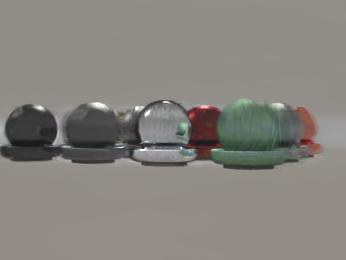} &
        \includegraphics[width=0.186\textwidth]{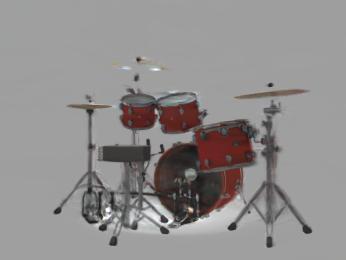} &
        \includegraphics[width=0.186\textwidth]{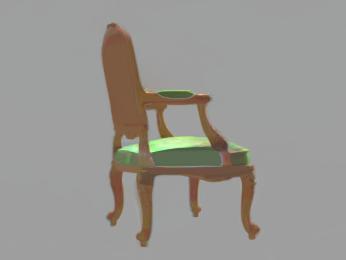} &
        \includegraphics[width=0.186\textwidth]{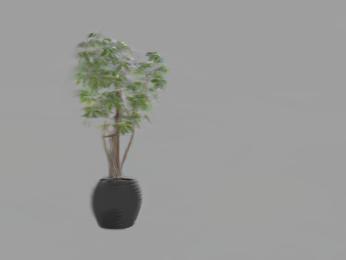} 
         \\

        \raisebox{36pt}{\rotatebox[origin=c]{90}{\tiny EventNeRF}}&
         \includegraphics[width=0.186\textwidth]{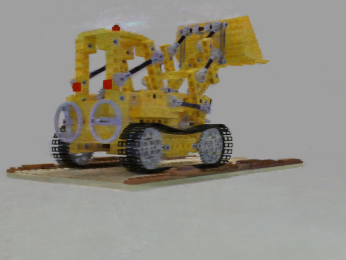} &
        \includegraphics[width=0.186\textwidth]{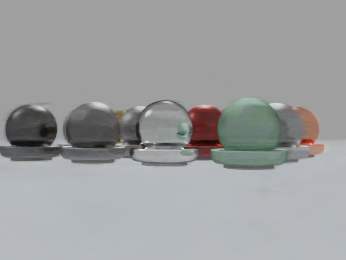} &
        \includegraphics[width=0.186\textwidth]{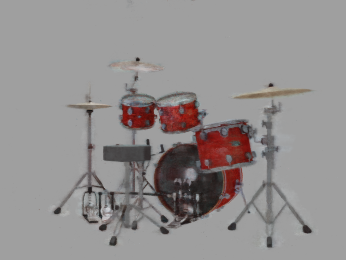} &
        \includegraphics[width=0.186\textwidth]{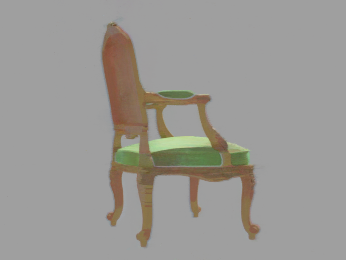} &
        \includegraphics[width=0.186\textwidth]{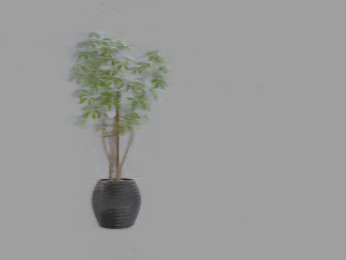} 
         \\
        
        \raisebox{36pt}{\rotatebox[origin=c]{90}{\tiny GT RGB}}&
         \includegraphics[width=0.186\textwidth]{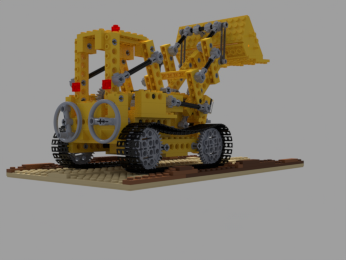} &
        \includegraphics[width=0.186\textwidth]{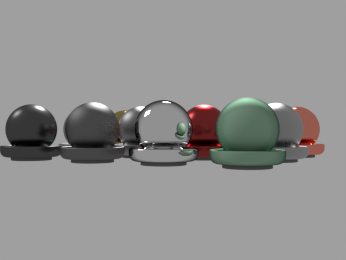} &
        \includegraphics[width=0.186\textwidth]{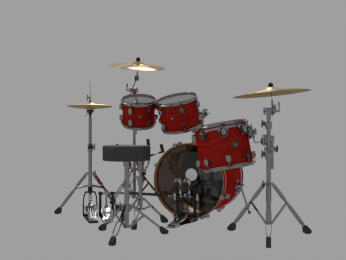} &
        \includegraphics[width=0.186\textwidth]{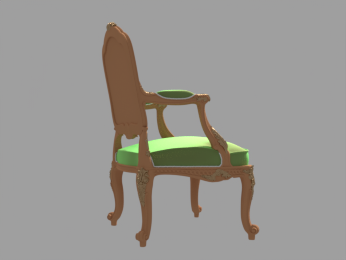} &
        \includegraphics[width=0.186\textwidth]{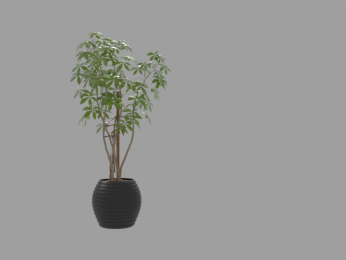} 
         \\

        \end{tabular}
    }
    \vspace{-8pt}
    \caption{Qualitative evaluation of novel view image synthesis on color event dataset. Ours (wo) denotes our method trained without ground-truth camera poses, while Ours (w) denotes the method trained with ground-truth camera poses.}
    \label{pic:color}
    \vspace{-1.9em}
\end{figure*}

We conducted experiments on the EventNeRF dataset \cite{rudnev2023eventnerf}, which focuses on object reconstruction. Due to the dataset's limited features, pose estimation is challenging; neither COLMAP nor DEVO can estimate camera poses on this dataset. As shown in Fig. \ref{pic:color}, our method can still successfully optimize both the 3D scene and camera poses even without ground-truth poses, though it produces minor artifacts. When trained with ground-truth poses, our method achieves improved novel view synthesis, with fewer artifacts and sharper textures.

\begin{table}[]
    \centering
    \scalebox{0.8}{
    \begin{tabular}{c|c|c|c|c}
    \toprule
    Method & \multicolumn{2}{c|}{Synthetic ($768 \times 480$)} & \multicolumn{2}{c}{Real-world ($1280 \times 720$)} \\
    \cmidrule(lr){2-3} \cmidrule(lr){4-5} & Training & Storage & Training & Storage \\
    \midrule
        ENeRF     & 12 hour & 253M & 12 hour & 253M\\
        EventNeRF     &  21 hour & 14M & 24 hour & 14M  \\
        Robust e-NeRF & 11 hour & 745M & 13 hour & 745M\\
        Ours & 0.5 hour & 65M & 2 hour & 55M\\
    \bottomrule
    \end{tabular} 
    }
    \caption{Average model efficiency comparison.}
    \label{tab:time_compaison}
\end{table}

\begin{table*}
    \centering
    \resizebox{\linewidth}{!}{
    \begin{tabular}{c*{15}{c}}
		\toprule
		\multirow{3}{*}{} & \multicolumn{3}{c}{\textit{room0}} & \multicolumn{3}{c}{\textit{room2}} & \multicolumn{3}{c}{\textit{office0}} & \multicolumn{3}{c}{\textit{office2}} & \multicolumn{3}{c}{\textit{office3}} \\
		\cmidrule(lr){2-4} \cmidrule(lr){5-7} \cmidrule(lr){8-10} \cmidrule(lr){11-13} \cmidrule(lr){14-16}
		& {\scriptsize PSNR$\uparrow$} & {\scriptsize SSIM$\uparrow$} & {\scriptsize LPIPS$\downarrow$}  & {\scriptsize PSNR$\uparrow$} & {\scriptsize SSIM$\uparrow$} & {\scriptsize LPIPS$\downarrow$}  & {\scriptsize PSNR$\uparrow$} & {\scriptsize SSIM$\uparrow$} & {\scriptsize LPIPS$\downarrow$}  & {\scriptsize PSNR$\uparrow$} & {\scriptsize SSIM$\uparrow$} & {\scriptsize LPIPS$\downarrow$}  & {\scriptsize PSNR$\uparrow$} & {\scriptsize SSIM$\uparrow$} & {\scriptsize LPIPS$\downarrow$}  \\
		\midrule
		E2GS* & 21.75& 0.77& 0.25& 23.11& 0.82& 0.20& 20.09 & 0.75 & 0.18& 18.62 & 0.78 & 0.20 &  20.13 & 0.84& 0.16\\ 
        EvGGS & 15.16 & 0.37 & 0.62 & 15.85 & 0.34 & 0.61 & 18.51 & 0.37 & 0.59 & 10.95 & 0.27 & 0.69 & 13.13 & 0.29 & 0.66 \\
        Ours       & \textbf{24.31} & \textbf{0.85} & \textbf{0.17} & \textbf{23.75}  & \textbf{0.79} & \textbf{0.23} & \textbf{25.64} & \textbf{0.54} & \textbf{0.30} & \textbf{21.74} & \textbf{0.82} & \textbf{0.23} & \textbf{21.18} & \textbf{0.88} & \textbf{0.13}   \\
		\bottomrule
	\end{tabular}
    }
    \caption{NVS performance comparison on Replica dataset. * denotes we removed the supervision of blurred images from the original E2GS.The result demonstrates that our method outperforms those two baseline methods.}
    \label{tab:nvs_suppl}
\end{table*}

\textbf{\section{Time Evaluations}}

As shown in Table \ref{tab:time_compaison}, our method has a significant advantage in training time compared to NeRF-based methods. Additionally, our method achieves an NVS rendering speed of approximately 500 FPS, whereas NeRF-based methods reach only about 0.5 FPS.

We mainly focus on demonstrating the effectiveness (\ie in terms of novel view synthesis and pose estimation) by exploiting 3D-GS representation for event camera, and have not tried to improve the efficiency of the proposed method. In particular, for the ease of the development, we still adopt the Adam optimizer with a small learning rate (\ie $1\text{e-}4$) from PyTorch for both motion and 3D-GS estimation. It requires around 0.3s and 1.7s per event chunk to converge for both tracking and mapping respectively.  We would further improve the efficiency by using a second-order optimization method (\eg levenberg-marquardt algorithm), which has been proved to converge much faster to the optimal solution compared to an first-order optimizer (\eg Adam).



\end{document}